\documentclass[twoside,11pt]{article}

\usepackage{amsmath}
\usepackage{enumitem}
\usepackage[preprint]{jmlr2e}
\usepackage{float}
\newtheorem{assumption}{Assumption}
\renewenvironment{proof}[1][Proof]
{\par\noindent{\bfseries #1\ }}
{\hfill\BlackBox\\[2mm]}
\usepackage{booktabs}
\usepackage{xcolor}

\usepackage{lastpage}

\jmlrheading{XX}{2026}{1-\pageref{LastPage}}
{10/26}{XX/XX}{XX-0000}
{Jingyi Jin, Martha White and Hao Wang}

\ShortHeadings{Convergence of TDRC with Eligibility Traces}
{Jin, White and Wang}

\firstpageno{1}

\newcommand{\JY}[1]{\textcolor{black}{#1}}

\begin{document}

\title{On Convergence of Gradient Temporal-Difference Learning with Regularized Corrections and Eligibility Traces
}

\author{\name Jingyi Jin\thanks{Corresponding author.}
\email jjin12@ualberta.ca \\
\addr Department of Mathematical and Statistical Sciences\\
Interdisciplinary Lab for Mathematical Ecology and Epidemiology\\
University of Alberta\\
Edmonton, Alberta, Canada
\AND
\name Martha White
\email whitem@ualberta.ca \\
\addr Department of Computing Science\\
University of Alberta and the Alberta Machine Intelligence Institute (Amii)\\
Edmonton, Alberta, Canada
\AND
\name Hao Wang
\email hao8@ualberta.ca \\
\addr Department of Mathematical and Statistical Sciences\\
Interdisciplinary Lab for Mathematical Ecology and Epidemiology\\
University of Alberta\\
Edmonton, Alberta, Canada}

\editor{To be assigned}

\maketitle

\makeatletter
\def\ps@arxivtitle{%
  \ps@plain
  \def\@oddfoot{\parbox{\textwidth}{\scriptsize
    \copyright~2026 Jingyi Jin, Martha White and Hao Wang.}}%
  \let\@evenfoot\@oddfoot
}
\makeatother
\thispagestyle{arxivtitle}

\begin{abstract}
Temporal-difference (TD) learning underlies many prediction and
control algorithms, but can diverge when combined with off-policy
learning, bootstrapping, and function approximation. TD with
Regularized Corrections (TDRC) is theoretically sound and practically
simple, but existing convergence theory covers only the one-step case
under i.i.d.\ sampling, while the Huber Bellman-error formulation has
not been extended to TDRC or eligibility traces. In this work, we
establish convergence of constrained TDRC($\lambda$) with diminishing stepsizes and non-i.i.d. off-policy data, for state-dependent and history-dependent $\lambda$-parameters. We also propose TDRC($\lambda$)-Huber, which preserves the TDRC($\lambda$) fixed point while reducing sensitivity to large Bellman errors. For both algorithms, we apply an ODE-based weak-convergence method from stochastic approximation theory. For the nonlinear mean dynamics of TDRC($\lambda$)-Huber, we give an integral small-gain condition that guarantees uniqueness and global asymptotic stability of the target equilibrium. We then establish convergence of the constrained algorithm, derive computable sufficient conditions for convergence, and identify a regularization-based convergence mechanism shared by linear TDRC($\lambda$) and TDRC($\lambda$)-Huber. When the mean TD matrix is positive stable, both constrained algorithms converge under sufficiently strong regularization without requiring positive
definiteness.
\end{abstract}


\section{Introduction}

Off-policy value estimation with function approximation is a central problem in reinforcement learning (RL) \citep{patterson2022generalized}. 
Temporal-difference (TD) learning provides an efficient bootstrapping method for estimating value functions and forms the basis of many prediction and control algorithms. 
However, combining off-policy learning, bootstrapping, and function approximation can lead to divergence. 
Several widely used algorithms, including classical TD, semi-gradient TD, and Q-learning, may diverge in this setting \citep{baird1995residual,tsitsiklis1997analysis,SuttonRichardS.2018RLAI,elelimy2025deep}. 
Conversely, algorithms with theoretical guarantees, such as Gradient TD methods \citep{sutton2009fast}, Emphatic TD \citep{sutton2016etd}, and Hybrid TD methods \citep{hackman2012faster,white2016investigating}, are often difficult to use in practice \citep{ghiassian2020gradient}. 
Designing off-policy TD algorithms that
are both theoretically sound and practically effective therefore
remains an important problem.

TD with Regularized Corrections (TDRC) was introduced as a
gradient-based algorithm that addresses some of these limitations
\citep{ghiassian2020gradient,patterson2022generalized}. TDRC adds an
$\ell_2$ regularizer to the secondary update of TD with Corrections
(TDC). This regularizer is designed to retain the theoretical
soundness of TDC while improving its practical behavior. In
particular, the regularization parameter $\beta$ controls the
interpolation between TDC and TD: setting $\beta=0$ recovers TDC,
whereas increasing $\beta$ makes TDRC behave more like TD. The
resulting algorithm has been studied from both empirical and
theoretical perspectives. Empirically, regularization has been shown
to reduce variance and accelerate the learning of the secondary
weights without incurring much bias
\citep{ghiassian2020gradient,patterson2022generalized}. More recent
experiments have further shown that the multistep extension
TDRC($\lambda$) performs effectively in deep reinforcement learning
and often exhibits stable and fast learning among the gradient-TD
variants considered in \citep{elelimy2025deep}. On the theoretical
side, \citep{ghiassian2020gradient} established convergence of
one-step TDRC under the assumption of i.i.d.\ state transitions.

Most value function learning algorithms, including TDRC, are based on the mean
squared (projected) Bellman error. A related difficulty is
that squared errors are sensitive to outliers, which can both skew the
solution of the objective and produce high-magnitude and high-variance
updates. To reduce the influence of such errors,
\citep{patterson2023robust} formulated Huber and absolute Bellman
errors and derived gradient-based algorithms for their optimization.
In particular, their TDC-Huber algorithm clips the estimate of the
secondary variable before using it in the gradient-correction update.
The resulting Huber algorithms showed improved stability and reduced
sensitivity to meta-parameters.

Despite these developments, two related gaps remain. First, the
existing convergence analysis of TDRC covers only the one-step case
under i.i.d.\ state transitions
\citep{ghiassian2020gradient}. 
For multistep
gradient-TD algorithms with positive $\lambda$, the eligibility
traces are constructed from a sequence of non-i.i.d.\ off-policy
transitions and gather information about the multistep Bellman
operator \citep{Yugtd}. Consequently, the existing one-step analysis
does not directly cover TDRC($\lambda$), for which no corresponding
convergence result is currently available.  Second, existing Huber gradient-TD algorithms are based on one-step
updates \citep{patterson2023robust}, and their extension to TDRC($\lambda$) with general
eligibility traces has not been proposed and analyzed. Such an extension may
combine the improved stability and reduced sensitivity to
meta-parameters observed for TDC-Huber with the regularization and
multistep credit assignment of TDRC($\lambda$). However, the
extension is not immediate: the $\lambda$-parameters induce a
generalized Bellman operator and a state--trace process, while Huber
clipping makes the coupled mean field nonlinear. It therefore remains
to determine whether the resulting algorithm preserves the
TDRC($\lambda$) fixed point and converges using non-i.i.d.\
off-policy data.

In this work, we address the two gaps identified above. We first
establish convergence results for TDRC($\lambda$) with
state-dependent and history-dependent $\lambda$-parameters, using
non-i.i.d.\ off-policy data. We then formulate a Huber extension of
TDRC($\lambda$) and establish its convergence under the same
settings. More specifically, our contributions are as follows.
\begin{enumerate}

\item
\textbf{Convergence of TDRC($\lambda$).}
We establish convergence of constrained single-time-scale
TDRC($\lambda$) with state-dependent and history-dependent
$\lambda$-parameters using non-i.i.d.\ off-policy data.

\item
\textbf{TDRC($\lambda$)-Huber and a regularization-based
convergence mechanism.}
We propose a new algorithm, TDRC($\lambda$)-Huber, that accommodates
state-dependent and history-dependent $\lambda$-parameters. We show that the proposed
algorithm preserves the same fixed point as TDRC($\lambda$) and
establish convergence of the constrained algorithm under an integral
small-gain condition. Our analysis further identifies a
large-$\beta$ convergence mechanism arising from regularization.
This mechanism applies to both linear TDRC($\lambda$) and
TDRC($\lambda$)-Huber. The clipping induced by the Huber loss reduces sensitivity to large
Bellman errors while preserving this convergence guarantee.




\item
\JY{\textbf{Small-gain analysis for TDRC($\lambda$)-Huber.}
We formulate the nonlinear mean dynamics of TDRC($\lambda$)-Huber in feedback form. This representation separates the coupled dynamics into a Hurwitz linear subsystem and a globally Lipschitz nonlinear feedback term arising from clipping. We then apply an integral small-gain condition to establish stability of the resulting nonlinear system. This analysis shows that the target equilibrium is unique and globally asymptotically stable.}

\item
\textbf{Computable convergence conditions and the role of the
regularization parameter $\beta$.}
We derive computable sufficient conditions
for convergence, covering state-dependent and history-dependent
$\lambda$-parameters as well as the one-step case. These conditions clarify the roles of $\beta$ and the secondary
stepsize $\eta$ in the convergence analysis.
\end{enumerate}

The scope of these results can be compared through the matrix
$A_\lambda$, defined formally in
Section~\ref{subsec:projected-bellman-error}, which governs the mean
TD($\lambda$) dynamics. In the classical on-policy analysis,
$A_\lambda$ is positive definite
\citep{tsitsiklis1997analysis}. In this case, the linear
TDRC($\lambda$) condition holds for all $\beta\geq0$ and
$\eta>0$. When $A_\lambda$ is nonsingular but not positive
definite, the sufficient condition in
Lemma~\ref{lem:tdrc-hurwitz-condition}, obtained by directly
analyzing the coefficient matrix of the coupled linear mean ODE,
involves coupled bounds on $\beta$ and $\eta$. For the
positive-stable subclass, a different feedback representation,
developed in Section~\ref{sec:huber-associated-mean-odes}, reveals
an additional large-$\beta$ convergence mechanism driven by
regularization. For every $\eta>0$, the resulting computable
sufficient condition holds for both linear TDRC($\lambda$) and
TDRC($\lambda$)-Huber when $\beta$ is sufficiently large.

Our proofs combine properties of the state--trace process with the ODE-based weak-convergence method from stochastic approximation theory \citep{kushner2003}. Several TD methods with eligibility traces, including LSTD($\lambda$), ETD($\lambda$), and GTD($\lambda$), have been analyzed within this framework \citep{Yu2012lstd,yu2015convergence,yu2016weak,Yugtd}. Our treatment of the primary and secondary iterates differs from the two-time-scale analysis of GTDb in Section~3.2 of \citep{Yugtd}. In that analysis, the secondary iterate evolves on a faster time scale, and the slow mean ODE is obtained by replacing the secondary variable with its equilibrium for each fixed value of the primary variable. In TDRC($\lambda$), the two updates use stepsizes $\alpha_n$ and $\eta\alpha_n$, respectively, where $\eta$ is fixed. The iterates therefore evolve on the same time scale and are analyzed jointly through a coupled mean ODE. For TDRC($\lambda$), this ODE is linear, so its stability can be studied directly. For TDRC($\lambda$)-Huber, Huber clipping introduces a nonlinear correction to the mean ODE. We separate the resulting dynamics into a linear
subsystem and a globally Lipschitz nonlinear feedback term, and
establish stability through an integral small-gain condition. These
ODE stability results allow the weak-convergence method to be applied
to the constrained iterates.

The paper is organized as follows. In
Section~\ref{sec:preliminaries}, we introduce the off-policy evaluation setting, generalized Bellman equations, and properties of
the state--trace process. In Section~\ref{sec:tdrc-lambda}, we present
TDRC($\lambda$) and TDRC($\lambda$)-Huber and characterize their
target fixed points. In Section~\ref{sec:tdrc-convergence}, we give
the convergence analysis of constrained TDRC($\lambda$). In
Section~\ref{sec:huber-convergence}, we analyze the convergence of constrained
TDRC($\lambda$)-Huber. In Section~\ref{sec:role-of-beta}, we discuss the role of
$\beta$ and illustrate the convergence conditions using a four-state
off-policy prediction problem. We conclude the paper in
Section~\ref{sec:conclusions}. The appendices contain the deferred
proofs, supporting stability results, stochastic approximation
verifications, and derivations of the computable conditions.

\section{Preliminaries}\label{sec:preliminaries}

In this section, we first describe the off-policy evaluation problem. We then present two choices of $\lambda$-parameters, the associated generalized Bellman equations, and the linear function approximation framework. Finally, to prepare for the subsequent convergence analysis, we introduce the state--trace process underlying the TD algorithms and review the relevant properties established by \citep[Section~2.2]{Yugtd}.

\subsection{Off-policy Evaluation and Linear Function Approximation}

We consider policy evaluation in a finite Markov decision process
(MDP). At time $t$, the agent observes $S_t\in\mathcal S$ and selects
$A_t\in\mathcal A$. Conditional on $S_t=s$ and $A_t=a$, the
environment transitions to $S_{t+1}=s'$ with probability
$p(s'\mid s,a)$ and generates a random reward $R_{t+1}$. We
assume that, conditional on $(S_t,A_t,S_{t+1})=(s,a,s')$, the reward
has mean $r(s,s')$ and finite variance.
The policy to be evaluated is a stationary \textit{target
policy} $\pi:\mathcal S\times\mathcal A\to[0,1]$, where
$\pi(a\mid s)$ denotes the probability of selecting action $a$ in
state $s$.

The performance of the target policy is evaluated through its expected discounted sum of
future rewards:
\begin{equation}
v_\pi(s)
:=
\mathbb{E}_s^\pi
\left[
r_\pi(S_0)
+
\sum_{n=1}^{\infty}
\left(
\prod_{k=1}^{n}\gamma(S_k)
\right)
r_\pi(S_n)
\right],
\qquad \text{for all }
s\in\mathcal{S}.
\label{value-function}
\end{equation}
Here, $r_\pi:\mathcal S\to\mathbb R$ is the expected one-step
reward under $\pi$, $\gamma:\mathcal S\to[0,1]$ is the state-dependent discount
factor, and $\mathbb E_s^\pi$ denotes expectation for the Markov
chain induced by $\pi$ with $S_0=s$. The function $v_\pi$ in \eqref{value-function} is called the \textit{value function} of $\pi$. Condition~\ref{cond:target-behavior}(i)
below ensures that $v_\pi$ is well defined.

In the off-policy setting, transitions and rewards are generated by
a \textit{behavior policy}
$b:\mathcal S\times\mathcal A\to[0,1]$. Let $P$ and $P^o$ denote
the transition matrices on $\mathcal S$ induced by the target and
behavior policies, respectively, and let
$\Gamma=\operatorname{diag}(\gamma(s):s\in\mathcal S)$. We impose
the following standard condition, adapted from Condition~2.1 of
\citep{Yugtd}.

\begin{condition}[Target and behavior policies]
\label{cond:target-behavior}
The transition matrices $P$ and $P^o$ satisfy the following:
\begin{itemize}
\item[(i)] The inverse $(I-P\Gamma)^{-1}$ exists.

\item[(ii)] For all $s,s'\in\mathcal{S}$,
\[
P^o_{ss'}=0
\quad\Longrightarrow\quad
P_{ss'}=0,
\]
and the Markov chain induced by $P^o$ is irreducible.
\end{itemize}
\end{condition}

Under Condition~\ref{cond:target-behavior}(i), the value function $v_\pi$ is well-defined and is the unique solution of the Bellman equation
\begin{equation*}
v_\pi=r_\pi+P\Gamma v_\pi,
\end{equation*}
or, equivalently, $v_\pi=(I-P\Gamma)^{-1}r_\pi$.
Moreover, Condition~\ref{cond:target-behavior}(ii) ensures that $P^o$ has a unique invariant probability distribution $\xi$. Let $\Xi$ denote the diagonal matrix whose diagonal entries are the components of $\xi$. For $s,s'\in\mathcal{S}$, define the importance-sampling ratio
\[
\rho(s,s')
=
\frac{P_{ss'}}{P^o_{ss'}},
\]
with the convention that $\rho(s,s')=0$ when $P^o_{ss'}=0$. These ratios correct for the difference between the transition dynamics induced by the target and behavior policies. For simplicity, we write $\rho_t=\rho(S_t,S_{t+1})$.

To approximate the value function $v_\pi$, let
$\phi:\mathcal{S}\to\mathbb{R}^d$ be a feature mapping, where 
   \[ \phi(s)
    =
    \bigl(\phi_1(s),\ldots,\phi_d(s)\bigr)^\top\] is the feature vector associated with state $s$. The component
functions $\phi_1,\ldots,\phi_d$ span the linear approximation
subspace
    $\mathcal{L}_\phi
    :=
    \operatorname{span}
    \{\phi_1,\ldots,\phi_d\}$.
Each approximate value function in $\mathcal{L}_\phi$ can be
parameterized as $v_\theta(s)
    =
    \phi(s)^\top\theta$. Let $\Phi\in\mathbb{R}^{|\mathcal{S}|\times d}$ denote the
corresponding feature matrix whose row associated with state $s$
is $\phi(s)^\top$. After identifying each function on the finite
state space with the vector of its values at all states, the
approximation subspace can equivalently be written as
\begin{equation*}
    \mathcal{L}_\phi
    =
    \operatorname{col}(\Phi)
    =
    \left\{
        \Phi\theta:
        \theta\in\mathbb{R}^d
    \right\}
    \subseteq\mathbb{R}^{|\mathcal{S}|}.
\end{equation*}
Thus, in matrix notation, the approximate value function is
$v_\theta=\Phi\theta\in\mathcal{L}_\phi$, and the objective is to
find a function in $\mathcal{L}_\phi$ that approximates $v_\pi$.

\subsection{Generalized Bellman Equations and
\texorpdfstring{$\lambda$}{lambda}-Parameters}

To incorporate multistep credit assignment, we consider the
eligibility-trace sequence $\{e_t\}$ defined by
\begin{equation*}
    e_t
    =
    \lambda_t\gamma_t\rho_{t-1}e_{t-1}
    +
    \phi(S_t),
    \qquad t\geq 1,
\end{equation*}
where $e_0\in\mathbb{R}^d$ is an initial trace vector,
$\lambda_t\in[0,1]$ is the trace-decay parameter at time $t$, and
$\gamma_t=\gamma(S_t)$. Following \citep{Yugtd}, we consider two
choices of $\lambda_t$.

\begin{enumerate}
    \item[(i)] \emph{State-dependent $\lambda$.}
    In this case,
    \begin{equation*}
        \lambda_t=\lambda(S_t),
        \qquad
        \lambda:\mathcal{S}\rightarrow[0,1].
    \end{equation*}

    \item[(ii)] \emph{History-dependent $\lambda$.}
    In this case, $\lambda_t$ depends on the previous trace
    $e_{t-1}$ and an auxiliary memory state $y_t$. More precisely,
    \begin{equation}
        y_t=f(y_{t-1},S_t),
        \qquad
        \lambda_t=\lambda(y_t,e_{t-1}),
        \label{eq:history-dependent-lambda}
    \end{equation}
    where $f$ and $\lambda$ are prescribed functions.
\end{enumerate}

State-dependent $\lambda$ is a special case of history-dependent
$\lambda$. The main motivation for using history-dependent $\lambda$
is that it provides greater flexibility in controlling the eligibility
traces while allowing $\lambda_t$ to take values over a broad range.
Following \citet[Section~2.1.3]{Yugtd}, we restrict our attention to
history-dependent trace constructions satisfying the following additional conditions.
These conditions will be used in the next subsection to establish the
required properties of the state--trace process.

\begin{condition}[Memory-state process]
\label{cond:memory-state}
The memory states $\{y_t\}$ take values in a finite set
$\mathcal{M}$. Under the behavior policy $b$, the Markov chain
$\{(S_t,y_t)\}$ on $\mathcal{S}\times\mathcal{M}$ has a single
recurrent class.
\end{condition}

\begin{condition}[History-dependent $\lambda$]
\label{cond:history-dependent-lambda}
The function $\lambda:\mathcal{M}\times\mathbb{R}^d\rightarrow[0,1]$
satisfies the following properties. There exists a norm
$\|\cdot\|$ on $\mathbb{R}^d$ such that, for every
$y\in\mathcal{M}$:
\begin{enumerate}
    \item[(i)] for all $e,e'\in\mathbb{R}^d$, $\left\|
            \lambda(y,e)e-\lambda(y,e')e'
        \right\|
        \leq
        \|e-e'\|$;

    \item[(ii)] there exists a constant $C_y<\infty$ such that
    \[
        \left\|
            \gamma(s')\rho(s,s')\lambda(y,e)e
        \right\|
        \leq C_y
    \]
    for all $e\in\mathbb{R}^d$ and all possible state transitions
    $(s,s')$ that can lead to the memory state $y$.
\end{enumerate}
\end{condition}

The $\lambda$-parameters affect the TD algorithms through the
eligibility traces and also determine the generalized Bellman equation
to be solved. In particular, the target value function $v_\pi$ is the
unique solution of
\begin{equation}
    v_\pi=T^{(\lambda)}v_\pi,
    \label{eq:generalized-bellman-fixed-point}
\end{equation}
where $T^{(\lambda)}$ is an affine operator on
$\mathbb{R}^{|\mathcal{S}|}$ of the form
\begin{equation*}
    T^{(\lambda)}v
    =
    r_\pi^{(\lambda)}+P^{(\lambda)}v,
    \qquad
    v\in\mathbb{R}^{|\mathcal{S}|},
\end{equation*}
with $r_\pi^{(\lambda)}\in\mathbb{R}^{|\mathcal{S}|}$ and
$P^{(\lambda)}$ being a substochastic matrix. Under linear function
approximation, TD algorithms seek an approximate solution to
\eqref{eq:generalized-bellman-fixed-point} in the approximation
subspace $\operatorname{col}(\Phi)$. The detailed construction and
properties of $T^{(\lambda)}$ for state-dependent and
history-dependent $\lambda$ are provided by
\citep[Section~2.1.3]{Yugtd} and are not repeated here.

In most of our analysis, we treat the two choices of $\lambda$
together. We therefore collect the conditions for both cases in the
following assumption.
\begin{assumption}
\label{assump:lambda-conditions}
Condition~\ref{cond:target-behavior} holds. For the
history-dependent $\lambda$ defined in
\eqref{eq:history-dependent-lambda},
Conditions~\ref{cond:memory-state} and
\ref{cond:history-dependent-lambda} also hold.
\end{assumption}

\subsection{Properties of the State--Trace Process}
\label{sec:stprocess}
Following \citep[Section~2.2]{Yugtd}, we review the key properties of the state--trace process generated under
the behavior policy. These properties provide the foundation for the subsequent weak convergence analysis of the algorithms within the stochastic approximation framework \citep{kushner2003}.

In the state-dependent case, define the state--trace
process by $X_t=(S_t,e_t)$,
and in the history-dependent case by $X_t=(S_t,y_t,e_t)$.
To prepare for handling the TD updates, which also involve the successor state,
we further define $Z_t=(S_t,e_t,S_{t+1})$ in the state-dependent case, and $Z_t=(S_t,y_t,e_t,S_{t+1})$
in the history-dependent case. In both cases, we use
$\mathcal{Z}$ to denote the space of $Z_t$.

For an initial condition $x$ of the state--trace process, define its
occupation probability measure by
\begin{equation*}
    \mu_{x,n}(D)
    =
    \frac{1}{n+1}
    \sum_{t=0}^{n}
    \mathbf{1}_{\{X_t\in D\}},
\end{equation*}
for every Borel subset $D$ of the state--trace space. Thus,
$\mu_{x,n}(D)$ represents the proportion of time that the process
spends in $D$ during the first $n+1$ time steps.

Under Condition~\ref{cond:target-behavior} and, in the
history-dependent case, Conditions~\ref{cond:memory-state}
and~\ref{cond:history-dependent-lambda}, the state--trace process has
the following properties established by
\citep[Theorem~2.1 and Propositions~2.2--2.3]{Yugtd}. Let $\mathbb{P}_x$ denote the probability measure associated with the process starting from the initial condition $x$, and $\mathbb{P}_x$-a.s. means almost surely with respect to $\mathbb{P}_x$.

\begin{proposition}[Properties of the state--trace process]
\label{prop:state-trace-properties} The following statements hold:
\begin{enumerate}
    \item[(i)] The state--trace process $\{X_t\}$ is a weak Feller
    Markov chain and has a unique invariant probability measure
    $\zeta$. For every initial condition $x$, the sequence of
    occupation probability measures $\{\mu_{x,n}\}$ converges weakly
    to $\zeta$, $\mathbb{P}_x$-a.s.

    \item[(ii)] Let $\mathbb{E}_{\zeta}$ denote expectation with
    respect to the stationary state--trace process initialized
    according to $\zeta$. If
    $f:\mathcal{Z}\rightarrow\mathbb{R}^k$ is Lipschitz continuous
    in the trace variable $e$, then $\mathbb{E}_{\zeta}
        \left[\|f(Z_0)\|\right]
        <\infty$.
    Moreover, for every initial condition of the process,
    \begin{equation*}
        \lim_{n\to\infty}
        \frac{1}{n}
        \sum_{t=0}^{n-1}f(Z_t)
        =
        \mathbb{E}_{\zeta}\!\left[f(Z_0)\right]
    \end{equation*}
    both in mean and almost surely.

    \item[(iii)] For every initial trace $e_0$, the family of trace
    iterates $\{e_t\}_{t\geq0}$ is uniformly integrable; that is,
    \begin{equation*}
        \lim_{K\to\infty}
        \sup_{t\geq0}
        \mathbb{E}
        \left[
            \|e_t\|
            \mathbf{1}_{\{\|e_t\|>K\}}
        \right]
        =
        0.
    \end{equation*}

    \item[(iv)] For any function $f$ satisfying the conditions in
    part~(ii), let
    $\bar{f}
        =
        \mathbb{E}_{\zeta}[f(Z_0)]$.
    Then, for every compact set $D\subset\mathcal{Z}$,
    \begin{equation*}
        \lim_{m,n\to\infty}
        \frac{1}{m}
        \sum_{i=n}^{n+m-1}
        \bigl(f(Z_i)-\bar{f}\bigr)
        \mathbf{1}_{\{Z_n\in D\}}
        =
        0
        \qquad\text{in mean}.
    \end{equation*}
\end{enumerate}
\end{proposition}

The preceding properties will be used to verify the averaging
conditions needed to identify the mean ODEs associated with the
algorithms. The stability properties of these ODEs will then be used
to establish the convergence of the algorithms.

We next record several identities used in deriving the associated
mean ODEs. For $v\in\mathbb{R}^{|\mathcal{S}|}$, define the TD error by
\begin{equation*}
    \bar{\delta}_t(v)
    =
    \rho_t
    \left(
        r(S_t,S_{t+1})
        +
        \gamma_{t+1}v(S_{t+1})
        -
        v(S_t)
    \right).
\end{equation*}

The following identities are established in
\citep[Proposition~2.1]{Yugtd}:
\begin{align}
\mathbb{E}_{\zeta}
\left[
\phi(S_0)\phi(S_0)^\top
\right]
&=
\Phi^\top\Xi\Phi,
\\
\mathbb{E}_{\zeta}
\left[
e_0\bar{\delta}_0(v)
\right]
&=
\Phi^\top\Xi
\left(
T^{(\lambda)}v-v
\right),
\qquad
\forall\,v\in\mathbb{R}^{|\mathcal{S}|},
\\
\mathbb{E}_{\zeta}
\left[
e_0\rho_0
\left(
\phi_i(S_0)-\gamma_1\phi_i(S_1)
\right)
\right]
&=
\Phi^\top\Xi
\left(
I-P^{(\lambda)}
\right)\Phi_i,
\qquad
1\leq i\leq d,
\\
\mathbb{E}_{\zeta}
\left[
e_0\rho_0(1-\lambda_1)\gamma_1\phi_i(S_1)
\right]
&=
\Phi^\top\Xi P^{(\lambda)}\Phi_i,
\qquad
1\leq i\leq d.
\label{identities}
\end{align}
These identities will be used later to derive the mean fields and associated mean ODEs
of the algorithms.

\section{\texorpdfstring{TDRC($\lambda$)}{TDRC(lambda)}
and Its Huber Extension}
\label{sec:tdrc-lambda}

This section introduces TDRC($\lambda$) and its Huber extension.
We first define the mean squared projected Bellman error (MSPBE) and
characterize its unique minimizer $\theta_\lambda^*$. We then present
TDRC($\lambda$) and show that $(\theta_\lambda^*,0)$ is a fixed point
of its expected-update equations. Finally, we introduce one-step
TDRC-Huber and extend it to general eligibility traces. The pair
$(\theta_\lambda^*,0)$ remains a fixed point of the Huber extension,
while its uniqueness and stability are established in the subsequent
convergence analysis.

\subsection{Projected Bellman Error and Its Minimizer}
\label{subsec:projected-bellman-error}

To find a function $v\in\mathcal{L}_\phi$ that approximately satisfies
$v=T^{(\lambda)}v$, we consider the projected Bellman equation on the
linear approximation subspace
\begin{equation*}
\mathcal{L}_\phi
:=
\operatorname{col}(\Phi)
=
\left\{
\Phi\theta:\theta\in\mathbb{R}^d
\right\}.
\end{equation*}

Let $\xi$ denote the invariant probability distribution of the Markov
chain induced by the behavior policy, and let $\Xi$ be the diagonal
matrix with diagonal entries $\xi_s$. Since the behavior-policy Markov
chain is irreducible, $\xi_s>0$ for every $s\in\mathcal{S}$. Define the
weighted Euclidean norm by
\begin{equation*}
\lVert v\rVert_\xi^2
:=
\sum_{s\in\mathcal{S}}\xi_s v(s)^2
=
v^\top\Xi v.
\end{equation*}

Define
\begin{equation*}
C=\Phi^\top\Xi\Phi.
\end{equation*}
We assume that $\Phi$ has full column rank. Since $\Xi$ is positive
definite, $C$ is positive definite and hence nonsingular. The
projection onto $\mathcal{L}_\phi$ with respect to
$\lVert\cdot\rVert_\xi$ is therefore
\begin{equation*}
\Pi_\xi
=
\Phi C^{-1}\Phi^\top\Xi.
\end{equation*}

The projected Bellman equation associated with $T^{(\lambda)}$ is
\begin{equation}
\Phi\theta
=
\Pi_\xi T^{(\lambda)}(\Phi\theta).
\label{eq:projected-bellman-equation}
\end{equation}
Define
\begin{equation*}
A_\lambda
=
\Phi^\top\Xi\left(I-P^{(\lambda)}\right)\Phi,
\qquad
b_\lambda
=
\Phi^\top\Xi r_\pi^{(\lambda)}.
\end{equation*}
Then \eqref{eq:projected-bellman-equation} is equivalent to
\begin{equation*}
A_\lambda\theta=b_\lambda.
\end{equation*}

For $v_\theta=\Phi\theta$, the associated MSPBE is
\begin{align*}
J_\lambda(\theta)
&:=
\frac{1}{2}
\left\|
\Pi_\xi
\left(
T^{(\lambda)}v_\theta-v_\theta
\right)
\right\|_\xi^2
\nonumber=
\frac{1}{2}
\left(
b_\lambda-A_\lambda\theta
\right)^\top
C^{-1}
\left(
b_\lambda-A_\lambda\theta
\right).
\end{align*}

We assume throughout that $A_\lambda$ is nonsingular. Under this
assumption, the projected Bellman equation has the unique solution
\begin{equation*}
\theta_\lambda^*
=
A_\lambda^{-1}b_\lambda.
\end{equation*}
Moreover,
\[
J_\lambda(\theta_\lambda^*)=0.
\]
Since $J_\lambda(\theta)\geq0$ for every
$\theta\in\mathbb{R}^d$, with equality if and only if
$A_\lambda\theta=b_\lambda$, the vector $\theta_\lambda^*$ is also
the unique minimizer of $J_\lambda$. Consequently,
\begin{equation*}
v_{\theta_\lambda^*}
=
\Phi\theta_\lambda^*
\end{equation*}
is the unique element of $\mathcal{L}_\phi$ satisfying the projected
Bellman equation.

The MSPBE is used to characterize $\theta_\lambda^*$ and to relate the
algorithms to the gradient-TD framework. The proposed algorithms need
not follow the exact gradient of the MSPBE. Instead, we verify directly
that $(\theta_\lambda^*,0)$ is a fixed point of their expected-update
equations and establish convergence by analyzing the associated mean
ODEs.

\subsection{\texorpdfstring{TDRC($\lambda$)}{TDRC(lambda)}}

Recall from Section~\ref{sec:preliminaries} the importance-sampling
ratio $\rho_t$, the state-dependent discount factor $\gamma_t$, and
the eligibility trace $e_t$. For the linear approximation
$v_\theta=\Phi\theta$, define the sample TD error by
\begin{equation*}
\delta_t(\theta)
=
\rho_t
\left(
R_{t+1}
+
\gamma_{t+1}\phi(S_{t+1})^\top\theta
-
\phi(S_t)^\top\theta
\right).
\end{equation*}

The main idea of TDRC($\lambda$) is to regularize the secondary
weights using the $\ell_2$ regularizer
$\frac{\beta}{2}\|x\|_2^2$, where $\beta\geq0$ is the
regularization parameter. Let $\eta>0$ be the secondary-stepsize, so that $\eta\alpha_t$ is the stepsize for the secondary
update. The TDRC($\lambda$) updates are
\begin{align*}
x_{t+1}
&=
x_t
+
\eta\alpha_t
\left[
e_t\delta_t(\theta_t)
-
\phi(S_t)\phi(S_t)^\top x_t
-
\beta x_t
\right],
\\
\theta_{t+1}
&=
\theta_t
+
\alpha_t
\left[
e_t\delta_t(\theta_t)
-
\rho_t(1-\lambda_{t+1})\gamma_{t+1}
\phi(S_{t+1})e_t^\top x_t
\right].
\end{align*}
The primary update has the same form
as in TDC($\lambda$), while the secondary update includes the
additional term $-\beta x_t$. When $\beta=0$, the update reduces
to TDC($\lambda$). \JY{A detailed discussion of the role of $\beta$ is given in
Section~\ref{sec:role-of-beta}.}

Under linear function approximation, a \textit{fixed point} is a parameter
value that remains unchanged under the expected update
\citep[Section~9.4]{SuttonRichardS.2018RLAI}. Accordingly, a
parameter pair $(\theta,x)$ is called a \textit{fixed point} of
TDRC($\lambda$) if both expected updates are zero. As shown in
Section~\ref{subsec:tdrc-main-convergence}, taking stationary expectations
of the TDRC($\lambda$) updates and using the identities established
in Section~\ref{sec:stprocess} gives the associated mean ODE.
Therefore, a fixed point is equivalently an equilibrium of this ODE
and satisfies
\begin{align}
0
&=
b_\lambda-A_\lambda\theta
+
(A_\lambda^\top-C)x,
\label{eq:fixed-point-theta}
\\
0
&=
b_\lambda-A_\lambda\theta
-
(C+\beta I)x.
\label{eq:fixed-point-x}
\end{align}
The following proposition identifies this fixed point and establishes
its uniqueness.

\begin{proposition}[Fixed point of TDRC($\lambda$)]
\label{prop:tdrc-fixed-point}
Suppose that $A_\lambda$ and $A_\lambda+\beta I$ are nonsingular.
Then the expected TDRC($\lambda$) recursion has the unique fixed
point
\begin{equation*}
(\theta^*,x^*)
=
(\theta_\lambda^*,0),
\qquad
\theta_\lambda^*
=
A_\lambda^{-1}b_\lambda.
\end{equation*}
Moreover, $\theta_\lambda^*$ is the unique minimizer of
$J_\lambda(\theta)$.
\end{proposition}
The proof is given in
Appendix~\ref{app:tdrc-fixed-point-proof}.

Thus, although TDRC($\lambda$) does not follow the MSPBE gradient,
it preserves the unique
minimizer of $J_\lambda(\theta)$ as its unique
fixed point. \JY{The subsequent convergence analysis considers the
constrained algorithm, whose iterates are confined to a bounded
Euclidean ball $B(R)$, and establishes convergence to that unique fixed point from every initialization $U_0\in B(R)$. This is not a serious restriction in practice because, for any given finite initial point $U_0$, the radius $R$ can be chosen sufficiently large so that $U_0\in B(R)$, after which the convergence result applies.}

\subsection{\texorpdfstring{TDRC($\lambda$)-Huber}
{TDRC(lambda)-Huber}}

Squared errors are sensitive to large TD errors, which may produce
high-magnitude and high-variance updates. To address this issue,
\citep{patterson2023robust} introduced TDC-Huber in the one-step
setting, where $\lambda_t\equiv0$ and $e_t=\phi(S_t)$
for $t\geq1$.
This algorithm incorporates clipping into the correction term. Following
this approach, we obtain one-step TDRC-Huber and then extend it to
state-dependent and history-dependent eligibility traces, yielding
TDRC($\lambda$)-Huber.

For a threshold $\tau>0$, define the Huber loss by
\begin{equation*}
\ell_\tau(u)
=
\begin{cases}
\dfrac{1}{2}u^2,
& |u|\leq\tau,
\\[2mm]
\tau|u|-\dfrac{1}{2}\tau^2,
& |u|>\tau.
\end{cases}
\end{equation*}
The Huber loss admits the biconjugate representation
\begin{equation*}
\ell_\tau(u)
=
\max_{h\in[-\tau,\tau]}
\left\{
uh-\frac{1}{2}h^2
\right\}
\end{equation*}
\citep[Proposition~4.1]{patterson2023robust}.
Thus, the auxiliary variable $h$ in this representation is
constrained to $[-\tau,\tau]$. Following
\citet{patterson2023robust}, we impose the analogous bound on the
scalar correction estimate using the clipping function
\begin{equation*}
\operatorname{clip}_\tau(u)
=
\begin{cases}
-\tau,
& u<-\tau,
\\
u,
& |u|\leq\tau,
\\
\tau,
& u>\tau.
\end{cases}
\end{equation*}
We apply the same clipping construction to TDRC. We first introduce
the one-step version and then extend it to state-dependent and
history-dependent eligibility traces.

When $\lambda_t\equiv0$, the eligibility trace satisfies
$e_t=\phi(S_t)$ for $t\geq1$. We define the TDRC(0)-Huber recursion by
\begin{align*}
x_{t+1}
&=
x_t
+
\eta\alpha_t
\left[
\phi(S_t)\delta_t(\theta_t)
-
\phi(S_t)\phi(S_t)^\top x_t
-
\beta x_t
\right],
\\
\theta_{t+1}
&=
\theta_t
+
\alpha_t
\left[
\phi(S_t)\delta_t(\theta_t)
-
\rho_t\gamma_{t+1}\phi(S_{t+1})
\operatorname{clip}_\tau
\left(
\phi(S_t)^\top x_t
\right)
\right].
\end{align*}
This update combines TDC-Huber with
the regularized secondary update of TDRC. When $\beta=0$, it reduces
to TDC-Huber.

We extend this construction to eligibility traces by
replacing $\phi(S_t)$ in the correction term with $e_t$. The
resulting TDRC($\lambda$)-Huber update is
\begin{align*}
x_{t+1}
&=
x_t
+
\eta\alpha_t
\left[
e_t\delta_t(\theta_t)
-
\phi(S_t)\phi(S_t)^\top x_t
-
\beta x_t
\right],
\\
\theta_{t+1}
&=
\theta_t
+
\alpha_t
\left[
e_t\delta_t(\theta_t)
-
\rho_t(1-\lambda_{t+1})\gamma_{t+1}
\phi(S_{t+1})
\operatorname{clip}_\tau(e_t^\top x_t)
\right].
\end{align*}
Thus, TDRC(0)-Huber is obtained by setting
$\lambda_t\equiv0$. Moreover,
\begin{equation*}
\lim_{\tau\rightarrow\infty}
\operatorname{clip}_\tau(u)
=
u,
\end{equation*}
so TDRC($\lambda$)-Huber reduces to TDRC($\lambda$) as
$\tau\rightarrow\infty$.

To identify the target fixed point, define the stationary mean
correction
\begin{equation*}
\mathcal H_{\lambda,\tau}(x)
:=
\mathbb{E}_{\zeta}
\left[
\rho_0(1-\lambda_1)\gamma_1\phi(S_1)
\operatorname{clip}_\tau(e_0^\top x)
\right].
\end{equation*}
As derived in Section~\ref{sec:huber-associated-mean-odes}, the equilibria
of the associated nonlinear mean ODE, equivalently the fixed points
of the expected update, satisfy
\begin{align}
0
&=
b_\lambda-A_\lambda\theta
-
\mathcal H_{\lambda,\tau}(x),
\label{eq:huber-fixed-point-theta}
\\
0
&=
b_\lambda-A_\lambda\theta
-
(C+\beta I)x.
\label{eq:huber-fixed-point-x}
\end{align}

\begin{proposition}[Fixed point of
TDRC($\lambda$)-Huber]
\label{prop:tdrc-huber-fixed-point}
Suppose that $A_\lambda$ is nonsingular. Then
\begin{equation*}
(\theta^*,x^*)
=
(\theta_\lambda^*,0),
\qquad
\theta_\lambda^*
=
A_\lambda^{-1}b_\lambda,
\end{equation*}
is a fixed point of TDRC($\lambda$)-Huber.
Its $\theta$-component is the minimizer of $J_\lambda(\theta)$
identified in Section~\ref{subsec:projected-bellman-error}.
In particular, TDRC(0)-Huber has the fixed point
\begin{equation*}
(\theta_0^*,0),
\qquad
\theta_0^*
=
A_0^{-1}b_0.
\end{equation*}
\end{proposition}
The proof is given in
Appendix~\ref{app:tdrc-huber-fixed-point-proof}.


\JY{This proposition establishes the existence of the target fixed point; that is, it shows that \((\theta_\lambda^*,0)\) is a fixed point of TDRC($\lambda$)-Huber. The later analysis in Section \ref{sec:huber-convergence} establishes its uniqueness and convergence to it. Moreover, this point coincides with the unique fixed point of TDRC($\lambda$), and its \(\theta\)-component minimizes \(J_\lambda(\theta)\). }

\JY{More specifically, because clipping makes the expected update nonlinear in \(x\), the nonsingularity of \(A_\lambda\) alone does not rule out additional fixed points of TDRC($\lambda$)-Huber. The convergence conditions introduced later, including the required gain condition, establish that \((\theta_\lambda^*,0)\) is the unique globally asymptotically stable equilibrium of the associated nonlinear mean ODE. Under these conditions, it is therefore the unique fixed point of the expected recursion, and the stochastic approximation analysis establishes convergence to it. The case \(\lambda_t\equiv0\) is stated separately because the subsequent analysis treats the general-trace and one-step cases separately.}

\section{Convergence Analysis of
\texorpdfstring{TDRC($\lambda$)}{TDRC(lambda)}}
\label{sec:tdrc-convergence}

This section establishes the convergence of constrained
TDRC($\lambda$) with diminishing stepsizes, a setting commonly used
in practice. Following the weak convergence method for constrained
stochastic approximation \citep{kushner2003}, we first state the main
convergence results. We then develop the supporting analysis by
deriving the associated mean ODE and establishing the stability of
its equilibrium. Finally, we verify the required stochastic
approximation conditions and complete the proofs of the main results.

\subsection{Associated Mean ODE and Main Convergence Results}
\label{subsec:tdrc-main-convergence}

In this subsection, we formulate constrained TDRC($\lambda$) as a
projected stochastic approximation recursion, derive its associated
mean ODE, and establish the stability property needed for the main
convergence results.

Let
\[
U_n
=
\begin{pmatrix}
x_n\\
\theta_n
\end{pmatrix}
\in\mathbb{R}^{2d},
\qquad
B(R)
=
\left\{
u\in\mathbb{R}^{2d}:\|u\|_2\leq R
\right\},
\]
and let $\Pi_{B(R)}$ denote the Euclidean projection onto $B(R)$.
The constrained TDRC($\lambda$) recursion is
\begin{align}
U_{n+1}
=
\Pi_{B(R)}
\begin{pmatrix}
x_n+\eta\alpha_n
\left[
e_n\delta_n(\theta_n)
-\phi(S_n)\phi(S_n)^\top x_n
-\beta x_n
\right]
\\[2mm]
\theta_n+\alpha_n
\left[
e_n\delta_n(\theta_n)
-\rho_n(1-\lambda_{n+1})\gamma_{n+1}
\phi(S_{n+1})e_n^\top x_n
\right]
\end{pmatrix}.
\label{eq:constrained-tdrc-lambda}
\end{align}
The projection ensures that the iterates remain bounded. The radius
$R$ is chosen sufficiently large; a precise lower bound is given
below in \eqref{eq:tdrc-projection-radius}.

Following \citep{yu2016weak}, we impose the following
diminishing-stepsize condition. It includes slowly decreasing
stepsizes such as $\alpha_n=(n+1)^{-c}$ for $c\in(0,1]$.

\begin{assumption}[Diminishing stepsizes]
\label{assump:stepsize}
The deterministic sequence $\{\alpha_n\}$ satisfies
\[
\alpha_n>0,
\qquad
\alpha_n\longrightarrow0,
\qquad
\sum_{n=0}^{\infty}\alpha_n=\infty.
\]
Moreover, there exists a sequence of positive integers
$m_n\rightarrow\infty$ such that
\begin{equation*}
\lim_{n\rightarrow\infty}
\sup_{0\leq j\leq m_n}
\left|
\frac{\alpha_{n+j}}{\alpha_n}-1
\right|
=0.
\end{equation*}
\end{assumption}

We next rewrite the update in the more concise form used in the
weak-convergence analysis of \citep{Yugtd}. Let $Z_n$ be the
state--trace process defined in
Section~\ref{sec:stprocess}. From now on, we write
\[
Z_n=(S_n,y_n,e_n,S_{n+1})
\]
in both cases, treating the memory state $y_n$ as a dummy variable
in the state-dependent case. Let $\mathcal Z$ denote the state
space of $\{Z_n\}$.

For $z=(s,y,e,s')$, define
\begin{align*}
\bar\delta(s,s',v_\theta)
&=
\rho(s,s')
\left[
r(s,s')
+\gamma(s')\phi(s')^\top\theta
-\phi(s)^\top\theta
\right],
\\
k(\theta,x,z)
&=
e\,\bar\delta(s,s',v_\theta)
-\phi(s)\phi(s)^\top x,
\\
h(\theta,x,z)
&=
\eta
\left[
k(\theta,x,z)-\beta x
\right].
\end{align*}
For state-dependent $\lambda$, define
\begin{equation*}
\begin{aligned}
g(\theta,x,z)
&=
e\,\bar\delta(s,s',v_\theta)
-\rho(s,s')
\left(1-\lambda(s')\right)
\gamma(s')\phi(s')e^\top x.
\end{aligned}
\end{equation*}
For history-dependent $\lambda$, define
\begin{equation*}
\begin{aligned}
g(\theta,x,z)
&=
e\,\bar\delta(s,s',v_\theta)
-\rho(s,s')
\left(1-\lambda(f(y,s'),e)\right)
\gamma(s')\phi(s')e^\top x,
\end{aligned}
\end{equation*}
where $f$ is the function in
\eqref{eq:history-dependent-lambda} that determines the next memory
state.

The functions $k$ and $g$ follow the notation of
\citep[Eqs.~(3.43)--(3.46)]{Yugtd}. The function $k$ corresponds
to the secondary update of TDC($\lambda$), while $h$ includes the
regularization term $-\beta x$ and the secondary-stepsize multiplier
$\eta$ used in TDRC($\lambda$).

Let
\[
\omega_{n+1}
=
\rho_n
\left(
R_{n+1}-r(S_n,S_{n+1})
\right),
\]
which has zero mean given the history. The constrained
update can then be written as
\begin{align*}
U_{n+1}
=
\Pi_{B(R)}
\begin{pmatrix}
x_n+\alpha_n
\left[
h(\theta_n,x_n,Z_n)
+\eta e_n\omega_{n+1}
\right]
\\[2mm]
\theta_n+\alpha_n
\left[
g(\theta_n,x_n,Z_n)
+e_n\omega_{n+1}
\right]
\end{pmatrix}.
\end{align*}

Let
\[
u
=
\begin{pmatrix}
x\\
\theta
\end{pmatrix},
\qquad
F(u,z)
=
\begin{pmatrix}
h(\theta,x,z)\\
g(\theta,x,z)
\end{pmatrix},
\qquad
\varepsilon_{n+1}
=
\begin{pmatrix}
\eta e_n\omega_{n+1}\\
e_n\omega_{n+1}
\end{pmatrix}.
\]
Then the update has the compact form
\begin{equation}
U_{n+1}
=
\Pi_{B(R)}
\left[
U_n+\alpha_n
\left(
F(U_n,Z_n)+\varepsilon_{n+1}
\right)
\right].
\label{eq:tdrc-sa-form}
\end{equation}

For each fixed $(\theta,x)$, let
\[
\bar h(\theta,x)
:=
\mathbb E_\zeta
\left[
h(\theta,x,Z_0)
\right],
\qquad
\bar g(\theta,x)
:=
\mathbb E_\zeta
\left[
g(\theta,x,Z_0)
\right],
\]
where $\mathbb E_\zeta$ denotes expectation with respect to the
stationary distribution of the state--trace process. By the
stationary identities in Section~\ref{sec:stprocess},
\begin{align}
\bar h(\theta,x)
&=
\eta
\left[
\Phi^\top\Xi
\left(
T^{(\lambda)}v_\theta-v_\theta
\right)
-(\Phi^\top\Xi\Phi+\beta I)x
\right]=
\eta
\left[
b_\lambda-A_\lambda\theta
-(C+\beta I)x
\right],
\label{eq:tdrc-mean-h}
\\
\bar g(\theta,x)
&=
\Phi^\top\Xi
\left(
T^{(\lambda)}v_\theta-v_\theta
\right)
-
\left(
\Phi^\top\Xi P^{(\lambda)}\Phi
\right)^\top x
=
b_\lambda-A_\lambda\theta
+(A_\lambda^\top-C)x.
\end{align}
The final equality follows from $-
\left(
\Phi^\top\Xi P^{(\lambda)}\Phi
\right)^\top
=
A_\lambda^\top-C$.

Define
\begin{equation*}
G^\lambda
=
\begin{pmatrix}
-\eta(C+\beta I)&-\eta A_\lambda\\
A_\lambda^\top-C&-A_\lambda
\end{pmatrix},
\qquad
g^\lambda
=
\begin{pmatrix}
\eta b_\lambda\\
b_\lambda
\end{pmatrix}.
\end{equation*}
Then
\begin{equation*}
\begin{pmatrix}
\bar h(\theta,x)\\
\bar g(\theta,x)
\end{pmatrix}
=
G^\lambda u+g^\lambda.
\end{equation*}

The associated projected mean ODE is
\begin{equation}
\dot u
=
G^\lambda u+g^\lambda+z,
\qquad
z\in-\mathcal N_{B(R)}(u),
\label{eq:tdrc-projected-mean-ode}
\end{equation}
where $z$ is the boundary-reflection term induced by the projection.
For the Euclidean ball,
\begin{equation*}
\mathcal N_{B(R)}(u)
=
\begin{cases}
\{0\},
& \|u\|_2<R,
\\
\{au:a\geq0\},
& \|u\|_2=R.
\end{cases}
\end{equation*}
Thus, $z=0$ in the interior. On the boundary, when the projection is
active, $z=-au$ for some $a\geq0$.

We now state the conditions used to establish the stability of this
ODE. Let
\[
H_\lambda
=
\frac{A_\lambda+A_\lambda^\top}{2}.
\]
A real square matrix is called Hurwitz if all its eigenvalues have
negative real parts, and positive stable if all its eigenvalues have
positive real parts.
The following lemma gives a sufficient condition for the
stability of the linear mean ODE.

\begin{lemma}[A sufficient Hurwitz condition]
\label{lem:tdrc-hurwitz-condition}
Suppose that $A_\lambda$ is nonsingular. If
\begin{equation}
0\leq\beta
<
\begin{cases}
+\infty,
& H_\lambda\succeq0,
\\[1ex]
\dfrac{1}{
\lambda_{\max}
\left(
-(A_\lambda A_\lambda^\top)^{-1}H_\lambda
\right)
},
& H_\lambda\not\succeq0,
\end{cases}
\qquad
\eta>
\max
\left\{
0,
-\lambda_{\min}
\left(
C^{-1}H_\lambda
\right)
\right\},
\label{eq:tdrc-spectral-condition}
\end{equation}
then $G^\lambda$ is Hurwitz and
$A_\lambda+\beta I$ is nonsingular.
\end{lemma}

The condition in Lemma~\ref{lem:tdrc-hurwitz-condition} can be
satisfied for every nonsingular $A_\lambda$. It covers all possible cases of $H_\lambda$. In particular, there is
no upper bound on $\beta$ when $H_\lambda\succeq0$, whereas a finite
positive upper bound is obtained when $H_\lambda$ has a negative
eigenvalue. The eigenvalues in
\eqref{eq:tdrc-spectral-condition} are real because
$(A_\lambda A_\lambda^\top)^{-1}H_\lambda$ and
$C^{-1}H_\lambda$ are similar to symmetric matrices. The proof, including its
relation to Conditions C1--C3 in
\citep[Appendix~H.3.1]{ghiassian2020gradient}, is given in
Appendix~\ref{app:tdrc-hurwitz-proof}.

Under \eqref{eq:tdrc-spectral-condition},
Lemma~\ref{lem:tdrc-hurwitz-condition} shows that $G^\lambda$ is Hurwitz, and its unique equilibrium is
\begin{equation*}
u^*
=
-(G^\lambda)^{-1}g^\lambda
=
\begin{pmatrix}
0\\
\theta_\lambda^*
\end{pmatrix},
\qquad
\theta_\lambda^*
=
A_\lambda^{-1}b_\lambda.
\end{equation*}
This is the same fixed point identified in
Proposition~\ref{prop:tdrc-fixed-point}.

Let $P_G\succ0$ be the unique solution of the Lyapunov equation
\begin{equation*}
(G^\lambda)^\top P_G+P_G G^\lambda=-I,
\end{equation*}
and define
\begin{equation}
R_0
=
\max
\left\{
\|u^*\|_2,
\frac{\|P_Gu^*\|_2}{\lambda_{\min}(P_G)}
\right\}.
\label{eq:tdrc-projection-radius}
\end{equation}
The following lemma shows that choosing $R>R_0$ preserves the
stability of $u^*$ under projection.

\begin{lemma}[Stability under Euclidean projection]
\label{sol-pro-ODE}
Suppose that \(G^\lambda\) is Hurwitz, and let
\[
u^*
=
-(G^\lambda)^{-1}g^\lambda.
\]
Let \(P_G\succ0\) be the unique solution of
\[
(G^\lambda)^\top P_G+P_GG^\lambda=-I.
\]
If
\begin{equation*}
R
>
R_0
:=
\max\left\{
\|u^*\|_2,\,
\frac{\|P_Gu^*\|_2}{\lambda_{\min}(P_G)}
\right\},
\end{equation*}
then \(u^*\in\operatorname{int}B(R)\) and is the unique globally
asymptotically stable equilibrium of
\eqref{eq:tdrc-projected-mean-ode} relative to \(B(R)\).
Consequently, the limit set of the projected ODE associated with
\(B(R)\) satisfies $L_{B(R)}=\{u^*\}$.
\end{lemma}
The radius $R$ of the constraint set can be chosen sufficiently large. Thus, although the lower bound $R_0$ may be conservative, the requirement $R>R_0$ is not a serious restriction. When $R>R_0$, $u^*$ lies in the interior of $B(R)$, and the projection does not change the limiting point. The projection is inactive near $u^*$ and only prevents the iterates from leaving a bounded region.

The proof is given in Appendix~\ref{app:tdrc-projection-proof}. The lemma
provides the ODE stability property required by the weak-convergence
analysis. We now state the corresponding convergence results.

For $k\geq0$ and $T>0$, define
\begin{equation*}
m(k,T)
=
\max
\left\{
m\geq k:
\sum_{j=k}^{m-1}\alpha_j\leq T
\right\},
\end{equation*}
and, for $D\subset\mathbb R^{2d}$ and $\delta>0$, define
\begin{equation*}
N_\delta(D)
=
\left\{
u:
\inf_{v\in D}\|u-v\|_2<\delta
\right\}.
\end{equation*}
For a singleton, we write
\[
N_\delta(u^*)
:=
N_\delta(\{u^*\})
=
\left\{
u:\|u-u^*\|_2<\delta
\right\}.
\]

\begin{theorem}[Convergence of constrained TDRC($\lambda$)]
\label{thm:tdrc-convergence}
Suppose that Assumption~\ref{assump:lambda-conditions} holds.
Suppose also that $A_\lambda$ is nonsingular and that
\eqref{eq:tdrc-spectral-condition} in
Lemma~\ref{lem:tdrc-hurwitz-condition} holds. Let
$\{\alpha_n\}$ be diminishing stepsizes satisfying
Assumption~\ref{assump:stepsize}. Choose the constraint-set radius $R>R_0$ as required by
Lemma~\ref{sol-pro-ODE}. For every initial condition of the
state--trace process, including every initial trace $e_0$, and every
$U_0\in B(R)$, there exists a
sequence $T_k\rightarrow\infty$ such that, for every $\delta>0$,
\begin{equation}
\limsup_{k\rightarrow\infty}
\mathbb P
\left(
U_i\notin N_\delta(u^*)
\text{ for some }
i\in[k,m(k,T_k)]
\right)
=0.
\label{eq:tdrc-window-convergence}
\end{equation}
Consequently,
\begin{equation*}
U_n\longrightarrow u^*
\qquad\text{in probability}.
\end{equation*}
\end{theorem}

\begin{corollary}[Convergence in mean]
\label{cor:tdrc-mean-convergence}
Under the conditions of
Theorem~\ref{thm:tdrc-convergence},
\begin{equation*}
\lim_{n\rightarrow\infty}
\mathbb E
\left[
\|U_n-u^*\|_2
\right]
=0.
\end{equation*}
\end{corollary}

The preceding analysis studies $G^\lambda$ directly and yields the
sufficient condition in
Lemma~\ref{lem:tdrc-hurwitz-condition}. A different feedback
representation reveals an additional large-$\beta$ convergence
mechanism driven by regularization. The feedback representation underlying this
result is developed in
Section~\ref{sec:huber-associated-mean-odes}, and the proof is given in
Appendix~\ref{app:tdrc-main-proof}.

\begin{corollary}[A large-$\beta$ condition for TDRC($\lambda$)]
\label{cor:tdrc-positive-stable-large-beta}
Suppose that Assumption~\ref{assump:lambda-conditions} holds and that
$A_\lambda$ is positive stable. Let $P_{A,\lambda}\succ0$ be the
unique solution of
\begin{equation*}
A_\lambda^\top P_{A,\lambda}
+
P_{A,\lambda}A_\lambda
=
I,
\end{equation*}
and define
\begin{equation*}
\kappa_2(P_{A,\lambda})
:=
\frac{\lambda_{\max}(P_{A,\lambda})}
     {\lambda_{\min}(P_{A,\lambda})},
\qquad
\overline L_\lambda
:=
\gamma_{\max}\phi_{\max}^2
\,\xi^\top(I-P\Gamma)^{-1}\mathbf 1.
\end{equation*}
If $\eta>0$ and $\beta\geq0$ satisfy
\begin{equation*}
\beta
>
2\overline L_\lambda
\lambda_{\max}(P_{A,\lambda})
\sqrt{\kappa_2(P_{A,\lambda})}
\|A_\lambda\|_2
-
\lambda_{\min}(C),
\end{equation*}
then $G^\lambda$ is Hurwitz. Consequently, under the stepsize and
projection-radius conditions of
Theorem~\ref{thm:tdrc-convergence}, all conclusions of
Theorem~\ref{thm:tdrc-convergence} and
Corollary~\ref{cor:tdrc-mean-convergence} hold.
\end{corollary}

Thus, the direct matrix analysis and the feedback representation
provide sufficient conditions in different ranges of $\beta$.
\JY{The stochastic approximation conditions required to relate the
constrained recursion \eqref{eq:tdrc-sa-form} to the projected mean
ODE are verified in Appendix~\ref{app:tdrc-sa-verification}, following
the weak-convergence method of \citep[Theorem~8.2.3]{kushner2003}
and the organization of \citep[Section~3.1.2]{Yugtd}. Together with
Lemma~\ref{sol-pro-ODE}, this verification completes the proof of the
main convergence result. The proofs of
Theorem~\ref{thm:tdrc-convergence} and
Corollaries~\ref{cor:tdrc-mean-convergence} and
\ref{cor:tdrc-positive-stable-large-beta} are given in
Appendix~\ref{app:tdrc-main-proof}.}

\paragraph{Remark (correction to the original one-step TDRC condition).}
\JY{For $\lambda=0$, the bound on $\beta$ in Lemma~\ref{lem:tdrc-hurwitz-condition} corrects
the corresponding condition in Theorem~3.1 of
\citep{ghiassian2020gradient}. Conditions C1--C3 in their
Appendix H.3.1 are valid, but the derivation from Eqs.~(18) to (20) leads to the incorrect bound $\beta<-\lambda_{\max}(H_0^{-1}A_0A_0^\top)$.
Lemma~\ref{lem:tdrc-hurwitz-condition} corrects this bound: there is no upper bound on $\beta$ when $H_0\succeq0$, whereas otherwise the corrected bound is $\beta<
\frac{1}{
\lambda_{\max}\!\left(
-(A_0A_0^\top)^{-1}H_0
\right)}$.
The corresponding proof is given in Appendix~\ref{app:tdrc-hurwitz-proof}.}

\section{Convergence Analysis of
\texorpdfstring{TDRC($\lambda$)-Huber}{TDRC(lambda)-Huber}}
\label{sec:huber-convergence}

This section establishes convergence of constrained
TDRC($\lambda$)-Huber with diminishing stepsizes. We first state the
main convergence results and computable conditions for general
eligibility traces and the one-step case. These computable conditions identify a large-$\beta$ convergence mechanism driven by regularization. We then derive the
corresponding nonlinear mean ODEs and establish their stability using
an integral small-gain argument. The supporting stability results,
derivations of the computable conditions, stochastic approximation
verifications, and proofs are given in the appendices.

\subsection{Main Convergence Results}
\label{sec:huber-main-convergence-results}

The case $\lambda_n\equiv0$ gives the one-step TDRC-Huber
algorithm, while TDRC($\lambda$)-Huber extends it to general
eligibility traces. We first give the convergence result for
TDRC($\lambda$)-Huber under an integral small-gain condition and
then derive a computable sufficient condition. We finally specialize
the analysis to the one-step case, for which specialized conditions are
available.

We use the joint parameter $u$, the joint iterate $U_n$, the
constraint set $B(R)$, and the Euclidean projection $\Pi_{B(R)}$
defined in Section~\ref{subsec:tdrc-main-convergence}. The constrained
TDRC($\lambda$)-Huber update is
\begin{align*}
U_{n+1}
=
\Pi_{B(R)}
\begin{pmatrix}
x_n+\eta\alpha_n
\left[
e_n\delta_n(\theta_n)
-\phi(S_n)\phi(S_n)^\top x_n
-\beta x_n
\right]
\\[2mm]
\theta_n+\alpha_n
\left[
e_n\delta_n(\theta_n)
-\rho_n(1-\lambda_{n+1})\gamma_{n+1}
\phi(S_{n+1})
\operatorname{clip}_\tau(e_n^\top x_n)
\right]
\end{pmatrix}.
\end{align*}
Compared with the constrained TDRC($\lambda$) update in
\eqref{eq:constrained-tdrc-lambda}, the auxiliary update is unchanged,
while the correction term in the primary update is clipped. As in
Section~\ref{subsec:tdrc-main-convergence}, the projection keeps the
iterates bounded, and the radius $R$ is chosen sufficiently large.

\paragraph{General eligibility traces.}

Define
\begin{equation*}
M_{\beta,\lambda}
:=
\begin{pmatrix}
-\eta(C+\beta I)&-\eta A_\lambda\\
0&-A_\lambda
\end{pmatrix},
\qquad
B_\lambda
:=
\begin{pmatrix}
0\\
-I
\end{pmatrix},
\qquad
\mathcal C_\lambda
:=
\begin{pmatrix}
I&0
\end{pmatrix}.
\end{equation*}
When $M_{\beta,\lambda}$ is Hurwitz, define the integral small gain
\begin{equation*}
\Gamma_2
\left(
M_{\beta,\lambda},
B_\lambda,
\mathcal C_\lambda
\right)
:=
\int_0^\infty
\left\|
\mathcal C_\lambda
e^{M_{\beta,\lambda}t}
B_\lambda
\right\|_2
\,dt.
\end{equation*}
This gain is used below to formulate an integral small-gain
condition that controls the effect of the nonlinear clipping term. Under the stated
condition, the target fixed point of the expected recursion,
equivalently the equilibrium of the mean ODE, is unique and globally
asymptotically stable. This is the main ODE stability property
required for the convergence theorem.

Also, define
\begin{equation*}
L_\lambda
:=
\mathbb E_\zeta
\left[
\rho_0(1-\lambda_1)\gamma_1
\|\phi(S_1)\|_2
\|e_0\|_2
\right].
\end{equation*}
The quantity $L_\lambda$ provides a Lipschitz bound for the
nonlinear mean correction, thereby bounding the magnitude of its
change in response to changes in $x$. Its properties are established
in Proposition~\ref{prop:huber-mean-field}.

\begin{theorem}[Convergence of constrained TDRC($\lambda$)-Huber]
\label{thm:huber-general-convergence}
Suppose that Assumption~\ref{assump:lambda-conditions} holds and that
$\{\alpha_n\}$ is a sequence of diminishing stepsizes satisfying
Assumption~\ref{assump:stepsize}. Suppose also that $A_\lambda$ is
positive stable, $\beta\geq0$, and $\eta>0$. Suppose that at least
one of the following conditions holds:
\begin{enumerate}
\item[\textnormal{(i)}]
The integral gain satisfies
\begin{equation}
L_\lambda
\Gamma_2
\left(
M_{\beta,\lambda},
B_\lambda,
\mathcal C_\lambda
\right)
<1.
\label{eq:huber-main-general-gain-condition}
\end{equation}

\item[\textnormal{(ii)}]
Let $P_{A,\lambda}\succ0$ be the unique solution of
\begin{equation}
A_\lambda^\top P_{A,\lambda}
+
P_{A,\lambda}A_\lambda
=
I,
\label{eq:huber-main-Alambda-lyapunov-equation}
\end{equation}
and define
\begin{equation*}
\kappa_2(P_{A,\lambda})
:=
\frac{
\lambda_{\max}(P_{A,\lambda})
}{
\lambda_{\min}(P_{A,\lambda})
}.
\end{equation*}
Also, let
\begin{equation*}
\overline L_\lambda
:=
\gamma_{\max}\phi_{\max}^2
\,\xi^\top(I-P\Gamma)^{-1}\mathbf 1,
\end{equation*}
which is the computable upper bound on $L_\lambda$ given in
Appendix~\ref{app:computable-Llambda-bound}. The parameter $\beta$
satisfies
\begin{equation}
\beta
>
2\overline L_\lambda
\lambda_{\max}(P_{A,\lambda})
\sqrt{\kappa_2(P_{A,\lambda})}
\|A_\lambda\|_2
-
\lambda_{\min}(C).
\label{com_con_1}
\end{equation}
\end{enumerate}

Define
\begin{equation*}
\theta_\lambda^*
:=
A_\lambda^{-1}b_\lambda,
\qquad
u_\lambda^*
:=
\begin{pmatrix}
0\\
\theta_\lambda^*
\end{pmatrix}.
\end{equation*}

Choose $R>R_{\lambda,\tau}$, where $R_{\lambda,\tau}$ is defined in
Proposition~\ref{prop:huber-general-trace-stability}, and let
$\{U_n\}_{n\geq0}$ be generated by the constrained
TDRC($\lambda$)-Huber recursion. For every initial condition of the
state--trace process, including every initial trace $e_0$, and every
$U_0\in B(R)$, there exists
a sequence $T_k\to\infty$ such that, for every $\delta>0$,
\begin{equation}
\limsup_{k\to\infty}
\mathbb P
\left(
U_i\notin N_\delta(u_\lambda^*)
\text{ for some }
i\in[k,m(k,T_k)]
\right)
=
0.
\label{eq:huber-general-window-convergence}
\end{equation}
Consequently,
\begin{equation}
U_n
\longrightarrow
u_\lambda^*
\qquad
\text{in probability}.
\label{eq:huber-general-convergence-probability}
\end{equation}
Moreover, since the iterates remain in the bounded set $B(R)$,
\begin{equation}
\lim_{n\to\infty}
\mathbb E
\left[
\|U_n-u_\lambda^*\|_2
\right]
=
0.
\label{eq:huber-general-mean-convergence}
\end{equation}
\end{theorem}

Condition~\textnormal{(i)} can be checked by evaluating the integral
gain numerically. Condition~\textnormal{(ii)} avoids this integration:
it can be checked by solving the Lyapunov equation
\eqref{eq:huber-main-Alambda-lyapunov-equation} and evaluating the
resulting explicit lower bound on $\beta$. The derivation showing that
condition~\textnormal{(ii)} implies condition~\textnormal{(i)} is
given in Appendix~\ref{app:huber-computable-conditions}.

\paragraph{The one-step case.}

We next specialize the analysis to $\lambda_n\equiv0$. Define
\begin{equation*}
M_{\beta,0}
:=
\begin{pmatrix}
-\eta(C+\beta I)&-\eta A_0\\
0&-A_0
\end{pmatrix},
\qquad
B_0
:=
\begin{pmatrix}
0\\
-D_0
\end{pmatrix},
\qquad
\mathcal C_0
:=
\begin{pmatrix}
\Phi&0
\end{pmatrix}.
\end{equation*}
The following result gives three sufficient conditions for
convergence. The first condition uses the integral gain directly.
The second bounds this gain through a Lyapunov equation, while the
third gives a more direct condition that avoids solving such an
equation.

\begin{corollary}[Convergence conditions for the one-step case]
\label{cor:huber-zero-convergence}
Suppose that Assumption~\ref{assump:lambda-conditions} holds. Let
$\{\alpha_n\}$ be a sequence of diminishing stepsizes satisfying
Assumption~\ref{assump:stepsize}. Suppose also that $A_0$ is positive
stable, $\beta\geq0$, and $\eta>0$. Define
\begin{equation*}
\theta_0^*
:=
A_0^{-1}b_0,
\qquad
u_0^*
:=
\begin{pmatrix}
0\\
\theta_0^*
\end{pmatrix}.
\end{equation*}

Suppose that at least one of the following conditions holds:
\begin{enumerate}
\item[\textnormal{(i)}]
The integral gain satisfies
\begin{equation}
\Gamma_2
\left(
M_{\beta,0},
B_0,
\mathcal C_0
\right)
<1.
\label{eq:huber-main-zero-gain-condition}
\end{equation}

\item[\textnormal{(ii)}]
Let $P_A\succ0$ be the unique solution of
\begin{equation}
A_0^\top P_A+P_AA_0=I,
\label{eq:huber-main-A0-lyapunov-equation}
\end{equation}
and define
\begin{equation*}
\kappa_2(P_A)
:=
\frac{
\lambda_{\max}(P_A)
}{
\lambda_{\min}(P_A)
}.
\end{equation*}
The parameter $\beta$ satisfies
\begin{equation}
\beta
>
2\lambda_{\max}(P_A)
\sqrt{\kappa_2(P_A)}
\|\Phi\|_2
\|A_0\|_2
\|D_0\|_2
-
\lambda_{\min}(C).
\label{com_con_2}
\end{equation}

\item[\textnormal{(iii)}]
The symmetric part
\begin{equation*}
H_0
:=
\frac{A_0+A_0^\top}{2}
\end{equation*}
is positive definite, and
\begin{equation}
\beta
>
\frac{
\|\Phi\|_2
\|A_0\|_2
\|D_0\|_2
}{
\lambda_{\min}(H_0)
}
-
\lambda_{\min}(C).
\label{com_con_3}
\end{equation}
\end{enumerate}

Choose $R>R_{0,\tau}$ and let $\{U_n\}_{n\geq0}$ be generated by the
constrained TDRC(0)-Huber recursion with $U_0\in B(R)$. Then all
conclusions of Theorem~\ref{thm:huber-general-convergence} hold with
$u_\lambda^*$ replaced by $u_0^*$.
\end{corollary}

Condition \textnormal{(i)}
can be evaluated numerically. Condition \textnormal{(ii)} gives an
explicit sufficient lower bound on $\beta$ by bounding the integral
gain through the solution of the Lyapunov equation
\eqref{eq:huber-main-A0-lyapunov-equation}. Although this condition is
also numerically computable, it requires solving an auxiliary
Lyapunov equation. Condition \textnormal{(iii)} avoids this step: all
quantities in the resulting bound can be computed directly from the
problem matrices. Both conditions
\textnormal{(ii)} and \textnormal{(iii)} imply condition
\textnormal{(i)} and hence guarantee convergence. The two explicit
conditions are compared numerically in
Section~\ref{subsubsec:four-state-certificates}. The derivation showing that conditions~\textnormal{(ii)} and \textnormal{(iii)} imply condition~\textnormal{(i)} is
given in Appendix~\ref{app:huber-computable-conditions}.

We refer to condition~\textnormal{(ii)} of
Theorem~\ref{thm:huber-general-convergence} for general eligibility
traces and conditions~\textnormal{(ii)} and \textnormal{(iii)} of
Corollary~\ref{cor:huber-zero-convergence} for the one-step case
collectively as the \emph{computable conditions}. These conditions
provide explicit lower bounds on $\beta$ while imposing no restriction
on $\eta$ beyond $\eta>0$. Thus, for any chosen $\eta>0$, convergence
is guaranteed by taking $\beta$ sufficiently large, making parameter
selection more flexible in practice.

\subsection{Mean ODE and Feedback Representations}
\label{sec:huber-associated-mean-odes}

We first introduce the nonlinear mean field generated by the Huber
correction term. We then derive feedback representations of the mean
ODE for general eligibility traces and for the one-step case.

\subsubsection{State-Dependent and History-Dependent Choices of
\texorpdfstring{$\lambda$}{lambda}}

\begin{proposition}
\label{prop:huber-mean-field}
Under Assumption~\ref{assump:lambda-conditions},
define
\begin{equation*}
\mathcal H_{\lambda,\tau}(x)
:=
\mathbb E_\zeta\!\left[
\rho_0(1-\lambda_1)\gamma_1
\phi(S_1)
\operatorname{clip}_\tau(e_0^\top x)
\right],
\qquad x\in\mathbb R^d.
\end{equation*}
Then $\mathcal H_{\lambda,\tau}(0)=0$.
Moreover, $\mathcal H_{\lambda,\tau}$ is globally Lipschitz:
\begin{equation*}
\left\|
\mathcal H_{\lambda,\tau}(x)
-\mathcal H_{\lambda,\tau}(y)
\right\|_2
\leq
L_\lambda\|x-y\|_2,
\qquad x,y\in\mathbb R^d,
\end{equation*}
where
\begin{equation*}
L_\lambda
:=
\mathbb E_\zeta\!\left[
\rho_0(1-\lambda_1)\gamma_1
\|\phi(S_1)\|_2\|e_0\|_2
\right]
<\infty.
\end{equation*}
Consequently,
\begin{equation}
\|\mathcal H_{\lambda,\tau}(x)\|_2
\leq
L_\lambda\|x\|_2,
\qquad x\in\mathbb R^d.
\label{eq:huber-mean-field-linear-growth}
\end{equation}
The mean field is also uniformly bounded:
\begin{align}
\|\mathcal H_{\lambda,\tau}(x)\|_2
&\leq
\tau\,
\mathbb E_\zeta\!\left[
\rho_0(1-\lambda_1)\gamma_1
\|\phi(S_1)\|_2
\right]
\notag\\
&=:
\overline H_{\lambda,\tau}
<\infty.
\label{eq:huber-general-H-uniform-bound}
\end{align}
\end{proposition}
The proof is given in
Appendix~\ref{app:huber-mean-field-proofs}.

Recall that
\begin{equation*}
C
:=
\Phi^\top\Xi\Phi,
\qquad
A_\lambda
:=
\Phi^\top\Xi(I-P^{(\lambda)})\Phi,
\qquad
b_\lambda
:=
\Phi^\top\Xi r_\pi^{(\lambda)}.
\end{equation*}

We derive the mean ODE following the same steps as in
Section~\ref{subsec:tdrc-main-convergence}. For state-dependent
$\lambda$, define
\begin{equation*}
\begin{aligned}
g_\tau(\theta,x,z)
&=
e\,\bar\delta(s,s',v_\theta)
-\rho(s,s')
\left(1-\lambda(s')\right)
\gamma(s')\phi(s')
\operatorname{clip}_\tau(e^\top x).
\end{aligned}
\end{equation*}
For history-dependent $\lambda$, define
\begin{equation*}
\begin{aligned}
g_\tau(\theta,x,z)
&=
e\,\bar\delta(s,s',v_\theta)
-\rho(s,s')
\left(1-\lambda(f(y,s'),e)\right)
\gamma(s')\phi(s')
\operatorname{clip}_\tau(e^\top x),
\end{aligned}
\end{equation*}
where $f$ is the function in
\eqref{eq:history-dependent-lambda} that determines the next memory
state.

Let
\[
\omega_{n+1}
=
\rho_n
\left(
R_{n+1}-r(S_n,S_{n+1})
\right),
\]
which has zero mean given the history. The constrained
update can then be written as
\begin{align*}
U_{n+1}
=
\Pi_{B(R)}
\begin{pmatrix}
x_n+\alpha_n
\left[
h(\theta_n,x_n,Z_n)
+\eta e_n\omega_{n+1}
\right]
\\[2mm]
\theta_n+\alpha_n
\left[
g_\tau(\theta_n,x_n,Z_n)
+e_n\omega_{n+1}
\right]
\end{pmatrix}.
\end{align*}

Let
\[
u
=
\begin{pmatrix}
x\\
\theta
\end{pmatrix},
\qquad
F_\tau(u,z)
=
\begin{pmatrix}
h(\theta,x,z)\\
g_\tau(\theta,x,z)
\end{pmatrix},
\qquad
\varepsilon_{n+1}
=
\begin{pmatrix}
\eta e_n\omega_{n+1}\\
e_n\omega_{n+1}
\end{pmatrix}.
\]
Then the update has the compact form
\begin{equation}
U_{n+1}
=
\Pi_{B(R)}
\left[
U_n+\alpha_n
\left(
F_\tau(U_n,Z_n)+\varepsilon_{n+1}
\right)
\right].
\label{eq:tdrchuber-sa-form}
\end{equation}

The secondary update is unchanged from TDRC($\lambda$). Hence, its
mean field is
\begin{align*}
\bar h(\theta,x)
&=
\eta
\left[
b_\lambda-A_\lambda\theta-(C+\beta I)x
\right],
\end{align*}
which coincides with \eqref{eq:tdrc-mean-h}. For the primary update,
define
\begin{align*}
\bar g_\tau(\theta,x)
&:=
\mathbb E_\zeta
\left[
g_\tau(\theta,x,Z_0)
\right]
\notag\\
&=
b_\lambda-A_\lambda\theta
-
\mathbb E_\zeta\!\left[
\rho_0(1-\lambda_1)\gamma_1
\phi(S_1)
\operatorname{clip}_\tau(e_0^\top x)
\right]
\notag\\
&=
b_\lambda-A_\lambda\theta
-\mathcal H_{\lambda,\tau}(x).
\end{align*}
Therefore, the unprojected mean ODE is
\begin{align*}
\dot x(t)
&=
-\eta(C+\beta I)x(t)
-\eta A_\lambda\theta(t)
+\eta b_\lambda,
\\
\dot\theta(t)
&=
-A_\lambda\theta(t)
+b_\lambda
-\mathcal H_{\lambda,\tau}(x(t)).
\end{align*}

Under the nonsingularity assumption in
Section~\ref{sec:tdrc-lambda}, $A_\lambda$ is nonsingular. Define
\begin{equation*}
\theta_\lambda^*
:=
A_\lambda^{-1}b_\lambda,
\qquad
u_\lambda^*
:=
\begin{pmatrix}
0\\
\theta_\lambda^*
\end{pmatrix}.
\end{equation*}

Let
\begin{equation*}
M_{\beta,\lambda}
:=
\begin{pmatrix}
-\eta(C+\beta I)&-\eta A_\lambda\\
0&-A_\lambda
\end{pmatrix},
\qquad
B_\lambda
:=
\begin{pmatrix}
0\\
-I
\end{pmatrix},
\qquad
\mathcal C_\lambda
:=
\begin{pmatrix}
I&0
\end{pmatrix},
\end{equation*}
and
\begin{equation*}
q_\lambda
:=
\begin{pmatrix}
\eta b_\lambda\\
b_\lambda
\end{pmatrix}.
\end{equation*}
Then
\begin{equation*}
q_\lambda=-M_{\beta,\lambda}u_\lambda^*,
\qquad
\mathcal C_\lambda u_\lambda^*=0.
\end{equation*}
Writing $u=(x^\top,\theta^\top)^\top$, the mean ODE can therefore
be expressed directly in the original coordinates as
\begin{align}
\dot u(t)
&=
M_{\beta,\lambda}u(t)
+q_\lambda
+B_\lambda
\mathcal H_{\lambda,\tau}
\left(\mathcal C_\lambda u(t)\right)
\notag\\
&=
M_{\beta,\lambda}\left(u(t)-u_\lambda^*\right)
+B_\lambda
\mathcal H_{\lambda,\tau}
\left(
\mathcal C_\lambda\left(u(t)-u_\lambda^*\right)
\right).
\label{eq:huber-general-feedback-ode}
\end{align}

\subsubsection{The One-Step Case}

In the one-step case, $e_n=\phi(S_n)$, and
$P^{(0)}=P\Gamma$. The nonlinear mean field then admits the
following explicit representation.

\begin{corollary}[The one-step mean field]
\label{cor:huber-one-step-mean-field}
Suppose that the conditions of
Proposition~\ref{prop:huber-mean-field} hold and that
$\lambda_n\equiv0$. Then
\begin{equation}
\mathcal H_{0,\tau}(x)
=
D_0
\Pi_{[-\tau,\tau]^{|\mathcal S|}}(\Phi x),
\qquad x\in\mathbb R^d,
\label{eq:huber-mean-field-lambda-zero}
\end{equation}
where
\begin{equation*}
D_0
:=
\Phi^\top(P\Gamma)^\top\Xi
=
\Phi^\top(P^{(0)})^\top\Xi.
\end{equation*}
Equivalently,
\begin{equation*}
\left[
\Pi_{[-\tau,\tau]^{|\mathcal S|}}(\Phi x)
\right]_s
=
\operatorname{clip}_\tau(\phi(s)^\top x),
\qquad s\in\mathcal S.
\end{equation*}
Moreover, $\mathcal H_{0,\tau}$ is globally Lipschitz:
\begin{equation}
\left\|
\Pi_{[-\tau,\tau]^{|\mathcal S|}}(v)
-
\Pi_{[-\tau,\tau]^{|\mathcal S|}}(w)
\right\|_2
\leq
\|v-w\|_2,
\qquad
v,w\in\mathbb R^{|\mathcal S|}.
\label{eq:proj-zero-explicit-lipschitz}
\end{equation}
\end{corollary}
The proof is given in
Appendix~\ref{app:huber-mean-field-proofs}.
When $\lambda_n\equiv0$, the one-step mean ODE is
\begin{align*}
\dot x(t)
&=
-\eta(C+\beta I)x(t)
-\eta A_0\theta(t)
+\eta b_0,
\\
\dot\theta(t)
&=
-A_0\theta(t)+b_0
-D_0\Pi_{[-\tau,\tau]^{|\mathcal S|}}(\Phi x(t)).
\end{align*}
Under the assumptions in Section~\ref{sec:tdrc-lambda}, $A_0$ is
nonsingular. Define
\begin{equation*}
\theta_0^*
:=
A_0^{-1}b_0,
\qquad
u_0^*
:=
\begin{pmatrix}
0\\
\theta_0^*
\end{pmatrix}.
\end{equation*}
Let
\begin{equation*}
M_{\beta,0}
:=
\begin{pmatrix}
-\eta(C+\beta I)&-\eta A_0\\
0&-A_0
\end{pmatrix},
\qquad
B_0
:=
\begin{pmatrix}
0\\
-D_0
\end{pmatrix},
\qquad
\mathcal C_0
:=
\begin{pmatrix}
\Phi&0
\end{pmatrix},
\end{equation*}
and
\begin{equation*}
q_0
:=
\begin{pmatrix}
\eta b_0\\
b_0
\end{pmatrix}.
\end{equation*}
Then
\begin{equation*}
q_0=-M_{\beta,0}u_0^*,
\qquad
\mathcal C_0u_0^*=0,
\end{equation*}
and the one-step mean ODE can be written as
\begin{align}
\dot u(t)
&=
M_{\beta,0}u(t)
+q_0
+B_0
\Pi_{[-\tau,\tau]^{|\mathcal S|}}
\left(\mathcal C_0u(t)\right)
\notag\\
&=
M_{\beta,0}\left(u(t)-u_0^*\right)
+B_0
\Pi_{[-\tau,\tau]^{|\mathcal S|}}
\left(
\mathcal C_0\left(u(t)-u_0^*\right)
\right).
\label{eq:huber-zero-feedback-ode}
\end{align}

Hence, in both the general and one-step cases, the mean ODE can be
written in the form
\begin{equation*}
\dot u(t)
=
M\left(u(t)-u^*\right)
+
BF\left(
\mathcal C\left(u(t)-u^*\right)
\right).
\end{equation*}

\subsection{Stability of the Mean ODEs}
\label{Stability_Criteria}

The feedback representations derived in
Section~\ref{sec:huber-associated-mean-odes} show that the general
eligibility-trace and one-step mean ODEs can both be written in the
common form
\[
\dot u(t)
=
M\bigl(u(t)-u^*\bigr)
+
BF\bigl(C(u(t)-u^*)\bigr).
\]
This common structure allows the same small-gain argument to be used
in both cases. We first establish general stability criteria for a
system of this form and its projected version, and then apply these
results to the general eligibility-trace and one-step
TDRC($\lambda$)-Huber mean ODEs to establish their stability.

\paragraph{General stability criteria.}

For a Hurwitz matrix $M$, define
\begin{equation*}
\Gamma_2(M,B,\mathcal C)
:=
\int_0^\infty
\|\mathcal C e^{Mt}B\|_2\,dt.
\end{equation*}

\begin{lemma}[Integral stability criterion]
\label{lem:integral-stability-criterion}
Suppose that $M$ is Hurwitz and that
$F:\mathbb R^p\to\mathbb R^q$ satisfies
\begin{equation*}
F(0)=0
\end{equation*}
and
\begin{equation}
\|F(v)-F(w)\|_2
\leq
L_F\|v-w\|_2,
\qquad v,w\in\mathbb R^p,
\label{eq:F-global-lipschitz}
\end{equation}
for some $L_F\geq0$. If
\begin{equation*}
L_F\Gamma_2(M,B,\mathcal C)<1,
\end{equation*}
then $u^*$ is the unique globally asymptotically stable equilibrium
of
\begin{equation}
\dot u(t)
=
M\left(u(t)-u^*\right)
+BF\left(
\mathcal C\left(u(t)-u^*\right)
\right).
\label{eq:abstract-nonlinear-ode}
\end{equation}
\end{lemma}

The proof is given in Appendix~\ref{app:huber-stability-proofs}. It applies
the small-gain argument to the deviation
$\widetilde u=u-u^*$.

We next consider the projected system
\begin{equation}
\dot u
=
M(u-u^*)
+
BF\left(\mathcal C(u-u^*)\right)
+
z,
\qquad
z\in-\mathcal N_{B(R)}(u),
\label{eq:abstract-projected-ode}
\end{equation}
where
\[
B(R)
:=
\left\{
u\in\mathbb R^m:\|u\|_2\leq R
\right\},
\]
and $\mathcal N_{B(R)}(u)$ denotes the normal cone of $B(R)$ at $u$.

\begin{lemma}[Stability under Euclidean projection]
\label{lem:huber-invariant-projection-set}
Suppose that $M$ is Hurwitz, that the unprojected system
\eqref{eq:abstract-nonlinear-ode} has $u^*$ as its unique globally
asymptotically stable equilibrium, and that $F$ is globally
Lipschitz and satisfies
\begin{equation*}
\|F(v)\|_2
\leq
\overline F,
\qquad
v\in\mathbb R^p.
\end{equation*}
Let $Q\succ0$ be the unique solution of
\begin{equation*}
M^\top Q+QM=-I,
\end{equation*}
and define
\begin{equation*}
K_Q
:=
\|QMu^*\|_2
+
\|QB\|_2\overline F
\end{equation*}
and
\begin{equation*}
R_Q
:=
\max\left\{
\|u^*\|_2,
2K_Q
\sqrt{
\frac{\lambda_{\max}(Q)}
     {\lambda_{\min}(Q)}
}
\right\}.
\end{equation*}
If
\[
R>R_Q,
\]
then $u^*\in\operatorname{int}B(R)$ and is the unique globally
asymptotically stable equilibrium of
\eqref{eq:abstract-projected-ode} relative to $B(R)$. Consequently,
\begin{equation*}
L_{B(R)}
=
\{u^*\}.
\end{equation*}
\end{lemma}

The proof is given in
Appendix~\ref{app:huber-projected-ode-proof}. It first shows that every
projected solution enters a positively invariant set contained in
$\operatorname{int}B(R)$. The projection term then vanishes, and the
conclusion of Lemma~\ref{lem:integral-stability-criterion} gives
convergence to $u^*$.

\paragraph{Application to the mean ODEs.}
Having established Lemmas~\ref{lem:integral-stability-criterion} and
\ref{lem:huber-invariant-projection-set}, we apply them to obtain
stability results for the projected mean ODEs corresponding to the
general eligibility-trace and one-step feedback representations
\eqref{eq:huber-general-feedback-ode} and
\eqref{eq:huber-zero-feedback-ode}, respectively.

For state-dependent and history-dependent choices of $\lambda$, the
general eligibility-trace case corresponds to
\begin{equation*}
M=M_{\beta,\lambda},
\qquad
B=B_\lambda,
\qquad
\mathcal C=\mathcal C_\lambda,
\qquad
F=\mathcal H_{\lambda,\tau},
\qquad
u^*=u_\lambda^*.
\end{equation*}
Under condition~\eqref{eq:huber-main-general-gain-condition} and
$R>R_{\lambda,\tau}$, the two lemmas imply that $u_\lambda^*$ is the
unique globally asymptotically stable equilibrium of the projected
mean ODE, with $L_{B(R)}=\{u_\lambda^*\}$.

For the one-step case, we instead take
\begin{equation*}
M=M_{\beta,0},
\qquad
B=B_0,
\qquad
\mathcal C=\mathcal C_0,
\qquad
F=\Pi_{[-\tau,\tau]^{|\mathcal S|}},
\qquad
u^*=u_0^*.
\end{equation*}
Under condition~\eqref{eq:huber-main-zero-gain-condition} and
$R>R_{0,\tau}$, the same argument implies that $u_0^*$ is the unique
globally asymptotically stable equilibrium of the projected mean ODE,
with $L_{B(R)}=\{u_0^*\}$.
The complete stability
statements, the definitions of $R_{\lambda,\tau}$ and
$R_{0,\tau}$, and the detailed proofs are given in
Appendix~\ref{app:huber-stability-results}.

Given these stability conclusions,
Appendix~\ref{app:huber-sa-verification} verifies the stochastic
approximation conditions needed to connect the constrained
recursion~\eqref{eq:tdrchuber-sa-form} to the corresponding projected
mean ODEs. Combining the ODE stability results with this stochastic
approximation argument yields the main convergence conclusions, whose
proofs are completed in Appendix~\ref{app:huber-main-proofs}.

\section{The Role of the Regularization Parameter
\texorpdfstring{$\beta$}{beta}}
\label{sec:role-of-beta}
This section discusses the role of the regularization parameter
$\beta$ in TDRC($\lambda$) and TDRC($\lambda$)-Huber. We first
examine its effect on the algorithms, their target equilibrium, and
their convergence conditions. In particular, we explain how a
different feedback representation reveals an additional
large-$\beta$ convergence mechanism driven by regularization. This
mechanism applies to both linear TDRC($\lambda$) and TDRC($\lambda$)-Huber when $A_\lambda$ is positive stable, including cases in
which $H_\lambda\not\succeq0$. We then derive a necessary condition
for the small-gain condition and examine how $\beta$ and $\eta$ enter
this condition and the computable sufficient conditions stated in
Section~\ref{sec:huber-main-convergence-results} and justified in
Appendix~\ref{app:huber-computable-conditions}. We also compare
TDRC($\lambda$)-Huber with TDC-Huber.
Finally, we illustrate these results using a four-state off-policy
prediction problem.


\subsection{Effect of \texorpdfstring{$\beta$}{beta} on the
Algorithm and Convergence}

The parameter $\beta$ controls the $\ell_2$ regularization applied
to the secondary update of TDC. This regularization can reduce the
variance of the auxiliary weights and help them converge more
quickly
\citep{patterson2023robust,ghiassian2020gradient}. When $\beta=0$,
TDRC reduces to TDC, whereas increasing $\beta$ makes the algorithm
behave more like TD. Thus, $\beta$ provides a way to interpolate
between TDC and TD.

Under the assumptions used in our analysis, this regularization does
not change the target equilibrium. For both TDRC($\lambda$) and
TDRC($\lambda$)-Huber, the equilibrium satisfies
\begin{equation*}
A_\lambda\theta_\lambda^*
=
b_\lambda,
\qquad
x^*=0.
\end{equation*}
The regularization term $-\beta x$ therefore vanishes at the
equilibrium. Hence, $\beta$ affects the transient dynamics and the
convergence conditions, but not the limiting parameter
$\theta_\lambda^*$.

For linear
TDRC($\lambda$), recall that
\[
H_\lambda
:=
\frac{A_\lambda+A_\lambda^\top}{2}.
\]
If $H_\lambda\succeq0$, the sufficient condition in
Lemma~\ref{lem:tdrc-hurwitz-condition} applies for every
$\beta\geq0$ and $\eta>0$. If $H_\lambda\not\succeq0$, that
condition requires $\beta$ to remain below a finite upper bound and
$\eta$ to exceed a lower bound. This restriction belongs to that
particular sufficient condition, rather than to linear
TDRC($\lambda$) itself.

The feedback representation and
Corollary~\ref{cor:tdrc-positive-stable-large-beta} together reveal
an additional large-$\beta$ convergence mechanism driven by
regularization. If $A_\lambda$ is positive stable, then, for every
$\eta>0$, the computable sufficient condition holds for both linear
TDRC($\lambda$) and TDRC($\lambda$)-Huber when $\beta$ is
sufficiently large. Thus,
regularization provides the large-$\beta$ convergence mechanism,
while the clipping induced by the Huber loss reduces sensitivity to large
Bellman errors and preserves
this guarantee.

\subsection{Effect of \texorpdfstring{$\beta$}{beta} on the
Small-Gain Conditions}

This subsection examines how $\beta$ and $\eta$ affect the
small-gain convergence analysis. We consider both a necessary
condition and computable sufficient conditions, and use them to
compare TDRC-Huber with TDC-Huber.

We first derive a \textbf{necessary condition} for the small-gain condition
for general \(\lambda\). Since $M_{\beta,\lambda}$ is Hurwitz,
\begin{align*}
\int_0^\infty
\mathcal C_\lambda
e^{M_{\beta,\lambda}t}
B_\lambda\,dt
&=
-\mathcal C_\lambda
M_{\beta,\lambda}^{-1}
B_\lambda
=
(C+\beta I)^{-1}.
\end{align*}
Therefore, the triangle inequality gives
\begin{align}
\Gamma_2
\left(
M_{\beta,\lambda},
B_\lambda,
\mathcal C_\lambda
\right)
&=
\int_0^\infty
\left\|
\mathcal C_\lambda
e^{M_{\beta,\lambda}t}
B_\lambda
\right\|_2\,dt
\notag\\
&\geq
\left\|
\int_0^\infty
\mathcal C_\lambda
e^{M_{\beta,\lambda}t}
B_\lambda\,dt
\right\|_2
\notag\\
&=
\left\|
(C+\beta I)^{-1}
\right\|_2
\notag\\
&=
\frac{1}{\lambda_{\min}(C)+\beta}.
\label{eq:beta-gain-lower-bound}
\end{align}
Consequently, the small-gain condition
\[
L_\lambda
\Gamma_2
\left(
M_{\beta,\lambda},
B_\lambda,
\mathcal C_\lambda
\right)
<1
\]
can hold only if
\begin{equation}
L_\lambda<\lambda_{\min}(C)+\beta,
\label{eq:beta-necessary-condition}
\end{equation}
or equivalently,
\begin{equation}
\beta>L_\lambda-\lambda_{\min}(C).
\label{eq:beta-necessary-lower-bound}
\end{equation}

This necessary condition shows the different roles of $\beta$ and
$\eta$ in the small-gain analysis. Increasing $\beta$ increases the
right-hand side of \eqref{eq:beta-necessary-condition} and therefore
makes the condition easier to satisfy. In contrast, changing $\eta$
may change the actual value of the integral gain
$\Gamma_2(M_{\beta,\lambda},B_\lambda,\mathcal C_\lambda)$, but it
does not change $L_\lambda$ or the lower bound $\frac{1}{\lambda_{\min}(C)+\beta}$
in \eqref{eq:beta-gain-lower-bound}. Thus, if
$\beta\leq L_\lambda-\lambda_{\min}(C)$, no choice of $\eta$ can make
the small-gain condition hold. However, choosing
$\beta>L_\lambda-\lambda_{\min}(C)$ does not guarantee the
small-gain condition, because the actual integral gain may still be
too large. Therefore, the lower bound on $\beta$ in
\eqref{eq:beta-necessary-lower-bound} is necessary but not
sufficient.

We next consider the computable \textbf{sufficient conditions}. In
the proofs, the integral gain is bounded above by expressions of
the form
\begin{equation*}
\frac{\mathsf K}{\lambda_{\min}(C)+\beta},
\end{equation*}
where $\mathsf K$ does not depend on $\beta$; see
\eqref{eq:huber-general-gain-final-bound},
\eqref{eq:huber-zero-lyapunov-integral-bound}, and
\eqref{eq:huber-zero-direct-integral-bound}. Requiring these upper
bounds to be less than one gives the lower bounds on $\beta$ stated
in condition~\textnormal{(ii)} of
Theorem~\ref{thm:huber-general-convergence} and in
Corollary~\ref{cor:huber-zero-convergence}. Hence, increasing $\beta$
not only relaxes the necessary condition but also makes these
computable sufficient conditions easier to satisfy.

This also clarifies the difference between TDRC-Huber and TDC-Huber.
When $\lambda=0$ and $\beta=0$, TDRC-Huber reduces to the TDC-Huber
algorithm of \citet{patterson2023robust}. For every $\eta>0$, the corresponding upper bounds take the forms
\begin{equation*}
\frac{\mathsf K_0}{\lambda_{\min}(C)+\beta}
\qquad\text{and}\qquad
\frac{\mathsf K_0}{\lambda_{\min}(C)}
\end{equation*}
for TDRC-Huber and TDC-Huber, respectively. The TDC-Huber upper bound
is independent of $\eta$. If this upper bound is greater than
or equal to one, it cannot verify the sufficient small-gain
condition. In contrast, $\beta$ provides an additional degree of
freedom in TDRC-Huber. Increasing $\beta$ reduces its upper bound and
may make it less than one, thereby verifying convergence of
TDRC-Huber even when the corresponding TDC-Huber bound cannot verify
convergence.

Together, these results clarify how $\beta$ and $\eta$ enter the
convergence analysis:
\begin{enumerate}
\item
\textbf{Necessary condition.}
If $\beta$ does not satisfy
\eqref{eq:beta-necessary-lower-bound}, no choice of $\eta$ can make
the small-gain condition hold. This does not imply divergence; it
means that convergence cannot be established through this small-gain
condition.
\item
\textbf{Computable sufficient conditions.}
For every $\eta>0$, increasing $\beta$ reduces the analytic
upper bounds on the integral gain. If one of these upper bounds
becomes less than one, the corresponding sufficient condition holds
and convergence is guaranteed.
\end{enumerate}

\subsection{A Four-State Illustration}
\label{subsubsec:four-state-certificates}

\begin{figure}[t]
    \centering
    \includegraphics[width=0.8\linewidth]
    {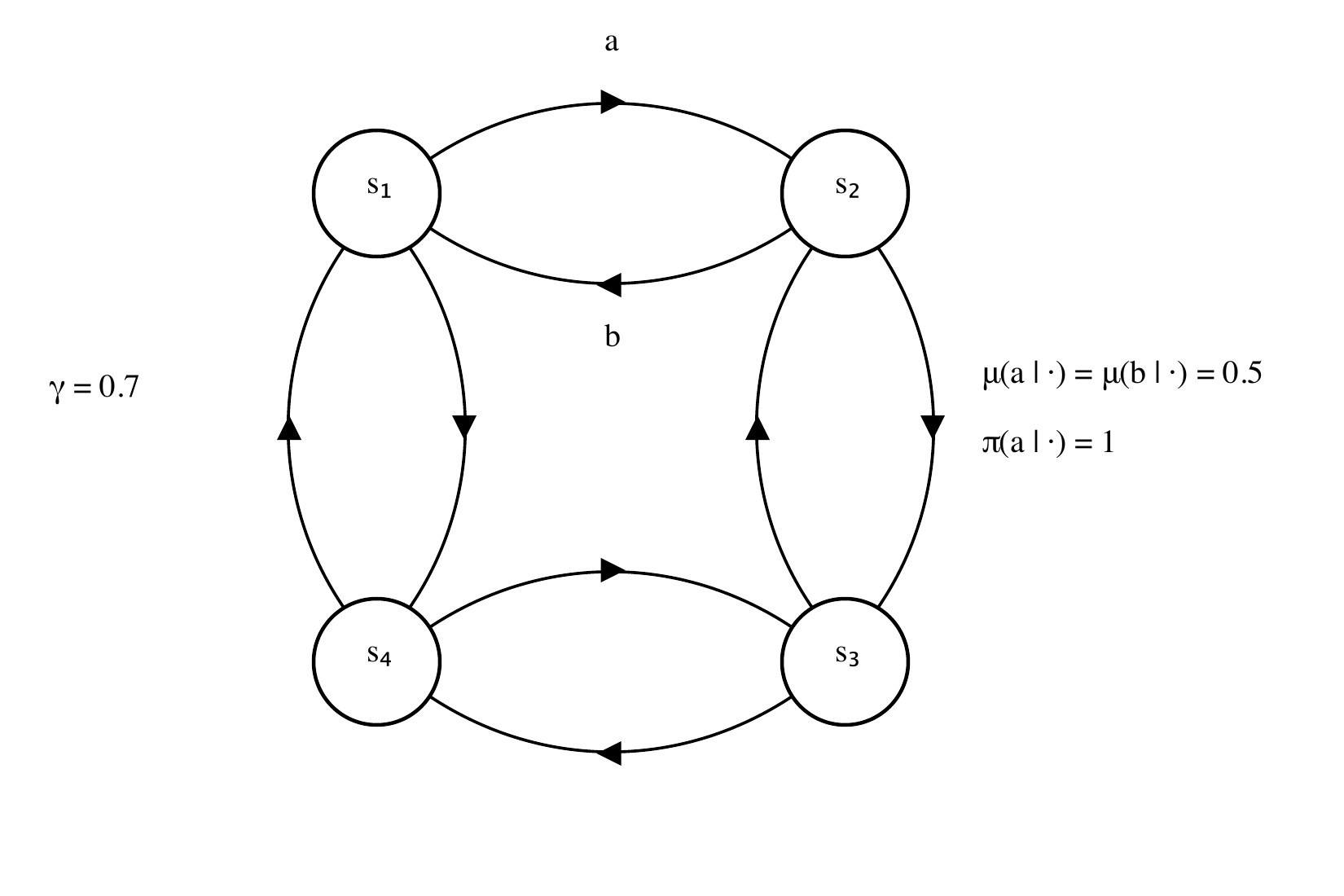}
    \caption{
        Four-state off-policy prediction problem. The target policy
        moves clockwise, whereas the behavior policy chooses the
        clockwise and counterclockwise actions with equal probability.
    }
    \label{fig:four-state-ring-mdp}
\end{figure}

We illustrate these effects using the four-state off-policy
prediction problem shown in
Figure~\ref{fig:four-state-ring-mdp}. The state space is
\[
\mathcal S=\{s_1,s_2,s_3,s_4\}.
\]
Let $a_{\mathrm{cw}}$ and $a_{\mathrm{ccw}}$ denote the actions that
move clockwise and counterclockwise, respectively. The target policy
always selects the clockwise action,
\[
\pi(a_{\mathrm{cw}}\mid s)=1,
\]
whereas the behavior policy $b$ selects the two actions uniformly:
\[
b(a_{\mathrm{cw}}\mid s)
=
b(a_{\mathrm{ccw}}\mid s)
=
\frac12.
\]
If $s_{\mathrm{cw}}$ and $s_{\mathrm{ccw}}$ are the corresponding
successor states, the transition-level importance-sampling ratios are
\[
\rho(s,s_{\mathrm{cw}})=2,
\qquad
\rho(s,s_{\mathrm{ccw}})=0.
\]
The behavior policy has the uniform stationary distribution
\[
\Xi=\frac14 I_4.
\]

Let $P_\pi$ denote the clockwise transition matrix. Set
$\gamma=0.7$ and choose
\[
\Phi
=
\frac1{12}
\begin{pmatrix}
     3 & -2\\
    -3 & -2\\
     3 &  0\\
    -3 &  2
\end{pmatrix},
\qquad
r_\pi
=
\begin{pmatrix}
    7/20\\
    -27/40\\
    1/4\\
    0
\end{pmatrix}.
\]

For $\lambda=0$, direct calculation gives
\[
C
=
\frac1{96}
\begin{pmatrix}
    6 & -1\\
    -1 & 2
\end{pmatrix},
\qquad
A_0
=
\frac1{960}
\begin{pmatrix}
    102 & -17\\
    -17 & 20
\end{pmatrix},
\]
and
\[
D_0
=
\frac7{480}
\begin{pmatrix}
    -3 & 3 & -3 & 3\\
    -2 & 0 & 2 & -2
\end{pmatrix},
\qquad
b_0
=
\begin{pmatrix}
    51/640\\[1mm]
    13/960
\end{pmatrix}.
\]
It follows that
\[
\theta_0^*
=
A_0^{-1}b_0
=
\begin{pmatrix}
1\\[1mm]
3/2
\end{pmatrix},
\qquad
v_{\theta_0^*}
=
\Phi\theta_0^*
=
\begin{pmatrix}
0\\
-1/2\\
1/4\\
0
\end{pmatrix}.
\]
A direct Bellman calculation verifies that
\[
r_\pi=(I-P_\pi\Gamma)v_{\theta_0^*}.
\]
Hence, $v_{\theta_0^*}=v_\pi$, so the target value function is
exactly representable by the chosen features. Moreover, $A_0$ is
symmetric positive definite, and therefore both the direct and
Lyapunov conditions apply.

\subsubsection{The One-Step Conditions}

For the numerical calculations, set $\eta=1$ and define
\begin{equation*}
\Gamma_0^{\mathrm{num}}(\beta)
:=
\Gamma_2(M_{\beta,0},B_0,\mathcal C_0).
\end{equation*}
The direct and Lyapunov upper bounds on this gain are
\begin{align*}
\widehat\Gamma_0^{\mathrm{dir}}(\beta)
&:=
\frac{
\|\Phi\|_2\|A_0\|_2\|D_0\|_2
}{
\lambda_{\min}(H_0)
\bigl(\lambda_{\min}(C)+\beta\bigr)
},
\\
\widehat\Gamma_0^{\mathrm{Lyap}}(\beta)
&:=
\frac{
2\lambda_{\max}(P_A)
\sqrt{\kappa_2(P_A)}
\|\Phi\|_2\|A_0\|_2\|D_0\|_2
}{
\lambda_{\min}(C)+\beta
}.
\end{align*}
By \eqref{eq:huber-zero-direct-integral-bound} and
\eqref{eq:huber-zero-lyapunov-integral-bound},
\begin{equation*}
\Gamma_0^{\mathrm{num}}(\beta)
\leq
\widehat\Gamma_0^{\mathrm{dir}}(\beta),
\qquad
\Gamma_0^{\mathrm{num}}(\beta)
\leq
\widehat\Gamma_0^{\mathrm{Lyap}}(\beta).
\end{equation*}
Hence, requiring either upper bound to be less than one guarantees
the one-step small-gain condition
\eqref{eq:huber-main-zero-gain-condition}. For this example, the
resulting requirements are
\begin{align*}
\widehat\Gamma_0^{\mathrm{dir}}(\beta)<1
&\quad\Longleftrightarrow\quad
\beta
>
\frac{
\|\Phi\|_2\|A_0\|_2\|D_0\|_2
}{
\lambda_{\min}(H_0)
}
-
\lambda_{\min}(C)
\approx 0.270,
\notag
\\
\widehat\Gamma_0^{\mathrm{Lyap}}(\beta)<1
&\quad\Longleftrightarrow\quad
\beta
>
2\lambda_{\max}(P_A)
\sqrt{\kappa_2(P_A)}
\|\Phi\|_2\|A_0\|_2\|D_0\|_2
-
\lambda_{\min}(C)
\approx 0.708.
\notag
\end{align*}
These are the lower bounds on $\beta$ from conditions
\textnormal{(iii)} and \textnormal{(ii)}, respectively, of
Corollary~\ref{cor:huber-zero-convergence}, evaluated for this
example.

\begin{table}[H]
    \centering
    \caption{
        Numerically evaluated integral gain and its analytic upper
        bounds for $\lambda=0$. A value below one verifies the
        corresponding small-gain condition.
    }
    \label{tab:four-state-lambda-zero-certificates}
    \begin{tabular}{c@{\qquad}ccc}
        \toprule
        $\beta$
        & $\Gamma_0^{\mathrm{num}}(\beta)$
        & $\widehat\Gamma_0^{\mathrm{dir}}(\beta)$
        & $\widehat\Gamma_0^{\mathrm{Lyap}}(\beta)$\\
        \midrule
        $0$   & $1.181$ & $15.696$ & $39.530$\\
        $1/2$ & $0.092$ & $0.556$  & $1.401$\\
        $1$   & $0.049$ & $0.283$  & $0.713$\\
        \bottomrule
    \end{tabular}
\end{table}

At $\beta=0$, the numerical gain exceeds one, and neither analytic
bound verifies the small-gain condition. At $\beta=1/2$, the
numerical gain and the direct bound are below one, whereas the
Lyapunov bound remains inconclusive. At $\beta=1$, the two analytic
bounds are below one and therefore verify the small-gain condition.
These results show that the value of $\beta$ required to verify
convergence depends on the sufficient condition being used. The
directly evaluated gain requires the smallest value of $\beta$,
followed by the direct analytic bound and then the more conservative
Lyapunov bound.

The two bounds trade sharpness for generality. The direct bound
requires
\[
H_0=\frac{A_0+A_0^\top}{2}\succ0,
\]
equivalently, $x^\top A_0x>0$ for every $x\neq0$. It is less
conservative in this example. The Lyapunov bound requires only that
$A_0$ be positive stable. Since $H_0\succ0$ implies positive
stability of $A_0$, but the converse need not hold, the Lyapunov
condition applies more broadly, although its bound can be more
conservative.

\subsubsection{The General-Trace Condition}

We next evaluate the sufficient small-gain condition for several
constant choices $\lambda(s)\equiv\lambda$. The stationary trace
distribution gives
\[
(L_0,L_{1/4},L_{1/2},L_{5/8})
=
(0.057888,0.038696,0.024293,0.018390).
\]

These values were obtained by deterministic summation from the
stationary trace distribution, rather than by simulation. Let $N$
denote the number of consecutive clockwise transitions immediately
preceding the current state. Conditional on $S_0=s$,
\[
\mathbb P(N=k\mid S_0=s)
=
2^{-(k+1)},
\qquad
k\geq0,
\]
and
\[
e_0
=
\sum_{j=0}^{k}
(2\lambda\gamma)^j\phi(s_{-j}),
\]
where $s_{-j}$ is the $j$th predecessor of $s$ under the clockwise
target dynamics. Therefore,
\[
L_\lambda
=
\frac{(1-\lambda)\gamma}{4}
\sum_{s\in\mathcal S}
\|\phi(s_{\mathrm{cw}})\|_2
\sum_{k=0}^{\infty}
2^{-(k+1)}
\left\|
\sum_{j=0}^{k}
(2\lambda\gamma)^j\phi(s_{-j})
\right\|_2.
\]
The series was truncated when the first omitted geometric weight was
at most $10^{-16}$. For the values of $\lambda$ considered here, the
resulting truncation error is less than $4\times10^{-17}$. In
particular,
\[
L_{5/8}-\lambda_{\min}(C)
\approx
1.6202\times10^{-5}>0,
\]
so the sign of this difference is not affected by numerical
truncation.
For each pair $(\lambda,\beta)$, define the small-gain quantity
\begin{equation}
G_\lambda^{\mathrm{num}}(\beta)
:=
L_\lambda
\Gamma_2
\left(
M_{\beta,\lambda},
B_\lambda,
\mathcal C_\lambda
\right).
\label{eq:four-state-general-numerical-product}
\end{equation}
The values with $\eta=1$ are reported in
Table~\ref{tab:four-state-general-numerical-product}.

\begin{table}[H]
    \centering
    \caption{
        Direct numerical evaluation of
        $G_\lambda^{\mathrm{num}}(\beta)$ with $\eta=1$. A value
        below one verifies the sufficient small-gain condition.
    }
    \label{tab:four-state-general-numerical-product}
    \begin{tabular}{c@{\qquad}cccc}
        \toprule
        $\lambda$
        & $\beta=0$
        & $\beta=0.02$
        & $\beta=0.05$
        & $\beta=0.1$\\
        \midrule
        $0$   & $3.465$ & $1.781$ & $1.072$ & $0.662$\\
        $1/4$ & $2.281$ & $1.162$ & $0.694$ & $0.427$\\
        $1/2$ & $1.417$ & $0.717$ & $0.426$ & $0.261$\\
        $5/8$ & $1.069$ & $0.539$ & $0.320$ & $0.196$\\
        \bottomrule
    \end{tabular}
\end{table}

At $\beta=0$, all four values are greater than one. Thus, the direct
evaluation with $\eta=1$ does not verify the small-gain condition. To
determine whether changing $\eta$ can make the condition hold, we use
the necessary condition
\eqref{eq:beta-necessary-lower-bound}.

For this example,
\[
\lambda_{\min}(C)
=
\frac{4-\sqrt{5}}{96}
\approx 0.018374.
\]
The necessary lower bounds on $\beta$ are therefore
\begin{align*}
L_0-\lambda_{\min}(C)
&\approx 0.039514,\\
L_{1/4}-\lambda_{\min}(C)
&\approx 0.020322,\\
L_{1/2}-\lambda_{\min}(C)
&\approx 0.005919,\\
L_{5/8}-\lambda_{\min}(C)
&\approx 0.000016.
\end{align*}
These lower bounds are all positive. Hence, when $\beta=0$, no choice
of $\eta$ can make the small-gain condition hold for any of the four
values of $\lambda$. In particular, when $\lambda=0$ and $\beta=0$,
this conclusion applies to the general-trace small-gain condition for
TDC-Huber. It does not address the separate one-step condition in
Corollary~\ref{cor:huber-zero-convergence}. This does not imply
divergence; it only shows that convergence cannot be verified through
the general-trace small-gain condition.

At $\beta=0.02$, the necessary condition is violated for
$\lambda=0$ and $\lambda=1/4$. Thus, no choice of $\eta$ can make the
small-gain condition hold in these two cases. For $\lambda=1/2$ and
$\lambda=5/8$, the necessary condition holds, and the corresponding
values of $G_\lambda^{\mathrm{num}}(\beta)$ with $\eta=1$ are below
one. Hence, the sufficient small-gain condition is verified for these
two values of $\lambda$.

At $\beta=0.05$, the necessary condition holds for all four values of
$\lambda$. However, with $\eta=1$,
$G_0^{\mathrm{num}}(0.05)=1.072>1$. This shows that the necessary
condition is not sufficient for the small-gain condition. The
sufficient condition is verified for the other three values of
$\lambda$. At $\beta=0.1$, it is verified in all four cases.

\paragraph{Remark.}
The values in the $\lambda=0$ row of
Table~\ref{tab:four-state-general-numerical-product} need not agree
with the corresponding values of
$\Gamma_0^{\mathrm{num}}(\beta)$ in
Table~\ref{tab:four-state-lambda-zero-certificates}. The former are
based on the general mean-field representation and its Lipschitz
constant $L_0$, whereas the latter are based on the explicit
one-step factorization involving $D_0$ and $\Phi$. These are two
different sufficient small-gain formulations of the same one-step
mean ODE.

Overall, this subsection gives the following three conclusions.
\begin{enumerate}[
    label=(\roman*),
    align=left,
    widest=iii,
    leftmargin=*
]
\item
For $\eta=1$, the reported values of
$G_\lambda^{\mathrm{num}}(\beta)$ decrease as $\beta$ increases,
making the sufficient convergence condition easier to satisfy.
\item
Increasing $\beta$ enlarges the range of $(\lambda,\beta)$ for
which the sufficient convergence condition can be verified.

\item
Whenever
$\beta\leq L_\lambda-\lambda_{\min}(C)$, no choice of $\eta>0$ can
make the general-trace small-gain condition hold.
\end{enumerate}
These conclusions are illustrated by the numerical values and
calculations above. The first two conclusions are consistent with
the dependence of $G_\lambda^{\mathrm{num}}(\beta)$ on $\beta$
through $M_{\beta,\lambda}$ in
\eqref{eq:four-state-general-numerical-product}. The third follows
from the necessary condition
\eqref{eq:beta-necessary-lower-bound} derived in the preceding
subsection.

\section{Conclusions}
\label{sec:conclusions}

In this paper, we established convergence results for constrained
single-time-scale TDRC($\lambda$) under linear function approximation.
Our analysis covers state-dependent and history-dependent
$\lambda$-parameters and non-i.i.d.\ off-policy data. By treating the
primary and secondary iterates jointly through a coupled linear mean
ODE, we proved convergence in probability and in mean to the target
fixed point.

We also introduced TDRC($\lambda$)-Huber by extending the Huber
Bellman-error formulation. We showed that the resulting algorithm
preserves the
fixed point of TDRC($\lambda$). Since the clipping induced by the Huber loss introduces a
nonlinear term into the mean ODE, we used an integral small-gain
condition to establish uniqueness and global asymptotic stability of
the equilibrium, and then proved convergence of the constrained
algorithm. We further derived computable sufficient conditions for
state-dependent and history-dependent $\lambda$-parameters, together
with specialized conditions for the one-step case. These results
also identify a regularization-based convergence mechanism shared by
linear TDRC($\lambda$) and TDRC($\lambda$)-Huber. For positive-stable
$A_\lambda$, sufficiently large $\beta$ gives a simple, computable
convergence guarantee for every $\eta>0$, including cases in which
$H_\lambda\not\succeq0$. The clipping induced by the Huber loss reduces sensitivity to large
Bellman errors while preserving this convergence guarantee and the target fixed point.
The four-state example illustrates how increasing $\beta$ can make
the sufficient convergence conditions easier to verify.

The analysis in this paper concerns constrained algorithms under linear
function approximation and diminishing stepsizes. One direction for
future work is to extend the convergence results to constant
stepsizes. Another is to extend the ODE-based small-gain approach
developed here beyond the Huber setting and apply it to a broader
class of algorithms with nonlinear mean ODEs. A further
direction is to develop less conservative stability analyses that
exploit the additional structure of the clipping residual, with the
aim of extending the Huber convergence analysis from positive-stable
$A_\lambda$ to the more general nonsingular setting covered by linear
TDRC($\lambda$).

\acks{We thank Dr. Huizhen Yu for helpful discussions. The research of Hao Wang was partially supported by the Natural Sciences and Engineering Research Council of Canada (Individual Discovery Grant RGPIN-2025-05734) and the Canada Research Chairs program (Tier 1 Canada
Research Chair Award).
}



\newpage

\appendix
\section{Proofs for Linear TDRC($\lambda$)}

\subsection{Proof of the Fixed-Point Result}
\label{app:tdrc-fixed-point-proof}

\begin{proof}[Proof of Proposition~\ref{prop:tdrc-fixed-point}]
Let
\[
C_\beta=C+\beta I,
\qquad
A_{\lambda,\beta}=A_\lambda+\beta I.
\]
Since $C$ is positive definite and $\beta\geq0$, $C_\beta$ is
positive definite and hence nonsingular. From
\eqref{eq:fixed-point-x},
\begin{equation}
x
=
C_\beta^{-1}
\left(
b_\lambda-A_\lambda\theta
\right).
\label{eq:auxiliary-fixed-point}
\end{equation}
Substituting \eqref{eq:auxiliary-fixed-point} into
\eqref{eq:fixed-point-theta} gives
\begin{align*}
0
&=
\left[
C_\beta+A_\lambda^\top-C
\right]
C_\beta^{-1}
\left(
b_\lambda-A_\lambda\theta
\right)
\nonumber =
A_{\lambda,\beta}^\top
C_\beta^{-1}
\left(
b_\lambda-A_\lambda\theta
\right).
\end{align*}
Since $A_{\lambda,\beta}$ and $C_\beta$ are nonsingular, it follows
that
\begin{equation*}
b_\lambda-A_\lambda\theta=0.
\end{equation*}
The nonsingularity of $A_\lambda$ then gives
\begin{equation*}
\theta
=
A_\lambda^{-1}b_\lambda
=
\theta_\lambda^*.
\end{equation*}
Substituting this result into
\eqref{eq:auxiliary-fixed-point} gives $x=0$.

Finally, $A_\lambda\theta^*=b_\lambda$ implies
$J_\lambda(\theta^*)=0$. Hence, $\theta_\lambda^*$ is the unique
minimizer of $J_\lambda(\theta)$.
\end{proof}

\subsection{Proof of the Sufficient Hurwitz Condition}
\label{app:tdrc-hurwitz-proof}

\begin{proof}[Proof of
Lemma~\ref{lem:tdrc-hurwitz-condition}]
Since \(A_\lambda\) is nonsingular,
\(A_\lambda A_\lambda^\top\succ0\). We first follow the characteristic-equation argument in
\citep[Appendix~H.3.1]{ghiassian2020gradient}. For each eigenvalue
\(s\in\mathbb C\) of \(G^\lambda\), there exists
\(v\in\mathbb C^d\setminus\{0\}\) such that
\begin{equation*}
s^2+
\left(
\eta\beta+\eta b_c+\mu
\right)s
+
\eta
\left(
\beta\mu+b_a
\right)
=0,
\end{equation*}
where
\begin{equation*}
b_c
=
\frac{v^*Cv}{\|v\|_2^2},
\qquad
b_a
=
\frac{v^*A_\lambda A_\lambda^\top v}{\|v\|_2^2},
\qquad
\mu
=
\frac{v^*A_\lambda v}{\|v\|_2^2}
=
\mu_r+\mathrm{i}\mu_i.
\end{equation*}
Since
\[
H_\lambda
=
\frac{A_\lambda+A_\lambda^\top}{2},
\]
we have
\[
\mu_r
=
\frac{v^*H_\lambda v}{\|v\|_2^2}.
\]

The conditions C1--C3 in
\citep[Appendix~H.3.1]{ghiassian2020gradient} are satisfied if
\begin{equation}
\beta\mu_r+b_a>0,
\qquad
\eta b_c+\mu_r>0
\label{eq:appendix-tdrc-two-scalar-conditions}
\end{equation}
for every \(v\neq0\). To see this, let
\[
q=\beta\mu_r+b_a,
\qquad
p=\eta b_c+\mu_r,
\]
and define
\[
\Omega=\eta\beta+p,
\qquad
\mathcal X=\mu_i^2+4\eta q.
\]
If \(p>0\) and \(q>0\), then
\[
\Omega>0,
\qquad
\Omega^2+\mathcal X>0,
\]
and
\[
\Omega^2q+\beta\mu_i^2p>0.
\]
These are C1--C3, respectively. It is therefore sufficient to verify
\eqref{eq:appendix-tdrc-two-scalar-conditions}.

By the definitions above,
\[
\beta\mu_r+b_a
=
\frac{
v^*(A_\lambda A_\lambda^\top+\beta H_\lambda)v
}{
\|v\|_2^2
},
\qquad
\eta b_c+\mu_r
=
\frac{
v^*(\eta C+H_\lambda)v
}{
\|v\|_2^2
}.
\]
Thus, \eqref{eq:appendix-tdrc-two-scalar-conditions} follows from
\begin{equation}
A_\lambda A_\lambda^\top+\beta H_\lambda\succ0,
\qquad
\eta C+H_\lambda\succ0.
\label{eq:tdrc-matrix-stability-condition}
\end{equation}

We now show that the conditions in
\eqref{eq:tdrc-spectral-condition} imply these two matrix
inequalities. If \(H_\lambda\succeq0\), then, since
\(A_\lambda\) is nonsingular, $A_\lambda A_\lambda^\top\succ0$, and
$\beta H_\lambda\succeq0$ for every \(\beta\geq0\). Therefore, $A_\lambda A_\lambda^\top+\beta H_\lambda\succ0$.

Suppose that
\(H_\lambda\not\succeq0\). For any
\(x\in\mathbb R^d\setminus\{0\}\), let $y=(A_\lambda A_\lambda^\top)^{1/2}x$.
Since \(A_\lambda\) is nonsingular, \(y\neq0\), and
\begin{align}
&x^\top
\left(
A_\lambda A_\lambda^\top+\beta H_\lambda
\right)x
\notag\\
&\quad=
y^\top
\left[
I+
\beta
(A_\lambda A_\lambda^\top)^{-1/2}
H_\lambda
(A_\lambda A_\lambda^\top)^{-1/2}
\right]y
\notag\\
&\quad\geq
\left[
1+
\beta\lambda_{\min}
\left(
(A_\lambda A_\lambda^\top)^{-1/2}
H_\lambda
(A_\lambda A_\lambda^\top)^{-1/2}
\right)
\right]
\|y\|_2^2.
\label{eq:appendix-tdrc-beta-quadratic-form}
\end{align}
The matrix
\((A_\lambda A_\lambda^\top)^{-1}H_\lambda\) is similar to $(A_\lambda A_\lambda^\top)^{-1/2}
H_\lambda
(A_\lambda A_\lambda^\top)^{-1/2}$. Therefore,
\[
\lambda_{\min}
\left(
(A_\lambda A_\lambda^\top)^{-1/2}
H_\lambda
(A_\lambda A_\lambda^\top)^{-1/2}
\right)
=
-\lambda_{\max}
\left(
-(A_\lambda A_\lambda^\top)^{-1}H_\lambda
\right).
\]
The upper bound on \(\beta\) gives $1-
\beta\lambda_{\max}
\left(
-(A_\lambda A_\lambda^\top)^{-1}H_\lambda
\right)
>0$.
It therefore follows from
\eqref{eq:appendix-tdrc-beta-quadratic-form} that $A_\lambda A_\lambda^\top+\beta H_\lambda\succ0$.

Next, for any \(x\in\mathbb R^d\setminus\{0\}\), let $y=C^{1/2}x$. Since \(C\succ0\), \(y\neq0\), and
\begin{align*}
x^\top(\eta C+H_\lambda)x
&=
y^\top
\left(
\eta I+C^{-1/2}H_\lambda C^{-1/2}
\right)y
\notag\\
&\geq
\left[
\eta+
\lambda_{\min}
\left(
C^{-1/2}H_\lambda C^{-1/2}
\right)
\right]
\|y\|_2^2.
\end{align*}
Since \(C^{-1}H_\lambda\) is similar to
\(C^{-1/2}H_\lambda C^{-1/2}\),
\[
\lambda_{\min}
\left(
C^{-1/2}H_\lambda C^{-1/2}
\right)
=
\lambda_{\min}(C^{-1}H_\lambda).
\]
The condition on \(\eta\) gives
\[
\eta+\lambda_{\min}(C^{-1}H_\lambda)>0,
\]
and hence
\[
\eta C+H_\lambda\succ0.
\]

We have proved \eqref{eq:tdrc-matrix-stability-condition}. Therefore,
\eqref{eq:appendix-tdrc-two-scalar-conditions} and Conditions C1--C3
hold. The root calculation in
\citep[Appendix~H.3.1]{ghiassian2020gradient} then gives
\[
\operatorname{Re}(s)<0
\]
for every eigenvalue \(s\) of \(G^\lambda\). Thus,
\(G^\lambda\) is Hurwitz.

It remains to prove that \(A_\lambda+\beta I\) is nonsingular.
Suppose, to the contrary, that
\(A_\lambda^\top+\beta I\) is singular. Since this matrix is real,
there exists a real vector \(v\neq0\) such that
\[
A_\lambda^\top v=-\beta v.
\]
It follows that
\[
v^\top A_\lambda A_\lambda^\top v
=
\|A_\lambda^\top v\|_2^2
=
\beta^2\|v\|_2^2.
\]
Moreover, since
\[
v^\top A_\lambda v
=
v^\top A_\lambda^\top v,
\]
we have
\[
v^\top H_\lambda v
=
\frac12
v^\top
\left(
A_\lambda+A_\lambda^\top
\right)v
=
v^\top A_\lambda^\top v
=
-\beta\|v\|_2^2.
\]
Consequently,
\[
\begin{aligned}
v^\top
\left(
A_\lambda A_\lambda^\top+\beta H_\lambda
\right)v
&=
\beta^2\|v\|_2^2
-\beta^2\|v\|_2^2=0,
\end{aligned}
\]
which contradicts
\(A_\lambda A_\lambda^\top+\beta H_\lambda\succ0\).
Therefore, \(A_\lambda^\top+\beta I\), and hence
\(A_\lambda+\beta I\), is nonsingular.
\end{proof}

\subsection{Proof of Projected Mean-ODE Stability}
\label{app:tdrc-projection-proof}
\begin{proof}[Proof of Lemma~\ref{sol-pro-ODE}]
Since
\[
R>R_0\geq\|u^*\|_2,
\]
the point \(u^*\) lies in the interior of \(B(R)\). Moreover,
\[
G^\lambda u^*+g^\lambda=0,
\]
so \(u^*\) is an equilibrium of the projected ODE.

Let
\[
\widetilde u=u-u^*
\]
and define
\[
V(u)=\widetilde u^\top P_G\widetilde u.
\]
Since
\[
(G^\lambda)^\top P_G+P_GG^\lambda=-I,
\]
every solution of the projected ODE satisfies, almost everywhere,
\begin{align*}
\dot V(u)
&=
2\widetilde u^\top P_G
\left(G^\lambda\widetilde u+z\right)=
-\|\widetilde u\|_2^2
+2\widetilde u^\top P_Gz.
\end{align*}

If \(\|u\|_2<R\), then \(z=0\). If \(u\) lies on the
boundary of \(B(R)\), then
\[
z=-au
\]
for some \(a\geq0\), where \(a=0\) includes the case in which the
projection is inactive. On the boundary,
\begin{align*}
\widetilde u^\top P_Gu
&=
u^\top P_Gu-(u^*)^\top P_Gu
\geq
\lambda_{\min}(P_G)R^2
-\|P_Gu^*\|_2R
>0,
\end{align*}
where the last inequality follows from \(R>R_0\). Therefore,
\[
2\widetilde u^\top P_Gz
=
-2a\widetilde u^\top P_Gu
\leq0.
\]
Together with the interior case, this gives
\begin{align}
\dot V(u)
&\leq
-\|\widetilde u\|_2^2
\leq
-\frac{1}{\lambda_{\max}(P_G)}V(u)
\qquad\text{almost everywhere}.
\label{eq:appendix-tdrc-exponential-bound}
\end{align}

Using the integrating factor $\exp\left(
\frac{t}{\lambda_{\max}(P_G)}
\right)$,
\eqref{eq:appendix-tdrc-exponential-bound} gives
\begin{align*}
\frac{d}{dt}
\left[
\exp\left(
\frac{t}{\lambda_{\max}(P_G)}
\right)
V(u(t))
\right]
\leq0
\qquad\text{almost everywhere}.
\end{align*}
Therefore,
\begin{equation*}
V(u(t))
\leq
\exp\left(
-\frac{t}{\lambda_{\max}(P_G)}
\right)
V(u(0)).
\end{equation*}
Since
\[
\lambda_{\min}(P_G)\|\widetilde u\|_2^2
\leq
V(u)
\leq
\lambda_{\max}(P_G)\|\widetilde u\|_2^2,
\]
we obtain
\begin{align}
\|u(t)-u^*\|_2
&\leq
\sqrt{
\frac{\lambda_{\max}(P_G)}
     {\lambda_{\min}(P_G)}
}
\exp\left(
-\frac{t}{2\lambda_{\max}(P_G)}
\right)
\|u(0)-u^*\|_2.
\label{eq:appendix-tdrc-state-exponential}
\end{align}
The multiplicative constant is finite, and the exponential factor
is at most one. Hence,
\eqref{eq:appendix-tdrc-state-exponential} implies Lyapunov
stability
\citep[Definition~1.31]{chicone2024ordinary}. The exponential
factor also converges to zero as \(t\to\infty\), so every solution
starting in \(B(R)\) converges to \(u^*\). Thus, \(u^*\) is
globally asymptotically stable relative to \(B(R)\); see
\citep[Definition~1.33]{chicone2024ordinary}. It is the unique
equilibrium in \(B(R)\), since any other equilibrium would
contradict \eqref{eq:appendix-tdrc-state-exponential}.

It remains to identify the limit set. Following
\citet[Section~3.1]{yu2016weak}, the limit set of the projected ODE
associated with \(B(R)\) is
\begin{equation}
L_{B(R)}
=
\bigcap_{\bar t>0}
\overline{
\bigcup_{u(0)\in B(R)}
\left\{
u(t):t\geq\bar t
\right\}
},
\label{eq:appendix-tdrc-limit-set-definition}
\end{equation}
where the union is taken over all solutions with initial conditions
in \(B(R)\), and the overline denotes closure. For every
\(u(0)\in B(R)\),
\[
\|u(0)-u^*\|_2
\leq
R+\|u^*\|_2.
\]
Therefore, \eqref{eq:appendix-tdrc-state-exponential} gives
\[
0
\leq
\lim_{t\to\infty}\|u(t)-u^*\|_2
\leq
\lim_{t\to\infty}
\sqrt{
\frac{\lambda_{\max}(P_G)}
{\lambda_{\min}(P_G)}
}
\exp\left(
-\frac{t}{2\lambda_{\max}(P_G)}
\right)
\left(R+\|u^*\|_2\right)
=
0.
\]
Thus, for every \(\delta>0\), there exists
\(\bar t_\delta<\infty\), independent of the initial condition and
the choice of solution, such that
\[
\|u(t)-u^*\|_2<\delta
\qquad
\text{for all }t\geq\bar t_\delta
\]
for every solution starting in \(B(R)\). Consequently,
\[
\bigcup_{u(0)\in B(R)}
\left\{
u(t):t\geq\bar t_\delta
\right\}
\subseteq
N_\delta(u^*).
\]
It follows from
\eqref{eq:appendix-tdrc-limit-set-definition} that
\[
L_{B(R)}\subseteq \overline{
\bigcup_{u(0)\in B(R)}
\left\{
u(t):t\geq\bar t_\delta
\right\}
}
\subseteq
\overline{N_\delta(u^*)}.
\]
Since \(\delta>0\) is arbitrary,
\[
L_{B(R)}
\subseteq
\bigcap_{\delta>0}
\overline{N_\delta(u^*)}
=
\{u^*\}.
\] Conversely, \(u(t)\equiv u^*\) is a solution of the projected ODE.
Therefore, for every \(\bar t>0\),
\[
u^*
\in
\bigcup_{u(0)\in B(R)}
\left\{
u(t):t\geq\bar t
\right\},
\]
and hence \(u^*\in L_{B(R)}\). Combining the two inclusions gives
\[
L_{B(R)}=\{u^*\}.
\]
\end{proof}

\subsection{Proofs of the Main Convergence Results}
\label{app:tdrc-main-proof}

\begin{proof}[Proof of Theorem~\ref{thm:tdrc-convergence}]
Appendix~\ref{app:tdrc-sa-verification} states and verifies the
hypotheses of the
weak-convergence theorem for constrained stochastic approximation
\citep[Theorem~8.2.3]{kushner2003}. The projected mean ODE is
\[
\dot u=G^\lambda u+g^\lambda+z,
\qquad
z\in-\mathcal N_{B(R)}(u).
\]
The theorem therefore yields a sequence \(T_k\to\infty\) such that,
for every \(\delta>0\),
\[
\limsup_{k\to\infty}
\mathbb P\left(
U_i\notin N_\delta(L_{B(R)})
\text{ for some }i\in[k,m(k,T_k)]
\right)=0,
\]
where \(L_{B(R)}\) is the limit set of the projected ODE.
Under \eqref{eq:tdrc-spectral-condition},
Lemma~\ref{lem:tdrc-hurwitz-condition} shows that \(G^\lambda\) is Hurwitz. Since \(R>R_0\), Lemma~\ref{sol-pro-ODE} gives
\[
L_{B(R)}=\{u^*\}.
\]
Substitution of this identity gives
\eqref{eq:tdrc-window-convergence} and hence convergence in
probability.
\end{proof}

\begin{proof}[Proof of Corollary~\ref{cor:tdrc-mean-convergence}]
Theorem~\ref{thm:tdrc-convergence} gives convergence in probability.
Because \(U_n\) and \(u^*\) belong to the bounded set \(B(R)\),
\citet[Theorem~10.3.6]{Dudley_2002} implies that convergence also
holds in mean.
\end{proof}

\begin{proof}[Proof of
Corollary~\ref{cor:tdrc-positive-stable-large-beta}]
For $\tau=\infty$, extend the definition of
$\operatorname{clip}_\tau$ by setting
\[
\operatorname{clip}_\infty(a):=a,
\qquad a\in\mathbb R.
\]
By \eqref{identities}, we have
\[
\mathcal H_{\lambda,\infty}(x)
=
\mathbb E_\zeta\!\left[
\rho_0(1-\lambda_1)\gamma_1
\phi(S_1)e_0^\top x
\right]
=
(C-A_\lambda^\top)x.
\]
Replacing $\operatorname{clip}_\tau$ by the identity map in the
proof of Proposition~\ref{prop:huber-mean-field} gives
\[
\mathcal H_{\lambda,\infty}(0)=0
\]
and
\[
\|\mathcal H_{\lambda,\infty}(x)
-\mathcal H_{\lambda,\infty}(y)\|_2
\leq
L_\lambda\|x-y\|_2.
\]
Under the stated lower bound on $\beta$, the integral-gain estimate
\eqref{eq:huber-general-gain-final-bound} gives
\[
L_\lambda
\Gamma_2
\left(
M_{\beta,\lambda},
B_\lambda,
\mathcal C_\lambda
\right)
<1.
\]
Lemma~\ref{lem:integral-stability-criterion}, applied with
$F=\mathcal H_{\lambda,\infty}$, therefore shows that
\[
\dot u
=
G^\lambda(u-u^*)
\]
is globally asymptotically stable. Hence, $G^\lambda$ is Hurwitz.
Lemma~\ref{sol-pro-ODE} and the same weak-convergence argument as in
the proof of Theorem~\ref{thm:tdrc-convergence} give the stated
convergence conclusions.
\end{proof}

\subsection{Verification of the Stochastic Approximation Conditions
for \texorpdfstring{TDRC($\lambda$)}{TDRC(lambda)}}
\label{app:tdrc-sa-verification}

This subsection states and verifies the four stochastic approximation
conditions required by the weak-convergence theorem.

Recall the notation
\[
u=
\begin{bmatrix}
x\\
\theta
\end{bmatrix},
\qquad
F(u,z)
=
\begin{bmatrix}
h(\theta,x,z)\\
g(\theta,x,z)
\end{bmatrix},
\qquad
\bar F(u)
=
\begin{bmatrix}
\bar h(\theta,x)\\
\bar g(\theta,x)
\end{bmatrix},
\]
and let
\[
\varepsilon_{n+1}
=
\begin{bmatrix}
\eta e_n\omega_{n+1}\\
e_n\omega_{n+1}
\end{bmatrix}.
\]
Following the weak-convergence approach to stochastic approximation
\citep{kushner2003}, we verify the following conditions to show that
the projected ODE
\[
\dot u
=
\bar F(u)+z,
\qquad
z\in-\mathcal N_{B(R)}(u),
\]
is the mean ODE associated with the constrained
TDRC($\lambda$) recursion \eqref{eq:tdrc-sa-form}.
\begin{enumerate}
\item
The following families of random variables are uniformly integrable:
\[
\{F(U_n,Z_n)\}_{n\geq0},
\qquad
\{F(u,Z_n):n\geq0,\ u\in B(R)\},
\]
and
\[
\{F(U_n,Z_n)+\varepsilon_{n+1}\}_{n\geq0}.
\]

\item
The family $\{Z_n\}_{n\geq0}$ is tight.

\item
For every compact set $D\subset\mathcal Z$, the function $F(u,z)$ is
continuous in $u$, uniformly over $z\in D$.

\item
The mean field $\bar F:B(R)\to\mathbb R^{2d}$ is continuous and, for
every fixed $u\in B(R)$ and every compact set
$D\subset\mathcal Z$, the conditional averaging property in
Proposition~\ref{prop:convergence-in-mean-F} holds.
\end{enumerate}

We verify these four conditions in turn.

\subsubsection{Verification of Condition 1}

\begin{proposition}
\label{prop:uniform-integrability-F}
For every initial trace $e_0$, the following families are uniformly
integrable:
\begin{enumerate}
\item $\{e_n\}_{n\geq0}$;

\item $\{F(u,Z_n):n\geq0,\ u\in B(R)\}$;

\item $\{F(U_n,Z_n)\}_{n\geq0}$;

\item
$\{F(U_n,Z_n)+\varepsilon_{n+1}\}_{n\geq0}$.
\end{enumerate}
\end{proposition}

\begin{proof}
The argument follows \citet[Section~3.1.2]{Yugtd}. By
Proposition~\ref{prop:state-trace-properties}\textnormal{(iii)}, the
trace sequence $\{e_n\}$ is uniformly integrable. In addition, the
functions $g$ and $k$ are Lipschitz continuous in the trace variable,
uniformly over bounded parameter sets and the remaining components of
$z$; see \citet[Section~3.1.2]{Yugtd}.
Recall that
\[
h(\theta,x,z)
=
\eta\bigl(k(\theta,x,z)-\beta x\bigr).
\]
Since the regularization term $-\beta x$ does not depend on the trace
variable, the Lipschitz continuity of $k$ in the trace variable
implies the same property for $h$.

The remaining argument is the same as that in
\citet[Section~3.1.2]{Yugtd}. It uses the boundedness of $B(R)$, the
finiteness of the state space, the uniform integrability of
$\{e_n\}$, the Lipschitz continuity in the trace variable stated
above, and the finite variance of the noise terms
$\{\omega_{n+1}\}$. These properties give uniform integrability for
the corresponding component functions. Finally,
\[
F(U_n,Z_n)
=
\begin{bmatrix}
h(\theta_n,x_n,Z_n)\\
g(\theta_n,x_n,Z_n)
\end{bmatrix}
\]
and
\[
F(U_n,Z_n)+\varepsilon_{n+1}
=
\begin{bmatrix}
h(\theta_n,x_n,Z_n)+\eta e_n\omega_{n+1}\\
g(\theta_n,x_n,Z_n)+e_n\omega_{n+1}
\end{bmatrix}.
\]
The stated uniform integrability results therefore follow from the
corresponding componentwise results.
\end{proof}

\subsubsection{Verification of Condition 2}

Condition~2 follows directly from
\citet[Section~3.1.2]{Yugtd}, since our definition of
$\{Z_n\}$ and the assumptions imposed on the state--trace process
are the same as those used therein.

\subsubsection{Verification of Condition 3}

\begin{proposition}
\label{prop:tdrc-uniform-continuity}
For every compact set $D\subset\mathcal Z$, the function $F(u,z)$ is
continuous in $u$, uniformly over $z\in D$.
\end{proposition}

\begin{proof}
Write $u=(x^\top,\theta^\top)^\top$ and fix a compact set
$D\subset\mathcal Z$. By the same argument as in
\citet[Section~3.1.2]{Yugtd}, the functions
$g(\theta,x,z)$ and $k(\theta,x,z)$ are continuous in $(\theta,x)$,
uniformly over $z\in D$.

It remains to consider
\[
h(\theta,x,z)
=
\eta\bigl(k(\theta,x,z)-\beta x\bigr).
\]
For every $u,u'$ and $z\in D$,
\begin{align*}
&\|h(\theta,x,z)-h(\theta',x',z)\|_2
\leq
\eta
\|k(\theta,x,z)-k(\theta',x',z)\|_2
+
\eta\beta\|x-x'\|_2.
\end{align*}
Since $\|x-x'\|_2\leq\|u-u'\|_2$ and $k$ is continuous in
$(\theta,x)$ uniformly over $z\in D$, the right-hand side converges
to zero uniformly over $z\in D$ as $u'\to u$. Hence, $h$ is
continuous in $(\theta,x)$ uniformly over $z\in D$.

Therefore,
\[
F(u,z)
=
\begin{pmatrix}
h(\theta,x,z)\\
g(\theta,x,z)
\end{pmatrix}
\]
is continuous in $u$, uniformly over $z\in D$.
\end{proof}

\subsubsection{Verification of Condition 4}

\begin{proposition}
\label{prop:convergence-in-mean-F}
The mean field $\bar F:B(R)\to\mathbb R^{2d}$ is continuous.
Moreover, for each fixed $u\in B(R)$ and each compact set
$D\subset\mathcal Z$,
\begin{equation}
\lim_{m,n\to\infty}
\frac{1}{m}
\sum_{i=n}^{n+m-1}
\mathbb E_n
\bigl[
F(u,Z_i)-\bar F(u)
\bigr]
\mathbf 1_{\{Z_n\in D\}}
=
0
\quad\text{in mean}.
\label{eq:tdrc-conditional-averaging}
\end{equation}
\end{proposition}

\begin{proof}
The mean field $\bar F(u)=G^\lambda u+g^\lambda$ is affine and hence
continuous. The argument follows
\citep[Lemma~3.4]{Yugtd}. Since
$\mathbf 1_{\{Z_n\in D\}}$ is measurable with respect to the information
available at time $n$, conditional Jensen's inequality shows that it
is sufficient to prove the stated convergence with $\mathbb E_n$
removed.

Fix $u=(x^\top,\theta^\top)^\top\in B(R)$. The function
$F(u,\cdot)$ is Lipschitz continuous in the trace variable. Indeed,
this property holds for $g(\theta,x,\cdot)$ and
$k(\theta,x,\cdot)$ by \citet[Section~3.1.2]{Yugtd}, and hence also
holds for
\[
h(\theta,x,z)
=
\eta\bigl(k(\theta,x,z)-\beta x\bigr).
\]
Moreover, by the definition of the mean field,
\[
\mathbb E_\zeta
\bigl[
F(u,Z_0)
\bigr]
=
\bar F(u).
\]

It therefore follows from
Proposition~\ref{prop:state-trace-properties} that, for every compact set
$D\subset\mathcal Z$,
\[
\frac{1}{m}
\sum_{i=n}^{n+m-1}
\bigl[
F(u,Z_i)-\bar F(u)
\bigr]
\mathbf 1_{\{Z_n\in D\}}
\longrightarrow 0
\quad\text{in mean}
\]
as $m,n\to\infty$. Conditional Jensen's inequality then gives
\eqref{eq:tdrc-conditional-averaging}.
\end{proof}

The four conditions above, together with
Lemma~\ref{sol-pro-ODE}, verify the hypotheses required to apply the
weak-convergence theorem to \eqref{eq:tdrc-sa-form}.

\section{Proofs for TDRC($\lambda$)-Huber}

\subsection{Proof of the Fixed-Point Result}
\label{app:tdrc-huber-fixed-point-proof}

\begin{proof}[Proof of
Proposition~\ref{prop:tdrc-huber-fixed-point}]
Since
\[
A_\lambda\theta_\lambda^*=b_\lambda
\qquad\text{and}\qquad
\operatorname{clip}_\tau(0)=0,
\]
substituting $(\theta,x)=(\theta_\lambda^*,0)$ into
\eqref{eq:huber-fixed-point-theta} and
\eqref{eq:huber-fixed-point-x} gives zero in both equations.
Therefore, $(\theta_\lambda^*,0)$ is a fixed point. The result for
TDRC(0)-Huber follows by setting $\lambda_t\equiv0$.
\end{proof}

\subsection{Proofs of the Mean-Field Representations}
\label{app:huber-mean-field-proofs}

\begin{proof}[Proof of Proposition~\ref{prop:huber-mean-field}]
Since $\operatorname{clip}_\tau(0)=0$,
$\mathcal H_{\lambda,\tau}(0)=0$. For arbitrary
$x,y\in\mathbb R^d$, the triangle inequality, the $1$-Lipschitz
property of the scalar clipping function, and the Cauchy--Schwarz
inequality give
\begin{align*}
&\left\|
\mathcal H_{\lambda,\tau}(x)
-\mathcal H_{\lambda,\tau}(y)
\right\|_2
\\
&\quad\leq
\mathbb E_\zeta\!\left[
\rho_0(1-\lambda_1)\gamma_1
\|\phi(S_1)\|_2
\left|
\operatorname{clip}_\tau(e_0^\top x)
-\operatorname{clip}_\tau(e_0^\top y)
\right|
\right]
\\
&\quad\leq
\mathbb E_\zeta\!\left[
\rho_0(1-\lambda_1)\gamma_1
\|\phi(S_1)\|_2
|e_0^\top(x-y)|
\right]
\\
&\quad\leq
L_\lambda\|x-y\|_2.
\end{align*}
By Proposition~\ref{prop:state-trace-properties}\textnormal{(ii)},
$\mathbb E_\zeta[\|e_0\|_2]<\infty$. The importance-sampling
ratios and feature vectors are bounded under
Condition~\ref{cond:target-behavior} and the finite-state setting.
Since $0\leq1-\lambda_1\leq1$ and $0\leq\gamma_1\leq1$,
\begin{equation*}
L_\lambda
\leq
\rho_{\max}\phi_{\max}
\mathbb E_\zeta[\|e_0\|_2]
<\infty.
\end{equation*}
Here,
\[
\rho_{\max}
:=
\max_{\{(s,s'):\,P^o_{ss'}>0\}}\rho(s,s'),
\qquad
\phi_{\max}
:=
\max_{s\in\mathcal S}\|\phi(s)\|_2.
\]
Setting $y=0$ gives
\eqref{eq:huber-mean-field-linear-growth}, while
$|\operatorname{clip}_\tau(a)|\leq\tau$ gives
\eqref{eq:huber-general-H-uniform-bound}.
\end{proof}

\begin{proof}[Proof of Corollary~\ref{cor:huber-one-step-mean-field}]
Let
\[
c_x
:=
\Pi_{[-\tau,\tau]^{|\mathcal S|}}(\Phi x).
\]
For each $i=1,\ldots,d$, using a calculation similar to that in the
proof of Proposition~2.1 of \citet{Yugtd}, we obtain
\begin{align*}
[\mathcal H_{0,\tau}(x)]_i
&=
\sum_{s\in\mathcal S}
\xi_s[c_x]_s
\mathbb E_\zeta\!\left[
\rho_0\gamma_1\phi_i(S_1)
\,\middle|\,
S_0=s
\right]
\\
&=
\sum_{s\in\mathcal S}
\xi_s[c_x]_s
(P\Gamma\Phi_i)(s)
\\
&=
c_x^\top\Xi P\Gamma\Phi_i.
\end{align*}
Therefore,
\begin{align*}
\mathcal H_{0,\tau}(x)
&=
\left(
c_x^\top\Xi P\Gamma\Phi
\right)^\top
=
\Phi^\top(P\Gamma)^\top\Xi c_x
=
D_0c_x,
\end{align*}
which proves \eqref{eq:huber-mean-field-lambda-zero}.

For arbitrary $v,w\in\mathbb R^{|\mathcal S|}$, the inequality
\[
\left|
\operatorname{clip}_\tau(a)
-
\operatorname{clip}_\tau(b)
\right|
\leq
|a-b|
\]
gives
\begin{align*}
\left\|
\Pi_{[-\tau,\tau]^{|\mathcal S|}}(v)
-
\Pi_{[-\tau,\tau]^{|\mathcal S|}}(w)
\right\|_2^2
&=
\sum_{s\in\mathcal S}
\left|
\operatorname{clip}_\tau(v_s)
-
\operatorname{clip}_\tau(w_s)
\right|^2
\leq
\sum_{s\in\mathcal S}
\left|
v_s-w_s
\right|^2
=
\|v-w\|_2^2.
\end{align*}
Hence,
\[
\left\|
\Pi_{[-\tau,\tau]^{|\mathcal S|}}(v)
-
\Pi_{[-\tau,\tau]^{|\mathcal S|}}(w)
\right\|_2
\leq
\|v-w\|_2,
\qquad
v,w\in\mathbb R^{|\mathcal S|}.
\]
This proves \eqref{eq:proj-zero-explicit-lipschitz}.
\end{proof}

\subsection{Proof of the Integral Stability Criterion}
\label{app:huber-stability-proofs}

\begin{proof}[Proof of
Lemma~\ref{lem:integral-stability-criterion}]
Let
\[
\widetilde u(t):=u(t)-u^*.
\]
Since $F(0)=0$, the global Lipschitz condition
\eqref{eq:F-global-lipschitz} gives
\begin{equation}
\|F(v)\|_2
=
\|F(v)-F(0)\|_2
\leq
L_F\|v\|_2,
\qquad v\in\mathbb R^p.
\label{eq:small-gain-linear-growth}
\end{equation}
Under the change of variables $\widetilde u=u-u^*$, the system becomes
\begin{equation*}
\dot{\widetilde u}
=
M\widetilde u
+
BF(\mathcal C\widetilde u).
\end{equation*}
Its vector field is globally Lipschitz. Indeed, for arbitrary
$\widetilde u_1,\widetilde u_2\in\mathbb R^m$,
\begin{align*}
&\left\|
M\widetilde u_1+BF(\mathcal C\widetilde u_1)
-
M\widetilde u_2-BF(\mathcal C\widetilde u_2)
\right\|_2
\notag\\
&\quad\leq
\|M\|_2
\|\widetilde u_1-\widetilde u_2\|_2
+
\|B\|_2
\left\|
F(\mathcal C\widetilde u_1)
-
F(\mathcal C\widetilde u_2)
\right\|_2
\notag\\
&\quad\leq
\left(
\|M\|_2
+
L_F\|B\|_2\|\mathcal C\|_2
\right)
\|\widetilde u_1-\widetilde u_2\|_2.
\end{align*}
Hence, for every initial condition, the unprojected system has a
unique solution defined for all $t\geq0$.

Let
\begin{equation*}
y(t)
:=
\mathcal C\widetilde u(t).
\end{equation*}
By the variation-of-constants formula,
\begin{equation}
\widetilde u(t)
=
e^{Mt}\widetilde u(0)
+
\int_0^t
e^{M(t-s)}BF(y(s))\,ds.
\label{eq:small-gain-variation-u}
\end{equation}
Multiplying both sides by $\mathcal C$ gives
\begin{equation*}
y(t)
=
\mathcal C e^{Mt}\widetilde u(0)
+
\int_0^t
\mathcal C e^{M(t-s)}BF(y(s))\,ds.
\end{equation*}
Using \eqref{eq:small-gain-linear-growth}, we obtain
\begin{align}
\|y(t)\|_2
&\leq
\|\mathcal C e^{Mt}\|_2
\|\widetilde u(0)\|_2+
L_F
\int_0^t
\left\|
\mathcal C e^{M(t-s)}B
\right\|_2
\|y(s)\|_2\,ds.
\label{eq:small-gain-output-bound}
\end{align}

Since $M$ is Hurwitz, it follows from
\citep[p.~90]{hespanha2018linear}
that there exist constants $K_M>0$ and
$\alpha_M>0$ such that
\begin{equation}
\|e^{Mt}\|_2
\leq
K_Me^{-\alpha_Mt},
\qquad t\geq0.
\label{eq:small-gain-M-exponential-bound}
\end{equation}

Using \eqref{eq:small-gain-M-exponential-bound}, we have
\[
\|\mathcal C e^{Mt}\|_2
\leq
\|\mathcal C\|_2\|e^{Mt}\|_2
\leq
K_M\|\mathcal C\|_2e^{-\alpha_Mt},
\qquad t\geq0.
\]
Consequently,
\begin{align*}
K_{\mathcal C}
:=
\sup_{t\geq0}\|\mathcal C e^{Mt}\|_2
&\leq
K_M\|\mathcal C\|_2
<\infty.
\end{align*}

Similarly,
\[
\|e^{Mt}B\|_2
\leq
\|e^{Mt}\|_2\|B\|_2
\leq
K_Me^{-\alpha_Mt}\|B\|_2,
\qquad t\geq0.
\]
Therefore,
\begin{align*}
\Gamma_{B,2}
:=
\int_0^\infty
\|e^{Mt}B\|_2\,dt
&\leq
K_M\|B\|_2
\int_0^\infty e^{-\alpha_Mt}\,dt=
\frac{K_M\|B\|_2}{\alpha_M}
<\infty.
\end{align*}

For $T>0$, define
\begin{align*}
Y_T
:=
\sup_{0\leq t\leq T}\|y(t)\|_2.
\end{align*}
For every $t\in[0,T]$, it follows from
\eqref{eq:small-gain-output-bound} that
\begin{align*}
\|y(t)\|_2
&\leq
K_{\mathcal C}\|\widetilde u(0)\|_2
+
L_FY_T
\int_0^t
\left\|
\mathcal C e^{M(t-s)}B
\right\|_2\,ds
\notag\\
&\leq
K_{\mathcal C}\|\widetilde u(0)\|_2
+
L_F\Gamma_2(M,B,\mathcal C)Y_T.
\end{align*}
Taking the supremum over $t\in[0,T]$ gives
\[
Y_T
\leq
K_{\mathcal C}\|\widetilde u(0)\|_2
+
L_F\Gamma_2(M,B,\mathcal C)Y_T.
\]
Therefore,
\begin{align*}
Y_T
\leq
\frac{K_{\mathcal C}}
{1-L_F\Gamma_2(M,B,\mathcal C)}
\|\widetilde u(0)\|_2.
\end{align*}
Since the right-hand side is independent of $T$, letting
$T\to\infty$ gives
\begin{align}
\sup_{t\geq0}\|y(t)\|_2
\leq
\frac{K_{\mathcal C}}
{1-L_F\Gamma_2(M,B,\mathcal C)}
\|\widetilde u(0)\|_2.
\label{eq:small-gain-y-bound}
\end{align}

We next show that $y(t)\to0$. Define
\[
\overline y
:=
\limsup_{t\to\infty}\|y(t)\|_2.
\]
By \eqref{eq:small-gain-y-bound},
$\overline y<\infty$. For every $\varepsilon>0$, there exists
$T_\varepsilon>0$ such that
\begin{align}
\|y(s)\|_2
\leq
\overline y+\varepsilon,
\qquad s\geq T_\varepsilon.
\label{eq:small-gain-limsup-bound}
\end{align}
For $t\geq T_\varepsilon$, splitting the integral in
\eqref{eq:small-gain-output-bound} at $T_\varepsilon$ gives
\begin{align}
\|y(t)\|_2
&\leq
\left\|
\mathcal C e^{Mt}
\right\|_2
\|\widetilde u(0)\|_2
\notag\\
&\quad+
L_F
\int_0^{T_\varepsilon}
\left\|
\mathcal C e^{M(t-s)}B
\right\|_2
\|y(s)\|_2\,ds
\notag\\
&\quad+
L_F
\int_{T_\varepsilon}^{t}
\left\|
\mathcal C e^{M(t-s)}B
\right\|_2
\|y(s)\|_2\,ds.
\label{eq:small-gain-y-split}
\end{align}

The first term on the right-hand side converges to zero by
\eqref{eq:small-gain-M-exponential-bound}. For the second term,
\eqref{eq:small-gain-M-exponential-bound} gives
\begin{align*}
&\int_0^{T_\varepsilon}
\left\|
\mathcal C e^{M(t-s)}B
\right\|_2
\|y(s)\|_2\,ds
\leq
\|\mathcal C\|_2K_M\|B\|_2
e^{-\alpha_M(t-T_\varepsilon)}
\int_0^{T_\varepsilon}\|y(s)\|_2\,ds,
\end{align*}
which also converges to zero as $t\to\infty$.

For the final term,
\eqref{eq:small-gain-limsup-bound} gives
\begin{align*}
&L_F
\int_{T_\varepsilon}^{t}
\left\|
\mathcal C e^{M(t-s)}B
\right\|_2
\|y(s)\|_2\,ds
\\
&\quad\leq
L_F(\overline y+\varepsilon)
\int_{T_\varepsilon}^{t}
\left\|
\mathcal C e^{M(t-s)}B
\right\|_2\,ds
\\
&\quad\leq
L_F\Gamma_2(M,B,\mathcal C)
(\overline y+\varepsilon).
\end{align*}
Taking the limit superior in
\eqref{eq:small-gain-y-split}, we obtain
\[
\overline y
\leq
L_F\Gamma_2(M,B,\mathcal C)
(\overline y+\varepsilon).
\]
Since $\varepsilon>0$ is arbitrary, we obtain \[
\overline y
\leq
L_F\Gamma_2(M,B,\mathcal C)\overline y.
\]
Since $L_F\Gamma_2(M,B,\mathcal C)<1$, it follows that
$\overline y=0$. Hence,
\begin{align*}
\lim_{t\to\infty}y(t)=0.
\end{align*}

Returning to \eqref{eq:small-gain-variation-u}, we have
\begin{align*}
\|\widetilde u(t)\|_2
&\leq
\|e^{Mt}\|_2
\|\widetilde u(0)\|_2
+
L_F
\int_0^t
\|e^{M(t-s)}B\|_2
\|y(s)\|_2\,ds.
\end{align*}
The first term converges to zero because $M$ is Hurwitz.
To treat the second term, fix $\varepsilon>0$ and choose
$T_\varepsilon>0$ such that
\[
\|y(s)\|_2
\leq
\varepsilon,
\qquad s\geq T_\varepsilon.
\]
Splitting the integral at $T_\varepsilon$, the integral over
$[0,T_\varepsilon]$ converges to zero by
\eqref{eq:small-gain-M-exponential-bound}, while
\begin{align*}
&L_F
\int_{T_\varepsilon}^{t}
\|e^{M(t-s)}B\|_2
\|y(s)\|_2\,ds
\\
&\quad\leq
L_F\varepsilon
\int_{T_\varepsilon}^{t}
\|e^{M(t-s)}B\|_2\,ds
\\
&\quad\leq
L_F\varepsilon\Gamma_{B,2}.
\end{align*}
First letting $t\to\infty$ and then letting
$\varepsilon$ tend to zero, we conclude that
\begin{align}
\lim_{t\to\infty}\widetilde u(t)=0.
\label{eq:small-gain-u-convergence}
\end{align}

It remains to verify Lyapunov stability of $u^*$. From
\eqref{eq:small-gain-variation-u},
\eqref{eq:small-gain-linear-growth}, and
\eqref{eq:small-gain-y-bound}, we obtain
\begin{align*}
\|\widetilde u(t)\|_2
&\leq
K_M\|\widetilde u(0)\|_2
+
L_F\Gamma_{B,2}
\sup_{s\geq0}\|y(s)\|_2
\notag\\
&\leq
\left(
K_M
+
\frac{
L_F\Gamma_{B,2}K_{\mathcal C}
}{
1-L_F\Gamma_2(M,B,\mathcal C)
}
\right)
\|\widetilde u(0)\|_2.
\end{align*}
Thus, $u^*$ is Lyapunov stable. Together with
\eqref{eq:small-gain-u-convergence}, this proves that $u^*$ is
globally asymptotically stable. It is also the unique equilibrium,
since a solution starting from any other equilibrium would be
constant and could not converge to $u^*$.
\end{proof}

\subsection{Proof of Projected Mean-ODE Stability}
\label{app:huber-projected-ode-proof}
\begin{proof}[Proof of
Lemma~\ref{lem:huber-invariant-projection-set}]
Since
\[
R>R_Q\geq\|u^*\|_2,
\]
we have $u^*\in\operatorname{int}B(R)$ and hence
$\mathcal N_{B(R)}(u^*)=\{0\}$. Since $u^*$ is an equilibrium of the
unprojected ODE, it is also an equilibrium of the projected ODE.

Let
\[
W(u):=u^\top Qu.
\]
Since
\[
M^\top Q+QM=-I,
\]
along every solution of the projected ODE,
\begin{align*}
\dot W(u)
&=
2u^\top Q
\left[
M(u-u^*)
+
BF\left(\mathcal C(u-u^*)\right)
+
z
\right]
\notag\\
&=
-\|u\|_2^2
-
2u^\top QMu^*
+
2u^\top QB
F\left(\mathcal C(u-u^*)\right)
+
2u^\top Qz
\notag\\
&\leq
-\|u\|_2^2
+
2\|QMu^*\|_2\|u\|_2
+
2\|QB\|_2\overline F\|u\|_2
+
2u^\top Qz
\notag\\
&=
-\|u\|_2^2
+
2K_Q\|u\|_2
+
2u^\top Qz.
\end{align*}

If $u\in\operatorname{int}B(R)$, then $z=0$. If $u$ lies on the
boundary of $B(R)$, then
\[
z=-au
\]
for some $a\geq0$. Hence,
\[
2u^\top Qz
=
-2a\,u^\top Qu
\leq
0.
\]
Together with the interior case, this gives
\begin{equation*}
\dot W(u)
\leq
-\|u\|_2
\left(
\|u\|_2-2K_Q
\right).
\end{equation*}

Since $R>R_Q$, we can choose $s$ such that
\begin{equation}
2K_Q
<
s
<
R
\sqrt{
\frac{\lambda_{\min}(Q)}
     {\lambda_{\max}(Q)}
}.
\label{eq:projected-choice-s}
\end{equation}
Define
\begin{equation*}
E_s
:=
\left\{
u:
u^\top Qu
\leq
\lambda_{\max}(Q)s^2
\right\}.
\end{equation*}
If $u\notin E_s$, then
\[
u^\top Qu
>
\lambda_{\max}(Q)s^2.
\]
Since
\[
u^\top Qu
\leq
\lambda_{\max}(Q)\|u\|_2^2,
\]
it follows that $\|u\|_2>s$. Therefore,
\begin{equation}
\dot W(u)
\leq
-s(s-2K_Q)
<0,
\qquad
u\notin E_s.
\label{eq:projected-uniform-decrease}
\end{equation}
The same inequality holds on $\partial E_s$, where
$\|u\|_2\geq s$. Thus, $E_s$ is positively invariant.

Moreover, every projected solution enters $E_s$ in finite time. To
see this, define the entrance time
\[
\tau_s
:=
\inf\left\{
t\geq0:u(t)\in E_s
\right\}.
\]
If $u(0)\in E_s$, then $\tau_s=0$. Suppose that
$u(0)\notin E_s$. For every $t<\tau_s$, the solution remains outside
$E_s$, and hence \eqref{eq:projected-uniform-decrease} gives
\[
\dot W(u(t))
\leq
-s(s-2K_Q).
\]
Integrating over $[0,t]$ yields
\begin{equation}
W(u(t))
\leq
W(u(0))
-
s(s-2K_Q)t,
\qquad
0\leq t<\tau_s.
\label{eq:projected-W-linear-decrease}
\end{equation}
On the other hand, since $u(t)\notin E_s$ for $t<\tau_s$,
\[
W(u(t))
>
\lambda_{\max}(Q)s^2.
\]
Combining this inequality with
\eqref{eq:projected-W-linear-decrease} gives
\[
t
<
\frac{
W(u(0))-\lambda_{\max}(Q)s^2
}{
s(s-2K_Q)
},
\qquad
0\leq t<\tau_s.
\]
Therefore,
\begin{equation*}
\tau_s
\leq
\frac{
W(u(0))-\lambda_{\max}(Q)s^2
}{
s(s-2K_Q)
}
<\infty.
\end{equation*}
Furthermore, for every $u(0)\in B(R)$,
\[
W(u(0))
\leq
\lambda_{\max}(Q)R^2.
\]
Hence,
\begin{equation*}
\tau_s
\leq
\frac{
\lambda_{\max}(Q)(R^2-s^2)
}{
s(s-2K_Q)
},
\end{equation*}
so the entrance time into $E_s$ is bounded uniformly over all initial
conditions in $B(R)$.

We next show that $E_s$ lies strictly inside the constraint set. If
$u\in E_s$, then
\[
\lambda_{\min}(Q)\|u\|_2^2
\leq
u^\top Qu
\leq
\lambda_{\max}(Q)s^2.
\]
Hence,
\[
\|u\|_2
\leq
s
\sqrt{
\frac{\lambda_{\max}(Q)}
     {\lambda_{\min}(Q)}
}
<R,
\]
where the final inequality follows from
\eqref{eq:projected-choice-s}. Therefore,
\begin{equation*}
E_s
\subset
\operatorname{int}B(R).
\end{equation*}

After a projected solution enters $E_s$, it remains in $E_s$ and
hence in the interior of $B(R)$. The projection term then vanishes,
and the solution follows the unprojected ODE
\eqref{eq:abstract-nonlinear-ode}. By the assumed global asymptotic
stability of the unprojected system, the solution converges to
$u^*$. Thus, $u^*$ is globally attractive for the projected ODE
relative to $B(R)$.

It remains to verify Lyapunov stability. Since
$u^*\in\operatorname{int}B(R)$, there exists a neighborhood of
$u^*$ contained in $\operatorname{int}B(R)$. By Lyapunov stability
of $u^*$ for the unprojected ODE, every solution starting
sufficiently close to $u^*$ remains in this neighborhood. The
projection term therefore remains zero, and the projected solution
coincides with the corresponding unprojected solution. Hence, $u^*$
is Lyapunov stable for the projected ODE. Together with global
attractivity, this proves that $u^*$ is globally asymptotically
stable relative to $B(R)$. It is the unique equilibrium in $B(R)$,
since any other equilibrium would generate a constant solution that
does not converge to $u^*$.

It remains to identify the limit set. The entrance time into $E_s$
is uniform over all initial conditions in $B(R)$, as shown above.
Moreover, convergence of the unprojected flow to $u^*$ is uniform on
the compact set $E_s$. It follows that convergence of the projected
solutions to $u^*$ is uniform over all initial conditions in $B(R)$.

Following \citep[Section~3.1]{yu2016weak}, the limit set of the
projected ODE associated with $B(R)$ is
\begin{equation}
L_{B(R)}
:=
\bigcap_{\bar t>0}
\overline{
\bigcup_{u(0)\in B(R)}
\left\{
u(t):t\geq\bar t
\right\}
},
\label{eq:abstract-projected-limit-set-definition}
\end{equation}
where the union is taken over all projected solutions with initial
conditions in $B(R)$, and the overline denotes closure.

For every $\delta>0$, uniform convergence gives a time
$\bar t_\delta<\infty$, independent of the initial condition and the
choice of solution, such that
\[
\|u(t)-u^*\|_2<\delta,
\qquad
t\geq\bar t_\delta,
\]
for every projected solution starting in $B(R)$. Consequently,
\[
\bigcup_{u(0)\in B(R)}
\left\{
u(t):t\geq\bar t_\delta
\right\}
\subset
N_\delta(u^*).
\]
It follows from
\eqref{eq:abstract-projected-limit-set-definition} that
\[
L_{B(R)}
\subseteq
\overline{N_\delta(u^*)}.
\]
Since $\delta>0$ is arbitrary,
\[
L_{B(R)}
\subseteq
\bigcap_{\delta>0}
\overline{N_\delta(u^*)}
=
\{u^*\}.
\]

Conversely, $u(t)\equiv u^*$ is a solution of the projected ODE.
Therefore, for every $\bar t>0$,
\[
u^*
\in
\bigcup_{u(0)\in B(R)}
\left\{
u(t):t\geq\bar t
\right\},
\]
and hence $u^*\in L_{B(R)}$. Combining the two inclusions gives
\[
L_{B(R)}
=
\{u^*\}.
\]
\end{proof}

\subsection{Stability Results for the Mean ODEs}
\label{app:huber-stability-results}

This subsection gives the complete stability results summarized after
Lemma~\ref{lem:huber-invariant-projection-set}. We first apply
Lemmas~\ref{lem:integral-stability-criterion} and
\ref{lem:huber-invariant-projection-set} to the unprojected and
projected mean ODEs for general eligibility traces, and then specialize
the argument to the one-step case. The resulting projection-radius
conditions are those used in
Theorem~\ref{thm:huber-general-convergence} and
Corollary~\ref{cor:huber-zero-convergence}.

\subsubsection{State-Dependent and History-Dependent Choices of
\texorpdfstring{$\lambda$}{lambda}}

\begin{proposition}[Stability for state-dependent and
history-dependent $\lambda$]
\label{prop:huber-general-trace-stability}
Suppose that Assumption~\ref{assump:lambda-conditions} holds.
Suppose also that $A_\lambda$ is positive stable, $\beta\geq0$, and
$\eta>0$.
Then $M_{\beta,\lambda}$ is Hurwitz. Suppose, in addition, that
\begin{equation}
L_\lambda
\Gamma_2
\left(
M_{\beta,\lambda},
B_\lambda,
\mathcal C_\lambda
\right)
<1.
\label{eq:huber-general-integral-gain-condition}
\end{equation}
Then the following conclusions hold.
\begin{enumerate}[label=\textnormal{(\roman*)}]
\item
The point $u_\lambda^*$ is the unique globally asymptotically stable
equilibrium of the unprojected mean ODE
\eqref{eq:huber-general-feedback-ode}.

\item
Let $Q_{\beta,\lambda}\succ0$ be the unique solution of
\begin{equation*}
M_{\beta,\lambda}^\top Q_{\beta,\lambda}
+
Q_{\beta,\lambda}M_{\beta,\lambda}
=
-I,
\end{equation*}
and define
\begin{equation*}
K_{\lambda,\tau}
:=
\left\|
Q_{\beta,\lambda}
M_{\beta,\lambda}
u_\lambda^*
\right\|_2
+
\left\|
Q_{\beta,\lambda}B_\lambda
\right\|_2
\overline H_{\lambda,\tau}
\end{equation*}
and
\begin{equation*}
R_{\lambda,\tau}
:=
\max\left\{
\|u_\lambda^*\|_2,
2K_{\lambda,\tau}
\sqrt{
\frac{\lambda_{\max}(Q_{\beta,\lambda})}
     {\lambda_{\min}(Q_{\beta,\lambda})}
}
\right\}.
\end{equation*}
For every $R>R_{\lambda,\tau}$,
$u_\lambda^*\in\operatorname{int}B(R)$ and is the unique globally
asymptotically stable equilibrium of the projected mean ODE
\begin{equation*}
\dot u
=
M_{\beta,\lambda}(u-u_\lambda^*)
+
B_\lambda
\mathcal H_{\lambda,\tau}
\left(
\mathcal C_\lambda(u-u_\lambda^*)
\right)
+
z,
\qquad
z\in-\mathcal N_{B(R)}(u),
\end{equation*}
relative to $B(R)$. Moreover, the limit set of the projected ODE associated with
\(B(R)\) satisfies
\begin{equation*}
L_{B(R)}
=
\{u_\lambda^*\}.
\end{equation*}
\end{enumerate}
\end{proposition}

\begin{proof}
Since $C+\beta I$ is positive definite and $A_\lambda$ is positive
stable, the block upper-triangular matrix $M_{\beta,\lambda}$ is
Hurwitz. Proposition~\ref{prop:huber-mean-field} gives
$\mathcal H_{\lambda,\tau}(0)=0$ and shows that
$\mathcal H_{\lambda,\tau}$ is globally Lipschitz with constant
$L_\lambda$. Lemma~\ref{lem:integral-stability-criterion} therefore
gives conclusion \textnormal{(i)}.

Proposition~\ref{prop:huber-mean-field} also gives the uniform bound
\eqref{eq:huber-general-H-uniform-bound}. Applying
Lemma~\ref{lem:huber-invariant-projection-set} with
\[
M=M_{\beta,\lambda},
\qquad
B=B_\lambda,
\qquad
\mathcal C=\mathcal C_\lambda,
\qquad
F=\mathcal H_{\lambda,\tau},
\qquad
u^*=u_\lambda^*
\]
gives conclusion \textnormal{(ii)}.
\end{proof}
\subsubsection{The One-Step Case}
\begin{corollary}[Stability in the one-step case]
\label{cor:huber-zero-trace-stability}
Suppose that Assumption~\ref{assump:lambda-conditions} holds, that
$\lambda_n\equiv0$, that $A_0$ is positive stable, and that
$\beta\geq0$ and $\eta>0$. Then $M_{\beta,0}$ is Hurwitz. Suppose,
in addition, that
\begin{equation}
\Gamma_2
\left(
M_{\beta,0},
B_0,
\mathcal C_0
\right)
<1.
\label{eq:huber-zero-integral-gain-condition}
\end{equation}
Then the following conclusions hold.
\begin{enumerate}[label=\textnormal{(\roman*)}]
\item
The point $u_0^*$ is the unique globally asymptotically stable
equilibrium of the unprojected one-step mean ODE
\eqref{eq:huber-zero-feedback-ode}.

\item
Let $Q_{\beta,0}\succ0$ be the unique solution of
\begin{equation*}
M_{\beta,0}^\top Q_{\beta,0}
+
Q_{\beta,0}M_{\beta,0}
=
-I,
\end{equation*}
and define
\begin{equation*}
K_{0,\tau}
:=
\left\|
Q_{\beta,0}M_{\beta,0}u_0^*
\right\|_2
+
\left\|
Q_{\beta,0}B_0
\right\|_2
\tau\sqrt{|\mathcal S|}
\end{equation*}
and
\begin{equation*}
R_{0,\tau}
:=
\max\left\{
\|u_0^*\|_2,
2K_{0,\tau}
\sqrt{
\frac{\lambda_{\max}(Q_{\beta,0})}
     {\lambda_{\min}(Q_{\beta,0})}
}
\right\}.
\end{equation*}
For every $R>R_{0,\tau}$,
$u_0^*\in\operatorname{int}B(R)$ and is the unique globally
asymptotically stable equilibrium of the projected one-step mean ODE
\begin{equation*}
\dot u
=
M_{\beta,0}(u-u_0^*)
+
B_0
\Pi_{[-\tau,\tau]^{|\mathcal S|}}
\left(
\mathcal C_0(u-u_0^*)
\right)
+
z,
\qquad
z\in-\mathcal N_{B(R)}(u),
\end{equation*}
relative to $B(R)$. Moreover, the limit set of the projected ODE associated with
\(B(R)\) satisfies
\begin{equation}
L_{B(R)}
=
\{u_0^*\}.
\label{eq:huber-zero-projected-limit-set}
\end{equation}
\end{enumerate}
\end{corollary}

\begin{proof}
The matrix $M_{\beta,0}$ is Hurwitz because it is block upper
triangular, $C+\beta I$ is positive definite, and $A_0$ is positive
stable. The componentwise clipping function fixes the origin and is
globally $1$-Lipschitz. Lemma~\ref{lem:integral-stability-criterion}
therefore gives conclusion \textnormal{(i)}.

Moreover,
\begin{equation*}
\left\|
\Pi_{[-\tau,\tau]^{|\mathcal S|}}(v)
\right\|_2
\leq
\tau\sqrt{|\mathcal S|},
\qquad
v\in\mathbb R^{|\mathcal S|}.
\end{equation*}
Applying Lemma~\ref{lem:huber-invariant-projection-set} with
\[
M=M_{\beta,0},
\qquad
B=B_0,
\qquad
\mathcal C=\mathcal C_0,
\qquad
F=\Pi_{[-\tau,\tau]^{|\mathcal S|}},
\qquad
u^*=u_0^*
\]
gives conclusion \textnormal{(ii)}.
\end{proof}

The preceding results establish stability of both the unprojected
and projected mean ODEs under the integral small-gain conditions
\eqref{eq:huber-general-integral-gain-condition} and
\eqref{eq:huber-zero-integral-gain-condition}. These conditions can
be checked by evaluating the corresponding integral gains
numerically. We next derive analytic upper bounds on these gains and
obtain explicit sufficient conditions in terms of the regularization
parameter $\beta$.

\subsection{Derivations of the Computable Small-Gain Conditions}
\label{app:huber-computable-conditions}

The main text states three computable certificates for the integral
small-gain conditions: condition~\textnormal{(ii)} of
Theorem~\ref{thm:huber-general-convergence} for general eligibility
traces and conditions~\textnormal{(ii)} and
\textnormal{(iii)} of Corollary~\ref{cor:huber-zero-convergence} for
the one-step case. We derive these certificates here.

\subsubsection{State-Dependent and History-Dependent Choices of
\texorpdfstring{$\lambda$}{lambda}}

We first derive the integral-gain bound used in
condition~\textnormal{(ii)} of
Theorem~\ref{thm:huber-general-convergence}. Recall from
Appendix~\ref{app:computable-Llambda-bound} that
$L_\lambda\leq\overline L_\lambda$.

\begin{proposition}[A condition for state-dependent and
history-dependent $\lambda$]
\label{prop:huber-general-explicit-condition}
Suppose that Assumption~\ref{assump:lambda-conditions} holds.
Suppose also that $A_\lambda$ is positive stable. Let $\eta>0$ and
let $\beta\geq0$
satisfy
\begin{equation}
\beta
>
2\overline L_\lambda
\lambda_{\max}(P_{A,\lambda})
\sqrt{\kappa_2(P_{A,\lambda})}
\|A_\lambda\|_2
-
\lambda_{\min}(C).
\label{eq:huber-general-explicit-beta-condition}
\end{equation}
Then $M_{\beta,\lambda}$ is Hurwitz and
\begin{equation}
L_\lambda
\Gamma_2
\left(
M_{\beta,\lambda},
B_\lambda,
\mathcal C_\lambda
\right)
<1.
\label{eq:huber-general-explicit-gain-condition}
\end{equation}
Consequently, all conclusions of
Proposition~\ref{prop:huber-general-trace-stability} hold.
\end{proposition}

\begin{proof}
Set
\begin{equation}
q_\beta
:=
\lambda_{\min}(C)+\beta.
\label{eq:q-beta}
\end{equation}
Since $C\succ0$ and $\beta\geq0$, we have $q_\beta>0$. Moreover,
\begin{equation}
\left\|
e^{-\eta(C+\beta I)t}
\right\|_2
=
e^{-\eta q_\beta t},
\qquad
t\geq0.
\label{eq:Cbeta-exponential-bound}
\end{equation}
Since $A_\lambda$ is positive stable, both diagonal blocks of
$M_{\beta,\lambda}$ are Hurwitz. Hence,
$M_{\beta,\lambda}$ is Hurwitz.

Let $P_{A,\lambda}\succ0$ be the solution of
\eqref{eq:huber-main-Alambda-lyapunov-equation}. Consider the
Lyapunov function
\[
V(y)
:=
y^\top P_{A,\lambda}y
\]
for the linear system
\[
\dot y=-A_\lambda y.
\]
The eigenvalue bounds for $P_{A,\lambda}$ give
\begin{equation}
\lambda_{\min}(P_{A,\lambda})\|y\|_2^2
\leq
V(y)
\leq
\lambda_{\max}(P_{A,\lambda})\|y\|_2^2.
\label{eq:Alambda-lyapunov-eigenvalue-bounds}
\end{equation}
Moreover,
\begin{align*}
\dot V(y)
&=
-y^\top
\left(
A_\lambda^\top P_{A,\lambda}
+
P_{A,\lambda}A_\lambda
\right)y
=
-\|y\|_2^2 \leq
-\frac{1}{\lambda_{\max}(P_{A,\lambda})}
V(y).
\end{align*}
Therefore,
\[
V(y(t))
\leq
\exp\left(
-\frac{t}{\lambda_{\max}(P_{A,\lambda})}
\right)
V(y(0)).
\]
Using \eqref{eq:Alambda-lyapunov-eigenvalue-bounds}, we obtain
\begin{align}
\|y(t)\|_2
&\leq
\sqrt{
\frac{
\lambda_{\max}(P_{A,\lambda})
}{
\lambda_{\min}(P_{A,\lambda})
}
}
\exp\left(
-\frac{t}{2\lambda_{\max}(P_{A,\lambda})}
\right)
\|y(0)\|_2.
\label{eq:Alambda-state-exponential-bound}
\end{align}

Since
\[
y(t)
=
e^{-A_\lambda t}y(0),
\]
\eqref{eq:Alambda-state-exponential-bound} implies that, for every
$y(0)\neq0$,
\begin{align}
\frac{
\left\|
e^{-A_\lambda t}y(0)
\right\|_2
}{
\|y(0)\|_2
}
&\leq
\sqrt{
\frac{
\lambda_{\max}(P_{A,\lambda})
}{
\lambda_{\min}(P_{A,\lambda})
}
}
\exp\left(
-\frac{t}{2\lambda_{\max}(P_{A,\lambda})}
\right).
\label{eq:Alambda-operator-ratio-bound}
\end{align}
By the definition of the induced matrix $2$-norm,
\[
\left\|
e^{-A_\lambda t}
\right\|_2
=
\sup_{y(0)\neq0}
\frac{
\left\|
e^{-A_\lambda t}y(0)
\right\|_2
}{
\|y(0)\|_2
}.
\]
Since the right-hand side of
\eqref{eq:Alambda-operator-ratio-bound} does not depend on $y(0)$,
taking the supremum over all $y(0)\neq0$ gives
\begin{equation}
\left\|
e^{-A_\lambda t}
\right\|_2
\leq
K_{A,\lambda}
e^{-\alpha_{A,\lambda}t},
\qquad
t\geq0,
\label{eq:Alambda-exponential-bound}
\end{equation}
where
\begin{equation}
K_{A,\lambda}
:=
\sqrt{
\frac{
\lambda_{\max}(P_{A,\lambda})
}{
\lambda_{\min}(P_{A,\lambda})
}
}
=
\sqrt{\kappa_2(P_{A,\lambda})},
\qquad
\alpha_{A,\lambda}
:=
\frac{1}{2\lambda_{\max}(P_{A,\lambda})}.
\label{eq:Alambda-decay-constants}
\end{equation}

The block upper-triangular structure of
$M_{\beta,\lambda}$ and the definitions of $B_\lambda$ and
$\mathcal C_\lambda$ give
\begin{equation*}
\mathcal C_\lambda
e^{M_{\beta,\lambda}t}
B_\lambda
=
\eta
\int_0^t
e^{-\eta(C+\beta I)(t-s)}
A_\lambda
e^{-A_\lambda s}\,ds.
\end{equation*}
Using \eqref{eq:Cbeta-exponential-bound} and
\eqref{eq:Alambda-exponential-bound}, we obtain
\begin{align*}
&
\left\|
\mathcal C_\lambda
e^{M_{\beta,\lambda}t}
B_\lambda
\right\|_2
\leq
\eta
K_{A,\lambda}
\|A_\lambda\|_2
\int_0^t
e^{-\eta q_\beta(t-s)}
e^{-\alpha_{A,\lambda}s}\,ds.
\end{align*}
Integrating over $t\in[0,\infty)$ and applying Tonelli's theorem
gives
\begin{align}
&
\Gamma_2
\left(
M_{\beta,\lambda},
B_\lambda,
\mathcal C_\lambda
\right)
\notag\\
&\quad\leq
\eta
K_{A,\lambda}
\|A_\lambda\|_2
\int_0^\infty
\int_0^t
e^{-\eta q_\beta(t-s)}
e^{-\alpha_{A,\lambda}s}
\,ds\,dt
\notag\\
&\quad=
\eta
K_{A,\lambda}
\|A_\lambda\|_2
\int_0^\infty
e^{-\alpha_{A,\lambda}s}
\left(
\int_s^\infty
e^{-\eta q_\beta(t-s)}\,dt
\right)ds
\notag\\
&\quad=
\frac{
K_{A,\lambda}\|A_\lambda\|_2
}{
\alpha_{A,\lambda}q_\beta
}.
\label{eq:huber-general-integral-gain-bound}
\end{align}
Since $L_\lambda\leq\overline L_\lambda$, substituting
\eqref{eq:q-beta} and \eqref{eq:Alambda-decay-constants} gives
\begin{align}
&
L_\lambda
\Gamma_2
\left(
M_{\beta,\lambda},
B_\lambda,
\mathcal C_\lambda
\right)
\leq
\frac{
2\overline L_\lambda
\lambda_{\max}(P_{A,\lambda})
\sqrt{\kappa_2(P_{A,\lambda})}
\|A_\lambda\|_2
}{
\lambda_{\min}(C)+\beta
}.
\label{eq:huber-general-gain-final-bound}
\end{align}
Condition~\eqref{com_con_1} makes the right-hand side strictly
smaller than one and therefore implies
\eqref{eq:huber-main-general-gain-condition}, as asserted in
condition~\textnormal{(ii)} of
Theorem~\ref{thm:huber-general-convergence}.

\end{proof}

\subsubsection{The One-Step Case}

When $\lambda_n\equiv0$, the explicit feedback representation
$\mathcal H_{0,\tau}(x)
=D_0\Pi_{[-\tau,\tau]^{|\mathcal S|}}(\Phi x)$ yields two useful
bounds on the integral gain. The first uses the Lyapunov equation in
condition~\textnormal{(ii)} of
Corollary~\ref{cor:huber-zero-convergence}; the second uses the
positive definiteness assumption in condition~\textnormal{(iii)}.

\paragraph{Lyapunov-based bound.}

\begin{corollary}[A numerically verifiable Lyapunov condition]
\label{cor:huber-zero-lyapunov-condition}
Suppose that Assumption~\ref{assump:lambda-conditions} holds.
Suppose also that $\lambda_n\equiv0$ and that $A_0$ is positive
stable. Let
$P_A\succ0$ be the unique solution of
\begin{equation}
A_0^\top P_A+P_AA_0=I,
\label{eq:A0-lyapunov-equation}
\end{equation}
and define
\begin{equation*}
\kappa_2(P_A)
:=
\frac{
\lambda_{\max}(P_A)
}{
\lambda_{\min}(P_A)
}.
\end{equation*}
Let $\eta>0$ and let $\beta\geq0$ satisfy
\begin{equation}
\beta
>
2\lambda_{\max}(P_A)
\sqrt{\kappa_2(P_A)}
\|\Phi\|_2
\|A_0\|_2
\|D_0\|_2
-
\lambda_{\min}(C).
\label{eq:huber-zero-lyapunov-beta-condition}
\end{equation}
Then $M_{\beta,0}$ is Hurwitz and
\begin{equation}
\Gamma_2
\left(
M_{\beta,0},
B_0,
\mathcal C_0
\right)
<1.
\label{eq:huber-zero-lyapunov-gain-condition}
\end{equation}
Consequently, all conclusions of
Corollary~\ref{cor:huber-zero-trace-stability} hold.
\end{corollary}

\begin{proof}
The calculation leading to
\eqref{eq:Alambda-exponential-bound}, specialized to $\lambda=0$,
gives
\begin{equation}
\|e^{-A_0t}\|_2
\leq
K_Ae^{-\alpha_At},
\qquad
t\geq0,
\label{eq:A0-lyapunov-exponential-bound}
\end{equation}
where
\begin{equation*}
K_A
:=
\sqrt{\kappa_2(P_A)},
\qquad
\alpha_A
:=
\frac{1}{2\lambda_{\max}(P_A)}.
\end{equation*}

Set
\[
q_\beta
:=
\lambda_{\min}(C)+\beta.
\]
Since $C\succ0$ and $\beta\geq0$, we have $q_\beta>0$. Therefore,
\[
\left\|
e^{-\eta(C+\beta I)t}
\right\|_2
=
e^{-\eta q_\beta t},
\qquad
t\geq0.
\]
Moreover, $A_0$ is positive stable, so both diagonal blocks of
$M_{\beta,0}$ are Hurwitz. Hence, $M_{\beta,0}$ is Hurwitz.

The block matrix-exponential formula gives
\begin{equation}
\mathcal C_0
e^{M_{\beta,0}t}
B_0
=
\eta\Phi
\int_0^t
e^{-\eta(C+\beta I)(t-s)}
A_0e^{-A_0s}D_0\,ds.
\label{eq:huber-zero-block-representation}
\end{equation}
Using 
\eqref{eq:A0-lyapunov-exponential-bound}, we obtain
\begin{align*}
&
\left\|
\mathcal C_0
e^{M_{\beta,0}t}
B_0
\right\|_2
\leq
\eta
\|\Phi\|_2
K_A
\|A_0\|_2
\|D_0\|_2
\int_0^t
e^{-\eta q_\beta(t-s)}
e^{-\alpha_As}\,ds.
\end{align*}
Applying the same integration calculation used in deriving
\eqref{eq:huber-general-integral-gain-bound} gives
\begin{align}
&
\Gamma_2
\left(
M_{\beta,0},
B_0,
\mathcal C_0
\right)
\leq
\frac{
\|\Phi\|_2
K_A
\|A_0\|_2
\|D_0\|_2
}{
\alpha_Aq_\beta
}
=
\frac{
2\lambda_{\max}(P_A)
\sqrt{\kappa_2(P_A)}
\|\Phi\|_2
\|A_0\|_2
\|D_0\|_2
}{
\lambda_{\min}(C)+\beta
}.
\label{eq:huber-zero-lyapunov-integral-bound}
\end{align}
Condition~\eqref{com_con_2} makes the right-hand side strictly
smaller than one. Thus, condition~\textnormal{(ii)} implies the
integral small-gain condition
\eqref{eq:huber-main-zero-gain-condition}.
\end{proof}

\paragraph{Direct bound.}

\begin{corollary}[A directly verifiable condition]
\label{cor:huber-zero-direct-condition}
Suppose that Assumption~\ref{assump:lambda-conditions} holds and that
$\lambda_n\equiv0$. Define
\begin{equation*}
H_0
:=
\frac{A_0+A_0^\top}{2},
\end{equation*}
and suppose that
\begin{equation*}
H_0\succ0.
\end{equation*}
Let $\eta>0$ and let $\beta\geq0$ satisfy
\begin{equation}
\beta
>
\frac{
\|\Phi\|_2
\|A_0\|_2
\|D_0\|_2
}{
\lambda_{\min}(H_0)
}
-
\lambda_{\min}(C).
\label{eq:huber-zero-direct-beta-condition}
\end{equation}
Then $M_{\beta,0}$ is Hurwitz and
\begin{equation}
\Gamma_2
\left(
M_{\beta,0},
B_0,
\mathcal C_0
\right)
<1.
\label{eq:huber-zero-direct-gain-condition}
\end{equation}
Consequently, all conclusions of
Corollary~\ref{cor:huber-zero-trace-stability} hold.
\end{corollary}

\begin{proof}
Consider the linear system
\[
\dot y(t)
=
-A_0y(t).
\]
Along every solution,
\begin{align*}
\frac{d}{dt}\|y(t)\|_2^2
&=
-y(t)^\top
\left(
A_0+A_0^\top
\right)y(t)
\notag\\
&=
-2y(t)^\top H_0y(t)
\notag\\
&\leq
-2\lambda_{\min}(H_0)
\|y(t)\|_2^2.
\end{align*}
Therefore,
\[
\|y(t)\|_2
\leq
e^{-\lambda_{\min}(H_0)t}
\|y(0)\|_2,
\qquad
t\geq0.
\]
Since $y(t)=e^{-A_0t}y(0)$ and the estimate holds for every initial
condition, the definition of the induced matrix $2$-norm gives
\begin{equation}
\|e^{-A_0t}\|_2
\leq
e^{-\lambda_{\min}(H_0)t},
\qquad
t\geq0.
\label{eq:A0-direct-exponential-bound}
\end{equation}
In particular, $-A_0$ is Hurwitz.

Set
\[
q_\beta
:=
\lambda_{\min}(C)+\beta.
\]
Since $C\succ0$ and $\beta\geq0$, we have $q_\beta>0$, and
\[
\left\|
e^{-\eta(C+\beta I)t}
\right\|_2
=
e^{-\eta q_\beta t},
\qquad
t\geq0.
\]
Thus, both diagonal blocks of $M_{\beta,0}$ are Hurwitz, and hence
$M_{\beta,0}$ is Hurwitz.

Applying \eqref{eq:A0-direct-exponential-bound} to
\eqref{eq:huber-zero-block-representation} gives
\begin{align*}
&
\left\|
\mathcal C_0
e^{M_{\beta,0}t}
B_0
\right\|_2
\leq
\eta
\|\Phi\|_2
\|A_0\|_2
\|D_0\|_2
\int_0^t
e^{-\eta q_\beta(t-s)}
e^{-\lambda_{\min}(H_0)s}\,ds.
\end{align*}
Using the same integration calculation as above, we obtain
\begin{align}
&
\Gamma_2
\left(
M_{\beta,0},
B_0,
\mathcal C_0
\right)
\leq
\frac{
\|\Phi\|_2
\|A_0\|_2
\|D_0\|_2
}{
\lambda_{\min}(H_0)q_\beta
}=
\frac{
\|\Phi\|_2
\|A_0\|_2
\|D_0\|_2
}{
\lambda_{\min}(H_0)
\left(
\lambda_{\min}(C)+\beta
\right)
}.
\label{eq:huber-zero-direct-integral-bound}
\end{align}
Condition~\eqref{com_con_3} makes the right-hand side strictly
smaller than one. Thus, condition~\textnormal{(iii)} also implies the
integral small-gain condition
\eqref{eq:huber-main-zero-gain-condition}.
\end{proof}

\subsection{Verification of the Stochastic Approximation Conditions
for \texorpdfstring{TDRC($\lambda$)-Huber}{TDRC(lambda)-Huber}}
\label{app:huber-sa-verification}

This subsection states and verifies the four stochastic approximation
conditions required by the weak-convergence theorem. Following
Section~\ref{sec:huber-main-convergence-results}, recall the notation
\[
u=
\begin{bmatrix}
x\\
\theta
\end{bmatrix},
\qquad
F_\tau(u,z)
=
\begin{bmatrix}
h(\theta,x,z)\\
g_\tau(\theta,x,z)
\end{bmatrix},
\qquad
\varepsilon_{n+1}
=
\begin{bmatrix}
\eta e_n\omega_{n+1}\\
e_n\omega_{n+1}
\end{bmatrix}.
\]

Set
\begin{align}
\bar F_\tau(u)
&:=\mathbb E_\zeta[F_\tau(u,Z_0)]=
\begin{pmatrix}
\bar h(\theta,x)\\
\bar g_\tau(\theta,x)
\end{pmatrix}=\begin{pmatrix}
-\eta(C+\beta I)x-\eta A_\lambda\theta+\eta b_\lambda\\
-A_\lambda\theta+b_\lambda-\mathcal H_{\lambda,\tau}(x)
\end{pmatrix}.
\label{eq:huber-F-and-Fbar}
\end{align}

Following the weak-convergence approach to stochastic approximation
\citep{kushner2003}, we verify the following conditions to show that
the projected ODE
\[
\dot u
=
\bar F_\tau(u)+z,
\qquad
z\in-\mathcal N_{B(R)}(u),
\]
is the mean ODE associated with the constrained
TDRC($\lambda$)-Huber recursion \eqref{eq:tdrchuber-sa-form}.
\begin{enumerate}
\item
The following families of random variables are uniformly integrable:
\[
\{F_\tau(U_n,Z_n)\}_{n\geq0},
\qquad
\{F_\tau(u,Z_n):n\geq0,\ u\in B(R)\},
\]
and
\[
\{F_\tau(U_n,Z_n)+\varepsilon_{n+1}\}_{n\geq0}.
\]

\item
The family \(\{Z_n\}_{n\geq0}\) is tight.

\item
For every compact set \(D\subset\mathcal Z\), the function
\(F_\tau(u,z)\) is continuous in \(u\), uniformly over \(z\in D\).

\item
The mean field \(\bar F_\tau:B(R)\to\mathbb R^{2d}\) is continuous
and, for every fixed \(u\in B(R)\) and every compact set
\(D\subset\mathcal Z\), the conditional averaging property in
Proposition~\ref{prop:huber-condition-4} holds.
\end{enumerate}

The verification of these four conditions is given in turn.

Since \(\lambda_n\equiv0\) is included in this setting, the
preceding verification also applies to
TDRC(0)-Huber.  Therefore, the four stochastic
approximation conditions need not be verified separately
for \(\lambda=0\).

\subsubsection{Verification of Condition 1}

We first record the trace-continuity property needed for uniform
integrability.

\begin{lemma}
\label{lem:huber-trace-lipschitz}
Let \(u=(x^\top,\theta^\top)^\top\). Under
Assumption~\ref{assump:lambda-conditions}, for each fixed \(u\),
\(g_\tau(\theta,x,\cdot)\) is Lipschitz continuous in the trace
variable \(e\). Moreover, the Lipschitz constant can be chosen
uniformly over \(u\in B(R)\).
\end{lemma}

\begin{proof}

In the state-dependent case, \(\lambda'=\lambda(s')\)
does not depend on \(e\). Hence, 
\begin{align}
&\left|
(1-\lambda(s'))
\left[
\operatorname{clip}_\tau(e^\top x)
-
\operatorname{clip}_\tau((e')^\top x)
\right]
\right|
\le
\left|
e^\top x-(e')^\top x
\right|
\le
\|x\|_2\|e-e'\|_2.
\label{eq:state-dependent-trace-bound}
\end{align}

We next consider the history-dependent case. Let
\(y'=f(y,s')\) be the next memory state. By
Condition~\ref{cond:history-dependent-lambda}\textnormal{(i)}, together with
equivalence of norms, there exists \(K<\infty\) such that
\begin{align*}
\left\|
\lambda(y',e)e-\lambda(y',e')e'
\right\|_2
\le
K\|e-e'\|_2.
\end{align*}
It follows that
\begin{align}
&\left\|
(1-\lambda(y',e))e
-
(1-\lambda(y',e'))e'
\right\|_2
\notag\\
&\quad\le
\|e-e'\|_2
+
\left\|
\lambda(y',e)e-\lambda(y',e')e'
\right\|_2
\notag\\
&\quad\le
(K+1)\|e-e'\|_2.
\label{eq:one-minus-lambda-trace-bound}
\end{align}

Recall that
\begin{align}
\operatorname{clip}_\tau(a)
=
a\min\left\{1,\frac{\tau}{|a|}\right\},
\label{eq:clip-multiplier-representation}
\end{align}
where the multiplier is taken to be \(1\) at \(a=0\).
For arbitrary \(a,b\in\mathbb R\),
\begin{align}
&|b|
\left|
\min\left\{1,\frac{\tau}{|a|}\right\}
-
\min\left\{1,\frac{\tau}{|b|}\right\}
\right|
\notag\\
&\quad=
\left|
b\min\left\{1,\frac{\tau}{|a|}\right\}
-
b\min\left\{1,\frac{\tau}{|b|}\right\}
\right|
\notag\\
&\quad=
\left|
(b-a)\min\left\{1,\frac{\tau}{|a|}\right\}
+
a\min\left\{1,\frac{\tau}{|a|}\right\}
-
b\min\left\{1,\frac{\tau}{|b|}\right\}
\right|
\notag\\
&\quad=
\left|
(b-a)\min\left\{1,\frac{\tau}{|a|}\right\}
+
\operatorname{clip}_\tau(a)
-
\operatorname{clip}_\tau(b)
\right|
\notag\\
&\quad\le
|b-a|
+
\left|
\operatorname{clip}_\tau(a)
-
\operatorname{clip}_\tau(b)
\right|
\notag\\
&\quad\le
2|a-b|.
\label{eq:clip-multiplier-bound}
\end{align}

Using \eqref{eq:clip-multiplier-representation}, we have
\begin{align*}
&(1-\lambda(y',e))
\operatorname{clip}_\tau(e^\top x)
=
\bigl[(1-\lambda(y',e))e\bigr]^\top x
\min\left\{1,\frac{\tau}{|e^\top x|}\right\}.
\end{align*}
Consequently,
\begin{align*}
&\left|
(1-\lambda(y',e))
\operatorname{clip}_\tau(e^\top x)
-
(1-\lambda(y',e'))
\operatorname{clip}_\tau((e')^\top x)
\right|
\notag\\
&\quad\le
\left\|
(1-\lambda(y',e))e
-
(1-\lambda(y',e'))e'
\right\|_2
\|x\|_2
\min\left\{1,\frac{\tau}{|e^\top x|}\right\}
\notag\\
&\qquad+
\left|
\bigl[(1-\lambda(y',e'))e'\bigr]^\top x
\right|
\left|
\min\left\{1,\frac{\tau}{|e^\top x|}\right\}
-
\min\left\{1,\frac{\tau}{|(e')^\top x|}\right\}
\right|.
\end{align*}
The first term on the right-hand side is bounded by
\[
(K+1)\|x\|_2\|e-e'\|_2
\]
by \eqref{eq:one-minus-lambda-trace-bound}. For the
second term, since \(0\le1-\lambda(y',e')\le1\), we have
\begin{align}
\left|
\bigl[(1-\lambda(y',e'))e'\bigr]^\top x
\right|
\le
|(e')^\top x|.
\label{eq:history-one-minus-lambda-inner-product-bound}
\end{align}
Applying \eqref{eq:history-one-minus-lambda-inner-product-bound} and
\eqref{eq:clip-multiplier-bound} with
\(a=e^\top x\) and \(b=(e')^\top x\) therefore gives
\begin{align*}
&\left|
\bigl[(1-\lambda(y',e'))e'\bigr]^\top x
\right|
\left|
\min\left\{1,\frac{\tau}{|e^\top x|}\right\}
-
\min\left\{1,\frac{\tau}{|(e')^\top x|}\right\}
\right|
\notag\\
&\qquad\le
|(e')^\top x|\left|
\min\left\{1,\frac{\tau}{|e^\top x|}\right\}
-
\min\left\{1,\frac{\tau}{|(e')^\top x|}\right\}
\right|
\notag\\
&\qquad\le
2\left|
(e-e')^\top x
\right|\le
2\|x\|_2\|e-e'\|_2.
\end{align*}
Combining these two estimates yields
\begin{align*}
&\left|
(1-\lambda(y',e))
\operatorname{clip}_\tau(e^\top x)
-
(1-\lambda(y',e'))
\operatorname{clip}_\tau((e')^\top x)
\right|
\notag\\
&\qquad\le
(K+3)\|x\|_2\|e-e'\|_2.
\end{align*}

The remaining trace-dependent term in \(g_\tau\) satisfies
\begin{align*}
&\left\|
e\bar\delta(s,s',v_\theta)
-
e'\bar\delta(s,s',v_\theta)
\right\|_2
\notag\\
&\qquad=
\left|
\bar\delta(s,s',v_\theta)
\right|
\|e-e'\|_2.
\end{align*}
Thus, in the history-dependent case,
\begin{align}
&\|g_\tau(\theta,x,z)-g_\tau(\theta,x,z')\|_2
\notag\\
&\quad\le
\Bigl(
|\bar\delta(s,s',v_\theta)|
+
\rho(s,s')\gamma(s')\|\phi(s')\|_2
(K+3)\|x\|_2
\Bigr)
\|e-e'\|_2.
\label{eq:gr-trace-lipschitz-bound}
\end{align}
The state-dependent case satisfies the analogous bound by
\eqref{eq:state-dependent-trace-bound}.

Finally, \(B(R)\) is bounded, and the state space, feature
vectors, mean rewards, discount factors, and importance-sampling
ratios are bounded. Hence, the coefficient in
\eqref{eq:gr-trace-lipschitz-bound} can be bounded
uniformly over \(u\in B(R)\) and the remaining components
of \(z\). This proves the result.
\end{proof}

\begin{proposition}
Under Assumptions~\ref{assump:lambda-conditions} and
\ref{assump:stepsize}, for every initial trace, the following
families of random variables are uniformly integrable:
\begin{enumerate}
\item \(\{e_n\}_{n\geq0}\);
\item \(\{F_\tau(u,Z_n):n\geq0,\ u\in B(R)\}\);
\item \(\{F_\tau(U_n,Z_n)\}_{n\geq0}\);
\item \(\{F_\tau(U_n,Z_n)+\varepsilon_{n+1}\}_{n\geq0}\).
\end{enumerate}
\end{proposition}

\begin{proof}
The argument is analogous to that of
Proposition~\ref{prop:uniform-integrability-F} and
\citep[Section~3.1.2]{Yugtd}. By
Proposition~\ref{prop:state-trace-properties}\textnormal{(iii)}, the sequence
\(\{e_n\}_{n\ge0}\) is uniformly integrable.  By
Lemma~\ref{lem:huber-trace-lipschitz},
\(g_\tau(\theta,x,z)\) is Lipschitz continuous in the trace
variable \(e\), uniformly with respect to
\((x,\theta)\in B(R)\) and the remaining components of
\(z\).

The component \(h\) is unchanged from the linear TDRC($\lambda$)
analysis in Appendix~\ref{app:tdrc-sa-verification}.
Hence, it retains the same Lipschitz continuity property
in the trace variable.  The corresponding componentwise
uniform integrability conclusions follow from:
\begin{enumerate}
\item the boundedness of \(B(R)\), and hence of the
      iterates;
\item the finiteness of the state space;
\item the uniform integrability of
      \(\{e_n\}_{n\ge0}\);
\item the Lipschitz continuity of \(g_\tau\) and \(h\) in
      the trace variable \(e\), uniformly with respect to
      \((\theta,x)\in B(R)\) and the remaining components
      of \(z\);
\item the standing finite-variance assumption on the
      reward-noise sequence \(\{\omega_{n+1}\}\).
\end{enumerate}
The argument is otherwise identical to that in
\citep[Section~3.1.2]{Yugtd}.

Finally, since
\begin{align*}
F_\tau(u,z)
=
\begin{bmatrix}
h(\theta,x,z)\\
g_\tau(\theta,x,z)
\end{bmatrix}
\end{align*}
and
\begin{align*}
F_\tau(U_n,Z_n)+\varepsilon_{n+1}
=
\begin{bmatrix}
h(\theta_n,x_n,Z_n)+\eta e_n\omega_{n+1}\\
g_\tau(\theta_n,x_n,Z_n)+e_n\omega_{n+1}
\end{bmatrix},
\end{align*}
the stated uniform integrability results follow from the
corresponding componentwise results.
\end{proof}

\subsubsection{Verification of Condition 2}

Condition~2 follows directly from
\citet[Section~3.1.2]{Yugtd}, since our definition of
$\{Z_n\}$ and the assumptions imposed on the state--trace process
are the same as those used therein.

\subsubsection{Verification of Condition 3}

\begin{proposition}
For every compact set $D\subset\mathcal Z$, the function
$F_\tau(u,z)$ is continuous in $u$, uniformly over $z\in D$.
\end{proposition}

\begin{proof}
Write $u=(x^\top,\theta^\top)^\top$ and fix a compact set
$D\subset\mathcal Z$. By the proof of
Proposition~\ref{prop:tdrc-uniform-continuity}, the functions
$g(\theta,x,z)$ and $h(\theta,x,z)$ are continuous in
$(\theta,x)$, uniformly over $z\in D$.

It remains to consider the clipping correction in $g_\tau$. Let
\[
c(z)
:=
\rho(s,s')(1-\lambda')\gamma(s')\phi(s')
\]
and define
\[
r_\tau(a)
:=
a-\operatorname{clip}_\tau(a),
\qquad
a\in\mathbb R.
\]
Then
\begin{equation}
g_\tau(\theta,x,z)
=
g(\theta,x,z)
+
c(z)r_\tau(e^\top x).
\label{eq:huber-g-from-g}
\end{equation}

The scalar function $r_\tau$ is uniformly continuous on
$\mathbb R$. Since $D$ is compact, both $c(z)$ and the trace
component $e$ are bounded on $D$. In particular, there exist
constants $C_D,E_D<\infty$ such that
\[
\|c(z)\|_2\leq C_D,
\qquad
\|e\|_2\leq E_D,
\qquad
z\in D.
\]
For every $u,u'$ and $z\in D$,
\begin{align}
&\left\|
c(z)r_\tau(e^\top x)
-
c(z)r_\tau(e^\top x')
\right\|_2
\notag\\
&\quad\leq
C_D
\left|
r_\tau(e^\top x)
-
r_\tau(e^\top x')
\right|
\notag\\
&\quad=
C_D
\left|
e^\top(x-x')
-
\left[
\operatorname{clip}_\tau(e^\top x)
-
\operatorname{clip}_\tau(e^\top x')
\right]
\right|
\notag\\
&\quad\leq
C_D|e^\top(x-x')|
+
C_D
\left|
\operatorname{clip}_\tau(e^\top x)
-
\operatorname{clip}_\tau(e^\top x')
\right|
\notag\\
&\quad\leq
2C_DE_D\|u-u'\|_2.
\label{eq:huber-correction-continuity}
\end{align}
Therefore, as $u'\to u$, the right-hand side of
\eqref{eq:huber-correction-continuity} converges to zero uniformly
over $z\in D$. Hence, the clipping correction is continuous in
$(\theta,x)$ uniformly over $z\in D$.

It follows from \eqref{eq:huber-g-from-g} and the continuity of $g$
in $(\theta,x)$ uniformly over $z\in D$ that $g_\tau$ has the same
property. Since $h$ also has this property,
\[
F_\tau(u,z)
=
\begin{pmatrix}
h(\theta,x,z)\\
g_\tau(\theta,x,z)
\end{pmatrix}
\]
is continuous in $u$, uniformly over $z\in D$.
\end{proof}

\subsubsection{Verification of Condition 4}

\begin{proposition}
\label{prop:huber-condition-4}
Under Assumption~\ref{assump:lambda-conditions}, the mean field
\(\bar F_\tau:B(R)\to\mathbb R^{2d}\) is continuous. Moreover, for
each fixed \(u\in B(R)\) and each compact set
\(D\subset\mathcal Z\),
\begin{align}
\lim_{m,n\to\infty}
\frac{1}{m}
\sum_{i=n}^{n+m-1}
\mathbb E_n\!\left[
F_\tau(u,Z_i)-\bar F_\tau(u)
\right]
\mathbf 1_{\{Z_n\in D\}}
=
0
\quad\text{in mean}.
\label{eq:huber-condition-4-averaging}
\end{align}
\end{proposition}

\begin{proof}
The continuity of \(\bar F_\tau\) follows from
\eqref{eq:huber-F-and-Fbar} and the global Lipschitz continuity of
\(\mathcal H_{\lambda,\tau}\) established in
Proposition~\ref{prop:huber-mean-field}.

The argument follows that of
Proposition~\ref{prop:convergence-in-mean-F}.
Since \(\mathbf 1_{\{Z_n\in D\}}\) is measurable with
respect to the information available at time \(n\),
conditional Jensen's inequality shows that it is
sufficient to prove the stated convergence with
\(\mathbb E_n\) removed.

For each fixed \(u=(x^\top,\theta^\top)^\top\in B(R)\),
the function \(h(\theta,x,\cdot)\) is Lipschitz continuous
in the trace variable by the proof of
Proposition~\ref{prop:uniform-integrability-F}, and
\(g_\tau(\theta,x,\cdot)\) has the same property by
Lemma~\ref{lem:huber-trace-lipschitz}.  Moreover, by the
definition of the mean fields,
\begin{align*}
\mathbb E_\zeta
\left[
h(\theta,x,Z_0)
\right]
&=
\bar h(\theta,x),
&
\mathbb E_\zeta
\left[
g_\tau(\theta,x,Z_0)
\right]
&=
\bar g_\tau(\theta,x).
\end{align*}

It therefore follows from
Proposition~\ref{prop:state-trace-properties}  that, for every compact
set \(D\subset\mathcal Z\),
\begin{align}
&\frac{1}{m}
\sum_{i=n}^{n+m-1}
\left[
h(\theta,x,Z_i)-\bar h(\theta,x)
\right]
\mathbf 1_{\{Z_n\in D\}}
\longrightarrow0
\quad\text{in mean},
\label{eq:huber-h-averaging}
\\
&\frac{1}{m}
\sum_{i=n}^{n+m-1}
\left[
g_\tau(\theta,x,Z_i)-\bar g_\tau(\theta,x)
\right]
\mathbf 1_{\{Z_n\in D\}}
\longrightarrow0
\quad\text{in mean}.
\label{eq:huber-gr-averaging}
\end{align}

Finally,
combining
\eqref{eq:huber-h-averaging} and
\eqref{eq:huber-gr-averaging},
\[
\frac{1}{m}
\sum_{i=n}^{n+m-1}
\bigl[
F_\tau(u,Z_i)-\bar F_\tau(u)
\bigr]
\mathbf 1_{\{Z_n\in D\}}
\longrightarrow 0
\quad\text{in mean}
\]
as $m,n\to\infty$. Conditional Jensen's inequality then gives
\eqref{eq:huber-condition-4-averaging}.
\end{proof}

\subsection{Proofs of the Main Convergence Results}\label{app:huber-main-proofs}

\begin{proof}[Proof of
Theorem~\ref{thm:huber-general-convergence}]
Under condition~\textnormal{(i)},
\eqref{eq:huber-main-general-gain-condition} holds directly. Under
condition~\textnormal{(ii)}, the derivation in
Appendix~\ref{app:huber-computable-conditions} shows that
\eqref{com_con_1} implies
\eqref{eq:huber-main-general-gain-condition}. Thus, the required
integral small-gain condition holds in either case.

Appendix~\ref{app:huber-sa-verification} states and verifies the
hypotheses of the weak-convergence theorem for constrained stochastic
approximation \citep[Theorem~8.2.3]{kushner2003}. The projected mean
ODE is
\begin{equation*}
\dot u
=
M_{\beta,\lambda}u
+q_\lambda
+B_\lambda
\mathcal H_{\lambda,\tau}(\mathcal C_\lambda u)
+z,
\qquad
z\in-\mathcal N_{B(R)}(u).
\end{equation*}
The theorem therefore yields a sequence \(T_k\to\infty\) such that,
for every \(\delta>0\),
\begin{equation}
\limsup_{k\to\infty}
\mathbb P\left(
U_i\notin
N_\delta(L_{B(R)})
\text{ for some }
i\in[k,m(k,T_k)]
\right)
=
0,
\label{eq:huber-general-limit-set-window-proof}
\end{equation}
where \(L_{B(R)}\) is the limit set of the projected mean ODE. Under
\eqref{eq:huber-main-general-gain-condition},
Proposition~\ref{prop:huber-general-trace-stability} applies. Since
\(R>R_{\lambda,\tau}\), it gives
\begin{equation*}
L_{B(R)}
=
\{u_\lambda^*\}.
\end{equation*}
Substitution of this identity into
\eqref{eq:huber-general-limit-set-window-proof} gives
\eqref{eq:huber-general-window-convergence} and hence convergence in
probability. Because \(U_n\) and \(u_\lambda^*\) belong to the bounded
set \(B(R)\), \citet[Theorem~10.3.6]{Dudley_2002} gives
\eqref{eq:huber-general-mean-convergence}.
\end{proof}

\begin{proof}[Proof of
Corollary~\ref{cor:huber-zero-convergence}]
Condition \textnormal{(i)} is
\eqref{eq:huber-main-zero-gain-condition}. Condition
\textnormal{(ii)} implies
\eqref{eq:huber-main-zero-gain-condition} by the Lyapunov-based
derivation in Appendix~\ref{app:huber-computable-conditions}, while
the direct derivation there gives the same implication under
condition~\textnormal{(iii)}. Hence, under any one
of conditions \textnormal{(i)}--\textnormal{(iii)},
Corollary~\ref{cor:huber-zero-trace-stability} gives
\eqref{eq:huber-zero-projected-limit-set}.

The stochastic approximation verification in
Appendix~\ref{app:huber-sa-verification} also applies when
\(\lambda_n\equiv0\). Therefore, the same weak-convergence argument
as in the proof of
Theorem~\ref{thm:huber-general-convergence} gives the one-step
versions of \eqref{eq:huber-general-window-convergence},
\eqref{eq:huber-general-convergence-probability}, and
\eqref{eq:huber-general-mean-convergence}, with
\(u_\lambda^*\) replaced by \(u_0^*\).
\end{proof}

\section{Derivation of a Computable Bound on
\texorpdfstring{$L_\lambda$}{L-lambda}}
\label{app:computable-Llambda-bound}

This appendix derives an upper bound on the Lipschitz constant
\begin{align}
L_\lambda
:=
\mathbb E_\zeta\!\left[
\rho_0(1-\lambda_1)\gamma_1
\|\phi(S_1)\|_2\|e_0\|_2
\right].
\label{eq:appendix-Llambda-definition}
\end{align}
The bound applies to the state-dependent and history-dependent
choices of $\lambda$ covered by
Assumption~\ref{assump:lambda-conditions}.

Let
\begin{align*}
\phi_{\max}
&:=
\max_{s\in\mathcal S}\|\phi(s)\|_2,
\\
\gamma_{\max}
&:=
\max_{s\in\mathcal S}\gamma(s),
\end{align*}
and let $\mathbf 1$ denote the all-ones vector of the appropriate
dimension. Under
Condition~\ref{cond:target-behavior}\textnormal{(i)},
\begin{align*}
(I-P\Gamma)^{-1}
=
\sum_{k=0}^{\infty}(P\Gamma)^k.
\end{align*}

\begin{proposition}[A computable bound on $L_\lambda$]
\label{prop:computable-Llambda-bound}
Under Assumption~\ref{assump:lambda-conditions},
\begin{align}
L_\lambda
\leq
\overline L_\lambda,
\label{eq:appendix-Llambda-bound}
\end{align}
where
\begin{align*}
\overline L_\lambda
:=
\gamma_{\max}\phi_{\max}^2
\,\xi^\top(I-P\Gamma)^{-1}\mathbf 1.
\end{align*}
\end{proposition}

\begin{proof}
By \citet[Lemma~2.1]{Yugtd} and the argument used in its proof, on a
double-ended stationary state--trace process,
\begin{align*}
e_0
=
\sum_{k=0}^{\infty}
\left(
\prod_{j=1}^{k}
\lambda_{1-j}\gamma_{1-j}\rho_{-j}
\right)
\phi(S_{-k}).
\end{align*}
Moreover, for every $k\geq0$,
\begin{align}
\mathbb E_\zeta\!\left[
\prod_{j=1}^{k}
\gamma_{1-j}\rho_{-j}
\right]
=
\xi^\top(P\Gamma)^k\mathbf 1.
\label{eq:appendix-importance-sampling-identity}
\end{align}
Here and below, the product corresponding to $k=0$ is understood to
be $1$.

Since $0\leq\lambda_n\leq1$ and all discount factors and
importance-sampling ratios are nonnegative, the triangle inequality
gives
\begin{align}
\|e_0\|_2
&\leq
\sum_{k=0}^{\infty}
\left(
\prod_{j=1}^{k}
\lambda_{1-j}\gamma_{1-j}\rho_{-j}
\right)
\|\phi(S_{-k})\|_2
\notag\\
&\leq
\phi_{\max}
\sum_{k=0}^{\infty}
\prod_{j=1}^{k}
\gamma_{1-j}\rho_{-j}.
\label{eq:appendix-trace-pointwise-bound}
\end{align}
Taking expectations in
\eqref{eq:appendix-trace-pointwise-bound} and using Tonelli's theorem
together with
\eqref{eq:appendix-importance-sampling-identity} yields
\begin{align}
\mathbb E_\zeta[\|e_0\|_2]
&\leq
\phi_{\max}
\sum_{k=0}^{\infty}
\xi^\top(P\Gamma)^k\mathbf 1
\notag\\
&=
\phi_{\max}\,
\xi^\top
\left(
\sum_{k=0}^{\infty}(P\Gamma)^k
\right)
\mathbf 1
\notag\\
&=
\phi_{\max}\,
\xi^\top(I-P\Gamma)^{-1}\mathbf 1.
\label{eq:appendix-expected-trace-bound}
\end{align}

Returning to \eqref{eq:appendix-Llambda-definition}, the bounds
$0\leq1-\lambda_1\leq1$,
$\gamma_1\leq\gamma_{\max}$, and
$\|\phi(S_1)\|_2\leq\phi_{\max}$, together with conditioning on the
state--trace history through time $0$, give
\begin{align}
L_\lambda
&\leq
\gamma_{\max}\phi_{\max}
\mathbb E_\zeta\!\left[
\|e_0\|_2
\sum_{s'\in\mathcal S}
P^o_{S_0s'}\rho(S_0,s')
\right]
\notag\\
&=
\gamma_{\max}\phi_{\max}
\mathbb E_\zeta[\|e_0\|_2],
\label{eq:appendix-conditional-Llambda-bound}
\end{align}
where the last equality follows from
\begin{align*}
\sum_{s'\in\mathcal S}
P^o_{ss'}\rho(s,s')
=
\sum_{s'\in\mathcal S}P_{ss'}
=
1.
\end{align*}
Combining \eqref{eq:appendix-expected-trace-bound} and
\eqref{eq:appendix-conditional-Llambda-bound} proves
\eqref{eq:appendix-Llambda-bound}.
\end{proof}

\clearpage
\vskip 0.2in
\bibliography{refs}

\end{document}